%% file: main.tex
\documentclass[10pt,twocolumn,letterpaper]{article}

\usepackage[pagenumbers]{cvpr} %

\input{preamble}
\usepackage{pifont}%
\newcommand{\cmark}{{\color{ForestGreen} \ding{51}}}%
\newcommand{\xmark}{{\color{red} \ding{55}}}%

\newcommand{\rom}[1]{\uppercase\expandafter{\romannumeral #1\relax}}

\usepackage{xcolor}
\definecolor{ForestGreen}{rgb}{0.13, 0.55, 0.13}
\usepackage{calc}
\newcommand{\rotatedCentering}[3]{\rotatebox{#1}{\hspace{(#2-\widthof{#3})/2}#3}}

\usepackage{amsthm}
\usepackage{amssymb}

\usepackage{colortbl}
\usepackage{tabulary}
\usepackage{tabularx}
\usepackage{etoolbox}
\usepackage{multirow}
\usepackage{cuted}
\usepackage[percentage]{overpic}
\usepackage{booktabs}
\usepackage{pgfplots}
\pgfplotsset{compat=newest}%

\definecolor{cvprblue}{rgb}{0.21,0.49,0.74}
\usepackage[pagebackref,breaklinks,colorlinks,citecolor=cvprblue]{hyperref}

\def\paperID{7307} %
\def\confName{CVPR}
\def\confYear{2024}

\title{Spectral Meets Spatial: Harmonising 3D Shape Matching and Interpolation}

\author{Dongliang Cao\textsuperscript{1}
~~~
Marvin Eisenberger\textsuperscript{2}
~~~
Nafie El Amrani\textsuperscript{1}
~~~
Daniel Cremers\textsuperscript{2}
~~~
Florian Bernard\textsuperscript{1}\\
[2mm]
$^1$ University of Bonn \qquad
$^2$ Technical University of Munich\\
}

\def\pathOurs{figs/ours/}
\def\pathAFMaps{figs/attentivefmaps/}
\def\pathURSSM{figs/urssm/}
\def\srcEnd{_M}
\def\trgtEnd{_N}

\begin{document}
\maketitle

\setkeys{Gin}{keepaspectratio}

\begin{strip}
  \centerline{
  \footnotesize
  \begin{tabular}{c}
    \setlength{\tabcolsep}{0pt}
    \includegraphics[width=\textwidth]{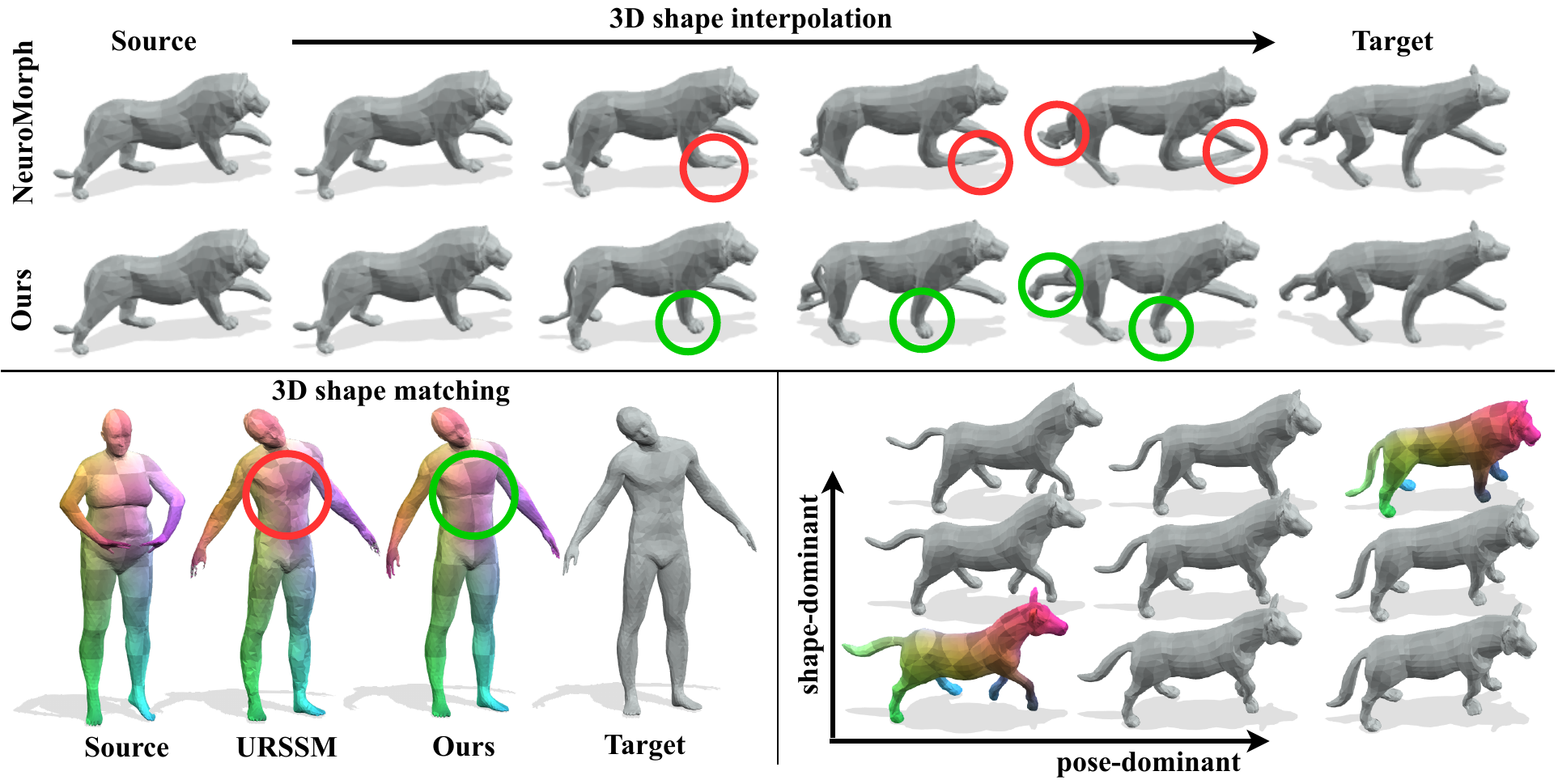}
  \end{tabular}
  }
\vspace{-0.4cm}
\captionof{figure}{\textbf{Top: 3D shape interpolation.} Compared to the SOTA shape interpolation method NeuroMorph~\cite{eisenberger2021neuromorph}, our method obtains more reliable interpolation even under large non-isometry. \textbf{Bottom left: 3D shape matching.} The point-wise correspondences found by the SOTA shape matching method URSSM~\cite{cao2023unsupervised} contains local mismatches. In contrast, our method enables smooth shape matching. \textbf{Bottom right: 3D shape matching and interpolation.} 
Our method is the first unsupervised method that obtains both accurate correspondences (shown as texture transfer) and realistic interpolation that capture both the pose-dominant and shape-dominant deformations.
}
\vspace{-0.3cm}
\label{fig:teaser}
\end{strip}

\input{sec/0_abstract}
\vspace{-0.2cm}
\input{sec/1_intro}

\input{sec/2_related_work}
\input{sec/3_background}
\input{sec/4_method}

\input{sec/5_experiments}
\input{sec/6_ablation}
\input{sec/7_conclusion}
\input{sec/8_acknowledgement}
{
    \small
    \bibliographystyle{ieeenat_fullname}
    \bibliography{main}
}

\input{sec/X_suppl}

\end{document}

%% file: sec/0_abstract.tex
\begin{abstract}
Although 3D shape matching and interpolation are highly interrelated, they are often studied separately and applied sequentially to relate different 3D shapes, thus resulting in sub-optimal performance. In this work we present a unified framework to predict both point-wise correspondences and shape interpolation between 3D shapes. To this end, we combine the deep functional map framework with classical surface deformation models to map shapes in both spectral and spatial domains. On the one hand, by incorporating spatial maps, our method obtains more accurate and smooth point-wise correspondences compared to previous functional map methods for shape matching. On the other hand, by introducing spectral maps, our method gets rid of commonly used but computationally expensive geodesic distance constraints that are only valid for near-isometric shape deformations. Furthermore, we propose a novel test-time adaptation scheme to capture both pose-dominant and shape-dominant deformations. Using different challenging datasets, we demonstrate that our method outperforms previous state-of-the-art methods for both shape matching and interpolation, even compared to supervised approaches.
\end{abstract}

%% file: sec/1_intro.tex
\section{Introduction}
\label{sec:intro}
Computing maps between 3D shapes is a fundamental problem in computer vision and computer graphics, since it opens the door to understanding different object categories and enables both shape analysis~\cite{loncaric1998survey} and shape generation~\cite{egger20203d}. One specific class of approaches builds shape relationships is statistical shape models (e.g.\ SMPL~\cite{loper2015smpl}). To do so, these methods require dense point-wise correspondences between large collections of 3D shapes. Even though shape matching has been studied extensively in the literature~\cite{tam2012registration,van2011survey}, finding accurate point-wise correspondences between non-rigidly deformed 3D shapes remains challenging due to different discretisation, non-isometry and partiality. Once shapes are in correspondence, shape interpolation methods can be applied to increase the shape diversity for shape generation~\cite{muralikrishnan2022glass,eisenberger2021neuromorph} and for different applications (e.g.\ animation~\cite{lewis2023pose}, interactive shape editing~\cite{huang2010interactive}). 

Despite the interrelation between shape matching and interpolation, prior works mainly consider them as two separate problems. In the context of shape matching, the functional map framework~\cite{ovsjanikov2012functional} is one of the most dominant pipelines and has been extended by many follow-up works~\cite{ren2018continuous,rodola2017partial,ren2021discrete,donati2022complex}. However, functional map methods only map shapes in the spectral domain and thus lead to local unsmooth point-wise correspondences (see URSSM in~\cref{fig:teaser} bottom left), which is undesirable for many downstream tasks (e.g.\ shape interpolation~\cite{eisenberger2020hamiltonian}, or statistical shape analysis~\cite{loper2015smpl,li2017flame}). Meanwhile, shape matching methods based on non-rigid registration (i.e.\ spatial maps) typically rely on time-consuming iterative optimisation schemes~\cite{huang2008non,li2008global} and careful initialisation~\cite{marin2020farm,bernard2020mina}, and thus often achieve worse performance due to the optimisation complexity~\cite{deng2022survey}. To overcome the above limitations, our method takes the best
of both worlds by combining spectral and spatial maps. For shape interpolation, most works assume that the shapes are already in correspondence and aim to design more efficient and realistic deformation energies~\cite{wirth2011continuum,heeren2012time,brandt2016geometric}. As a consequence, wrong correspondences often lead to undesirable interpolation trajectories. Only few approaches~\cite{eisenberger2019divergence,eisenberger2021neuromorph} explicitly consider shape matching and interpolation together, but mostly under overly strict assumptions (e.g.\ volume preservation).~\cref{tab:comparison} summarises the comparison between different learning-based shape matching approaches. Unlike prior methods, we propose the first unsupervised framework that harmonises spectral and spatial maps for both shape matching and interpolation. We summarise our main contributions as follows:
\begin{itemize}
    \item For the first time we fuse spectral and spatial maps to enable the joint unsupervised learning of both non-rigid 3D shape matching and interpolation.
    \item Our method predicts both pose-dominant and shape-dominant deformations, in contrast to prior interpolation methods that mainly focus on pose-dominant ones due to local deformation priors.
    \item We set the new state-of-the-art shape matching and interpolation performance on numerous challenging benchmarks, even compared to supervised methods.
\end{itemize}

\begin{table}[tbh!]
\centering
\small
\begin{tabular}{@{}lccccc@{}}
\toprule
Methods & Unsup. & Spectral & Spatial & Interp.  \\ \midrule
FMNet~\cite{litany2017deep} & \xmark & \cmark & \xmark  & \xmark \\
3D-CODED~\cite{groueix20183d} & \xmark & \xmark & \cmark  & \xmark  \\
GeomFMaps~\cite{donati2020deep} & \xmark & \cmark & \xmark  & \xmark \\
TransMatch~\cite{trappolini2021shape} & \xmark & \xmark & \cmark & \xmark \\
UnsupFMNet~\cite{halimi2019unsupervised} & \cmark & \cmark & \xmark  & \xmark  \\
SURFMNet~\cite{roufosse2019unsupervised} & \cmark & \cmark & \xmark  & \xmark \\
Deep Shells~\cite{eisenberger2020deep} & \cmark & \cmark & \cmark & \xmark \\
CorrNet3D~\cite{zeng2021corrnet3d} & \cmark & \xmark & \cmark & \xmark \\
NeuroMorph~\cite{eisenberger2021neuromorph} & \cmark & \xmark & \cmark & \cmark \\
AttnFMaps~\cite{li2022learning} & \cmark & \cmark & \xmark & \xmark \\
URSSM~\cite{cao2023unsupervised} & \cmark & \cmark & \xmark & \xmark \\
Ours & \cmark & \cmark & \cmark & \cmark \\ \bottomrule
\end{tabular}
\caption{\textbf{Learning-based shape matching method comparison.} Our method is the first unsupervised learning approach that operates in both spectral and spatial domains to enable accurate shape matching and realistic shape interpolation.}
\vspace{-0.3cm}
\label{tab:comparison}
\end{table}

%% file: sec/2_related_work.tex
\section{Related work}
\label{sec:related_work}
\subsection{Shape matching}
Shape matching aims to find point-wise correspondences between shapes. Classical shape matching approaches~\cite{windheuser2011geometrically,holzschuh2020simulated,roetzer2022scalable} establish correspondences by explicitly considering geometric relations. Other methods~\cite{huang2008non,ezuz2019elastic,eisenberger2019divergence,bernard2020mina} solve the problem based on non-rigid shape registration. However, directly finding point-wise correspondences often leads to complex optimisation problems that typically require time-consuming iterative optimisation strategies and are thus only applicable for low-resolution shapes~\cite{windheuser2011geometrically,roetzer2022scalable,roetzer2023fast}, which limits their applications in real-world settings.

In contrast, the functional map framework finds correspondences in the spectral domain~\cite{ovsjanikov2012functional}, where the correspondence relationship can be encoded into a small matrix, namely the functional map. Due to its simple yet efficient formulation, the functional map framework has been extended by many follow-up works, e.g.\ improving the matching accuracy~\cite{eynard2016coupled,ren2019structured}, extending it to more challenging scenarios (e.g.\ non-isometry~\cite{ren2018continuous,ren2021discrete,magnet2022smooth,eisenberger2020smooth}, partiality~\cite{rodola2017partial,litany2017fully}). Together with the development of deep learning, many learning-based approaches are proposed~\cite{litany2017deep,halimi2019unsupervised,roufosse2019unsupervised} and lead to state-of-the-art performance~\cite{cao2023self,attaiki2021dpfm,cao2023unsupervised,cao2022unsupervised}. Nevertheless, the map in the spectral domain does not guarantee smooth point-wise correspondences. Therefore, the conversion of functional map into smooth and accurate point-wise correspondences is still an open-problem that is extensively studied~\cite{vestner2017product,melzi2019zoomout,ezuz2019reversible,pai2021fast}. Unlike most functional map methods, our method harmonises the spectral and spatial maps together and thus leads to more accurate and smooth point-wise correspondences. 

\subsection{Shape interpolation}
Shape interpolation is a well-studied problem in computer graphics that aims to continuously deform a source
shape to a target shape. The most common strategy is to define an interpolation path in some higher dimensional space~\cite{brandt2016geometric,heeren2012time,heeren2014exploring,wirth2011continuum}. To this end, various deformation measurements (e.g.\ as-rigid-as-possible (ARAP)~\cite{sorkine2007rigid}, PriMo~\cite{botsch2006primo}, etc.) are proposed to optimise the interpolation path so that the local distortion between any consecutive shapes is minimised. Another direction is to interpolate intrinsic quantities like dihedral angles before reconstructing the 3D shape~\cite{alexa2000rigid,baek2015isometric}. Other methods formulate shape interpolation as a time-dependent gradient flow~\cite{charpiat2007generalized,eckstein2007generalized} by incorporating certain constraints (e.g.\ volume preservation~\cite{eisenberger2019divergence,eisenberger2020hamiltonian}). Nevertheless, most of them assume known correspondences between two shapes, which is unrealistic for real-world 3D shapes. The most relevant approach to our work is NeuroMorph~\cite{eisenberger2021neuromorph}, which is an unsupervised framework for both shape matching and interpolation. However, it has several limitations: In the context of shape matching, it is based on the geodesic distance preservation that is computationally expensive and only valid for near-isometric shape deformation. In the context of shape interpolation, it solely utilises the as-rigid-as-possible~\cite{sorkine2007rigid} to deform the source shape into the target shape, which can only model the low-frequency pose-dominant deformations (see Fig.\ 1 in~\cite{eisenberger2021neuromorph}). In contrast, we utilise more efficient spectral regularisation and thus lead to more accurate shape matching even under large non-isometry. Furthermore, our method is able to capture both pose-dominant and shape-dominant deformations based on our novel test-time adaptation. 

%% file: sec/3_background.tex
\section{Deep functional map in a nutshell}
\label{sec:background}
In this section we provide a summary of the popular deep functional map framework~\cite{roufosse2019unsupervised}. 
We consider a pair of 3D shapes $\mathcal{X}$ and $\mathcal{Y}$ represented as triangle meshes with $n_{\mathcal{X}}$ and $n_{\mathcal{Y}}$ (w.l.o.g.\ $n_{\mathcal{X}} \leq n_{\mathcal{Y}}$) vertices, respectively. Here we summarise the main steps of its pipeline:
\begin{enumerate}
    \item Compute the Laplacian matrices $\mathbf{L}_{\mathcal{X}}, \mathbf{L}_{\mathcal{Y}}$~\cite{pinkall1993computing} and the corresponding first $k$ eigenfunctions $\mathbf{\Phi}_{\mathcal{X}}, \mathbf{\Phi}_{\mathcal{Y}}$ and eigenvalues $\mathbf{\Lambda}_{\mathcal{X}}, \mathbf{\Lambda}_{\mathcal{Y}}$ in matrix notation, respectively.
    \item Compute feature vectors $\mathbf{F}_{\mathcal{X}}, \mathbf{F}_{\mathcal{Y}}$ defined on each shape via a learnable feature extractor $\mathcal{F}_{\theta}$.
    \item Compute the functional maps $\mathbf{C}_{\mathcal{XY}}, \mathbf{C}_{\mathcal{YX}}$ associated with the Laplacian eigenfunctions by solving the optimisation problem
    \begin{equation}
    \label{eq:fmap}
        \mathbf{C}_{\mathcal{XY}}=\arg\min_{\mathbf{C}}~ E_{\mathrm{data}}\left(\mathbf{C}\right)+\lambda E_{\mathrm{reg}}\left(\mathbf{C}\right).
    \end{equation}
    Here, minimising $E_{\mathrm{data}}\left(\mathbf{C}\right)=\left\|\mathbf{C}\mathbf{\Phi}_{\mathcal{X}}^{\dagger}\mathbf{F}_{\mathcal{X}}-\mathbf{\Phi}_{\mathcal{Y}}^{\dagger}\mathbf{F}_{\mathcal{Y}}\right\|^{2}_{F}$ enforces descriptor preservation, while minimising the regularisation term $E_{\mathrm{reg}}$ imposes certain structural properties (e.g.\ Laplacian commutativity~\cite{ovsjanikov2012functional}).
    \item During training, structural regularisation (e.g.\ orthogonality, bijectivity~\cite{roufosse2019unsupervised}) is imposed on the functional maps, i.e.
    \begin{equation}
        \label{eq:l_struct}
        L_{\mathrm{struct}} = \lambda_{\mathrm{bij}}L_{\mathrm{bij}} + \lambda_{\mathrm{orth}}L_{\mathrm{orth}}.
    \end{equation}
    \item During inference, the point-wise map $\mathbf{\Pi}_{\mathcal{YX}}$ is obtained based on the map relationship $\mathbf{C}_{\mathcal{XY}} = \mathbf{\Phi}_{\mathcal{Y}}^{\dagger}\mathbf{\Pi}_{\mathcal{YX}}\mathbf{\Phi}_{\mathcal{X}}$, e.g.\ either by nearest neighbour search in the spectral domain or by other post-processing techniques~\cite{vestner2017product,melzi2019zoomout,eisenberger2020smooth,ezuz2019reversible}.
\end{enumerate} 

%% file: sec/4_method.tex
\section{Our method}
\label{sec:method}
\begin{figure*}
  \centering
  \includegraphics[width=\linewidth]{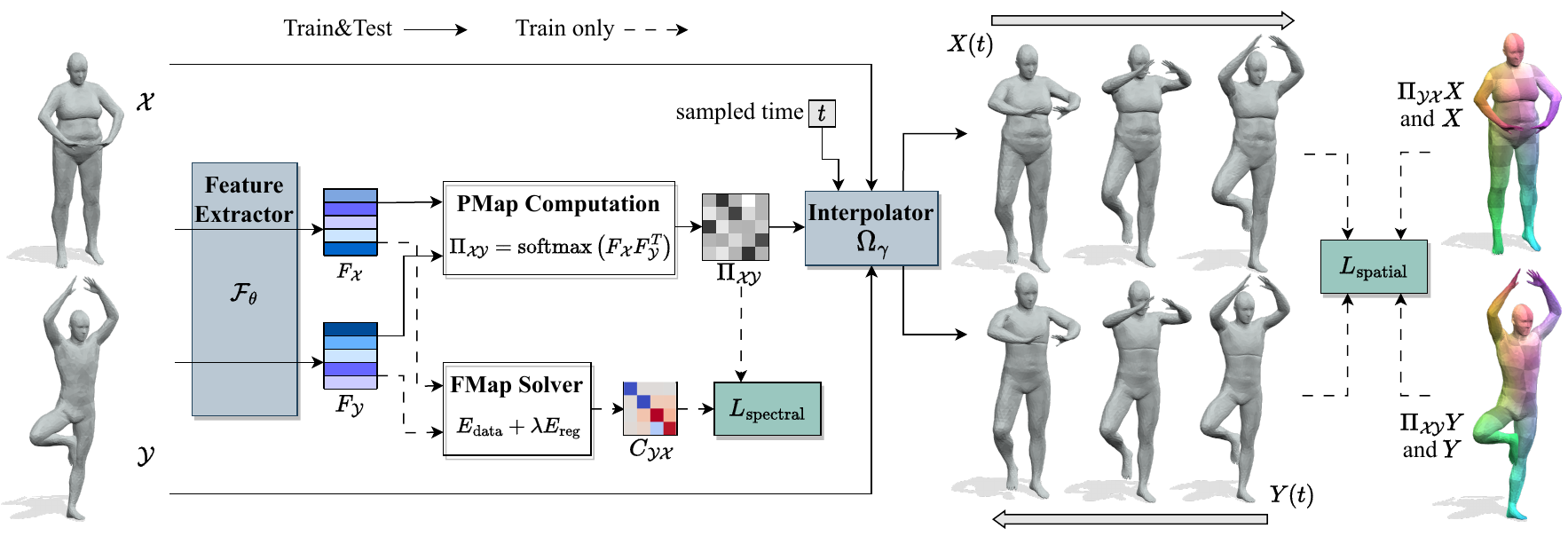}
  \caption{\textbf{Method overview.} First, the Siamese feature extractor $\mathcal{F}_{\theta}$ is used to extract features $\mathbf{F}_{\mathcal{X}}, \mathbf{F}_{\mathcal{Y}}$ from input shapes $\mathcal{X}, \mathcal{Y}$, respectively. The extracted features are then used to compute both bidirectional functional maps $\mathbf{C}_{\mathcal{XY}}, \mathbf{C}_{\mathcal{YX}}$ and point-wise maps $\mathbf{\Pi}_{\mathcal{XY}}, \mathbf{\Pi}_{\mathcal{YX}}$. Afterwards, the computed point-wise maps are used to bring the shape into correspondences. In the context of shape interpolation, a series of time steps $t$ is sampled and fed into the interpolator $\Omega_{\gamma}$ together with shape information to predict a series of interpolated shapes ${X}(t), Y(t)$. During training the spectral loss $L_{\mathrm{spectral}}$ is used to regularise the predicted functional maps and point maps, while the spatial loss $L_{\mathrm{spatial}}$ is used to regularise the interpolation trajectories of both shapes. }
  \label{fig:framework}
\end{figure*}

The framework of our method is depicted in~\cref{fig:framework}. Our method aims to predict both accurate point-wise correspondences and realistic shape interpolation paths in an unsupervised manner. Therefore, our framework can be divided into two sub-modules, namely the shape matching module (\cref{subsec:matching}) and the shape interpolation module (\cref{subsec:interpolation}). Our unsupervised loss and test-time adaptation are introduced in~\cref{subsec:unsup_loss} and~\cref{subsec:tta}, respectively. 

\subsection{Shape matching module}
\label{subsec:matching}
\noindent \textbf{Point-wise map computation.}
Point-wise map computation aims to find dense correspondences between two shapes. In theory, the point-wise map $\mathbf{\Pi}_{\mathcal{XY}}$ should be a (partial) permutation matrix, i.e.
    \begin{equation}
        \label{eq:permutation_mat}
        \left\{\mathbf{\Pi} \in\{0,1\}^{n_{\mathcal{X}} \times n_{\mathcal{Y}}}: \mathbf{\Pi} \mathbf{1}_{n_{\mathcal{Y}}} = \mathbf{1}_{n_{\mathcal{X}}}, \mathbf{1}_{n_{\mathcal{X}}}^{\top} \mathbf{\Pi} \leq \mathbf{1}_{n_{\mathcal{Y}}}^{\top}\right\},
    \end{equation}
where $\mathbf{\Pi}(i,j)$ indicates whether the vertex $\mathbf{X}^{i}$ in shape $\mathcal{X}$ corresponds to the vertex $\mathbf{Y}^{j}$ in shape $\mathcal{Y}$. In our work the point-wise correspondences $\mathbf{\Pi}_{\mathcal{XY}}$ between shapes $\mathcal{X}$ and $\mathcal{Y}$ are obtained based on similarity measurement of the extracted features. To this end, the Siamese feature extractor $\mathcal{F}_{\theta}$ (i.e.\ DiffusionNet~\cite{sharp2020diffusionnet}) is used to extract features $\mathbf{F}_{\mathcal{X}}, \mathbf{F}_{\mathcal{Y}}$ from input shapes, respectively. Following prior works~\cite{cao2023unsupervised,eisenberger2021neuromorph}, we use the softmax operator to produce a soft correspondence matrix to make the computation differentiable, i.e. $\mathbf{\Pi}_{\mathcal{XY}} = \mathrm{Softmax}\left( {\mathbf{F}_{\mathcal{X}}\mathbf{F}_{\mathcal{Y}}^{T}} \right).$
The softmax operator is applied in each row to ensure that elements are non-negative and $\mathbf{\Pi}_{\mathcal{XY}} \mathbf{1}_{n_{\mathcal{Y}}} = \mathbf{1}_{n_{\mathcal{X}}}$. In this way, $\mathbf{\Pi}_{\mathcal{XY}}$ can be interpreted as a soft assignment of vertices $\mathbf{Y}$ in shape $\mathcal{Y}$ to vertices $\mathbf{X}$ in shape $\mathcal{X}$. 

\noindent \textbf{Functional map computation.} The goal of the functional map computation is to impose spectral regularisation on the obtained point-wise maps based on the coupling relationship $\mathbf{C}_{\mathcal{XY}} = \mathbf{\Phi}_{\mathcal{Y}}^{\dagger}\mathbf{\Pi}_{\mathcal{YX}}\mathbf{\Phi}_{\mathcal{X}}$. Therefore, the functional map computation is only used for training as part of the spectral loss computation. The computation pipeline of functional maps is explained in~\cref{sec:background} and the details of the spectral loss can be found in~\cref{subsec:unsup_loss}.
\subsection{Shape interpolation module}
\label{subsec:interpolation}
The goal of the shape interpolation module $\Omega_{\gamma}$ is to predict the deformation field $\mathbf{\Delta}(t) \in \mathbb{R}^{n \times 3}$, where $t \in \left[0,1\right]$, that continuously deforms the shape $\mathcal{X}$ into the shape $\mathcal{Y}$, and vice versa. The deformation field of shape $\mathcal{X}$ forms the interpolation trajectory
    \begin{equation}
        \label{eq:deform}
        \mathbf{X}(t) = \mathbf{X} + \mathbf{\Delta}_{\mathcal{X}}(t). 
    \end{equation}
The interpolation trajectory shifts shape $\mathcal{X}$ from its original locations $\mathbf{X}(0) = \mathbf{X}$ into new locations $\mathbf{X}(1) \approx \mathbf{\Pi}_{\mathcal{XY}}\mathbf{Y}$, which are close to the corresponding vertices in shape $\mathcal{Y}$. Following NeuroMorph~\cite{eisenberger2021neuromorph}, we use EdgeConv~\cite{wang2019dynamic} blocks with residual connections~\cite{he2016deep} to build our interpolator $\Omega_{\gamma}$. Similarly, the input feature of shape $\mathcal{X}$ in the interpolator is the 7-dimensional feature vector $\mathbf{Z}_{\mathcal{X}} \in \mathbb{R}^{n_{\mathcal{X}} \times 7}$ that can be expressed in the form
    \begin{equation}
        \mathbf{Z}_{\mathcal{X}}(t) = \left(\mathbf{X}, \mathbf{\Pi}_{\mathcal{XY}}\mathbf{Y} - \mathbf{X}, t\mathbf{1}_{n_{\mathcal{X}}}\right),
    \end{equation}
where $\mathbf{X}$ is the vertices of the shape $\mathcal{X}$, $\mathbf{\Pi}_{\mathcal{XY}}\mathbf{Y} - \mathbf{X}$ is the vertex offsets between shape $\mathcal{X}$ and $\mathcal{Y}$, $t\mathbf{1}_{n_{\mathcal{X}}}$ is the sampled time that broadcasts to all vertices. 
The deformation fields are given by the output of the interpolator multiplied by time, i.e.\ $\mathbf{\Delta}_{\mathcal{X}}(t) = t\cdot\Omega_{\gamma}(\mathbf{Z}_{\mathcal{X}}(t))$. To this end, the interpolator can immediately predict a trivial solution (i.e.\ linear interpolation) by setting $\Omega_{\gamma}(\mathbf{Z}_{\mathcal{X}}(t)) = \mathbf{\Pi}_{\mathcal{XY}}\mathbf{Y} - \mathbf{X}$, since it satisfies the boundary conditions of the interpolation~\cite{eisenberger2021neuromorph}:
    \begin{align}
        \mathbf{X}(0) &= \mathbf{X} + 0\cdot(\mathbf{\Pi}_{\mathcal{XY}}\mathbf{Y} - \mathbf{X}) = \mathbf{X}, \\
        \mathbf{X}(1) &= \mathbf{X} + 1\cdot(\mathbf{\Pi}_{\mathcal{XY}}\mathbf{Y} - \mathbf{X}) = \mathbf{\Pi}_{\mathcal{XY}}\mathbf{Y}.
    \end{align}
However, the linear interpolation is a degenerate solution, since our goal is to obtain geometrically realistic shape interpolation. To prevent the interpolator prediction from falling into the trivial solution, spatial regularisation needs to be imposed, including a deformation energy~\cite{sorkine2007rigid} and additional regularisation, which will be introduced in~\cref{subsec:unsup_loss}.
\subsection{Unsupervised loss}
\label{subsec:unsup_loss}
In this section we introduce the spectral and spatial regularisation that enable unsupervised training of our method.
\noindent \textbf{Spectral regularisation.} To obtain accurate point-wise correspondences for shape matching and shape interpolation, we utilise a spectral regularisation based on the functional map framework. Unlike prior works~\cite{cosmo2020limp,eisenberger2021neuromorph} relying on geodesic distance preservation that is computationally expensive and only valid for near-isometric deformation~\cite{halimi2019unsupervised}, the spectral regularisation imposed on the functional maps is more efficient and works well also for non-isometric deformations~\cite{li2022learning,cao2023unsupervised,donati2022deep}. Unlike most previous deep functional map methods~\cite{roufosse2019unsupervised,eisenberger2020deep,donati2020deep} relying on off-the-shelf post-processing techniques~\cite{melzi2019zoomout,eisenberger2020smooth} to obtain point-wise maps (see~\cref{sec:background}), we aim to directly obtain point-wise map based on deep feature similarity. To this end, we combine structural regularisation in~\cref{eq:l_struct} with the coupling loss~\cite{cao2023unsupervised,ren2021discrete,sun2023spatially} to build our spectral regularisation  
\begin{equation}
    \label{eq:l_fmap}
    L_{\mathrm{spectral}} = \lambda_{\mathrm{struct}} L_{\mathrm{struct}} + \lambda_{\mathrm{couple}} L_{\mathrm{couple}},
\end{equation}
where
\begin{equation}
    \label{eq:couple}
    L_{\mathrm{couple}} = \left\|\mathbf{C}_{\mathcal{XY}} - \mathbf{\Phi}_{\mathcal{Y}}^{\dagger}\mathbf{\Pi}_{\mathcal{YX}}\mathbf{\Phi}_{\mathcal{X}}\right\|^{2}_{F}. 
\end{equation}

\noindent \textbf{Spatial regularisation.} Finding point-wise correspondences solely based on spectral regularisation leads to local mismatches, which is undesirable for many downstream tasks including shape interpolation. To compensate for this, one of our spatial regularisation terms aims to directly align two shapes in the spatial domain, i.e. 
\begin{equation}
    \label{eq:align}
    L_{\mathrm{align}} = \left\|\mathbf{X}(1) - \mathbf{\Pi}_{\mathcal{XY}}\mathbf{Y}\right\|^{2}_{F} + \left\|\mathbf{Y}(1) - \mathbf{\Pi}_{\mathcal{YX}}\mathbf{X}\right\|^{2}_{F}.
\end{equation}
As mentioned in~\cref{subsec:interpolation}, additional spatial regularisation for shape deformation is needed to avoid trivial degenerated solutions. To this end, similar to~\cite{eisenberger2021neuromorph,zhou2020unsupervised}, the ARAP deformation energy~\cite{sorkine2007rigid} is used to regularise the interpolation trajectory. ARAP is a low-frequency preserving deformation energy that encourages locally rigid transformations. During training, we sample $T+1$ discrete timesteps uniformly $\mathbf{X}_k:=\mathbf{X}(k/T)$ for $k\in\{0,\dots,T\}$ to compute the ARAP energy as
\begin{multline}
    \label{eq:arap}
    E_{\mathrm{arap}}(\mathbf{X}_k, \mathbf{X}_{k+1}) = \\
    \min_{\mathbf{R}_i\in SO(3)}\sum_{i=1}^{n_{\mathcal{X}}}\sum_{j\in \mathcal{N}_{i}}\left\|\mathbf{R}_i\mathbf{E}_k^{ij} - \mathbf{E}_{k+1}^{ij}\right\|^{2}_{F},
\end{multline}
where $\mathbf{E}_k^{ij} = \mathbf{X}_k^{i} - \mathbf{X}_k^{j}\in\mathbb{R}^3$ is the triangle edge between mesh vertices $\mathbf{X}_k^{j}$ and $\mathbf{X}_k^{i}$. The rotation matrices $\mathbf{R}_i$ can be computed in closed form for an efficient deformation (see~\cite{sorkine2007rigid} for details). Finally, our ARAP regularisation term can be expressed in the form
\begin{equation}
    \label{eq:arap_loss}
    L_{\mathrm{arap}} = 
    \sum_{k=0}^{T-1}E_{\mathrm{arap}}(\mathbf{X}_k, \mathbf{X}_{k+1}). 
\end{equation}
Analogously, the loss in~\cref{eq:arap_loss} is also applied to the reverse sequence for shape $\mathcal{Y}$. However, the ARAP deformation energy only enables low-frequency pose-dominant deformation. To capture the shape-dominant deformation, we propose a test-time adaptation described in~\cref{subsec:tta}. Apart from the ARAP regularisation, we propose two additional regularisation terms to encourage smooth and symmetric deformation trajectories from shape $\mathcal{X}$ to shape $\mathcal{Y}$, and vice versa. The symmetry loss encourages symmetric deformation trajectories from both directions, i.e.
\begin{equation}
    \label{eq:sym_loss}
    L_{\mathrm{sym}} =
    \sum_{k=1}^{T-1}\left\|\mathbf{X}_k - \mathbf{\Pi}_{\mathcal{XY}}\mathbf{Y}_{T-k}\right\|^{2}_{F}.
\end{equation}
This loss is also applied with the roles of $\mathbf{X}$ and $\mathbf{Y}$ swapped. Since we would like to also model the shape-dominant deformation, we only want $\mathbf{X}_k \approx \mathbf{\Pi}_{\mathcal{XY}}\mathbf{Y}_{T-k}$ allowing for shape-based variations. To this end, we introduce a temporal shape variance loss defined as
\begin{equation}
    \label{eq:var_loss}
    L_{\mathrm{var}} = \sum_{i=1}^{n_{\mathcal{X}}}\mathrm{Var}\left(\left\|\mathbf{X}_k - \mathbf{\Pi}_{\mathcal{XY}}\mathbf{Y}_{T-k}\right\|^{i}\right),
\end{equation}
where $\mathrm{Var}(\mathbf{X}^{i}) = \frac{1}{T-1}\sum_{k=1}^{T}(\mathbf{X}^{i}_k - \Bar{\mathbf{X}}^{i})^2 $ is the alignment error variance of the $i$-th vertex during the interpolation sequence and $ \Bar{\mathbf{X}}^{i} = \frac{1}{T}\sum_{k=1}^{T}\mathbf{X}^{i}_k$ is the mean. In this manner, we assume that shape variations are nearly constant throughout the interpolation sequence. Similar to~\cref{eq:sym_loss}, the loss is also applied for the reverse direction $\mathbf{Y}$. 
The spatial loss is the linear combination of the above losses, i.e.
\begin{equation}
    \label{eq:spatial}
    L_{\mathrm{spatial}} = \lambda_{\mathrm{align}} L_{\mathrm{align}} + \lambda_{\mathrm{arap}} L_{\mathrm{arap}} + \lambda_{\mathrm{sym}} L_{\mathrm{sym}} + \lambda_{\mathrm{var}} L_{\mathrm{var}}.
\end{equation}
The total loss is the sum of the spectral loss and spatial loss.
\subsection{Test-time adaptation}
\label{subsec:tta}

\begin{figure}[!ht]
    \begin{center}  \includegraphics[width=1.05\columnwidth]{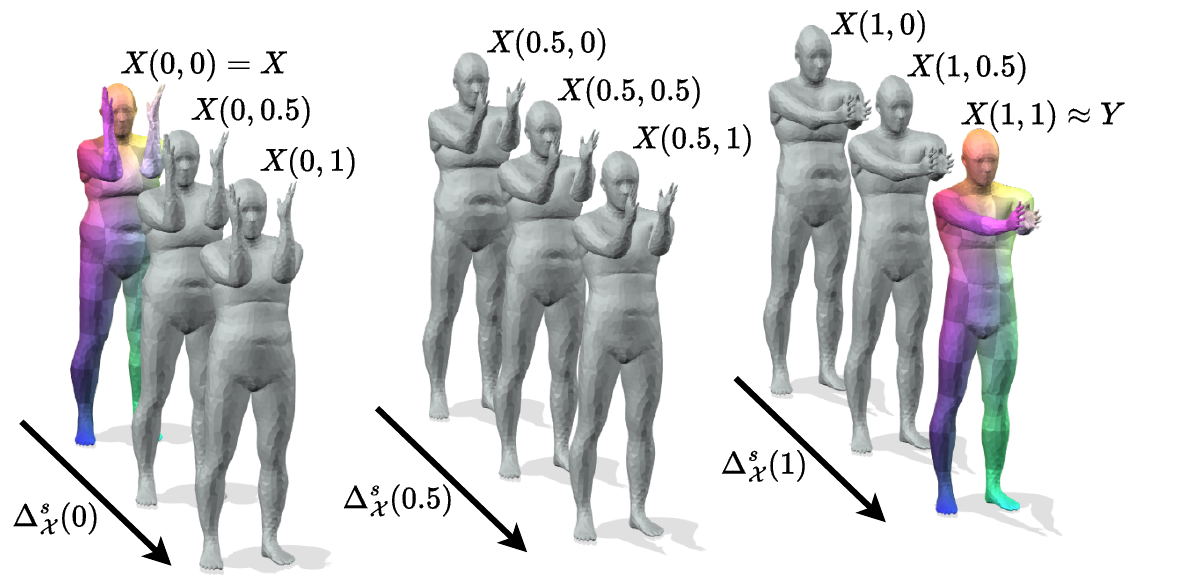}
    \caption{\textbf{Visualisation of our test-time adaptation.} Starting from pose-dominant deformation based on ARAP, we obtain $\mathbf{X}(t, t_s=0)$ (i.e.\ the last row). Afterwards, we optimise the shape-dominant deformation field $\mathbf{\Delta}_{s}(t)$ for each sampled time point. Finally, the linear interpolation is performed to obtain the shape-dominant interpolation trajectories (shown by arrows).}
    \vspace{-0.4cm}
    \label{fig:tta}
    \end{center}
\end{figure}

The spatial loss defined in~\cref{eq:spatial} encourages low-frequency pose-dominant deformations, while the shape-based deformations are not well-modelled. To this end, we propose a novel test-time adaptation to obtain more accurate correspondences and better capture the shape-dominant deformation. Specifically, we optimise for an additional shape-dominant deformation field $\mathbf{\Delta}_{s}(t) \in \mathbb{R}^{n_\mathcal{X} \times 3}$ (with $t \in \left[0,1\right]$) between $\mathbf{X}(t)$ and $\mathbf{Y}(1-t)$:
\begin{equation}
    \label{eq:tta}
    \mathbf{\Delta}_{s}(t) = \arg\min_{\mathbf{\Delta}} \mathrm{CD}\left(\mathbf{X}(t) + \mathbf{\Delta}, \mathbf{Y}(1-t)\right)
    + \lambda_{\mathrm{D}}L_{\mathrm{D}}(\mathbf{\Delta}),
\end{equation}
where we consider the standard Chamfer distance
\begin{multline}
    \label{eq:chamfer}
    \mathrm{CD}\left(\mathbf{S}_1, \mathbf{S}_2\right)=\frac{1}{\left|\mathbf{S}_1\right|} \sum_{\mathbf{x} \in \mathbf{S}_1} \min _{\mathbf{y} \in \mathbf{S}_2}\|\mathbf{x}-\mathbf{y}\|_2^2\\
    +\frac{1}{\left|\mathbf{S}_2\right|} \sum_{\mathbf{y} \in \mathbf{S}_2} \min _{\mathbf{x} \in \mathbf{S}_1}\|\mathbf{x}-\mathbf{y}\|_2^2,
\end{multline}
and $L_{\mathrm{D}}(\cdot)$ is the Dirichlet energy~\cite{magnet2022smooth} which promotes smooth deformations. The VectorAdam~\cite{ling2022vectoradam} optimiser is used to optimise the additional shape-based deformation field. Once we obtain the optimal $\mathbf{\Delta}_{s}(t)$, we use linear interpolation to obtain the shape-dominant deformation trajectory, i.e.
\begin{equation}
    \label{eq:linear}
    \mathbf{X}(t, t_s) = (1-t_s) \cdot (\mathbf{X}(t) + \mathbf{\Delta}_{s}(t)) +  t_s \cdot \mathbf{Y}(1-t), 
\end{equation}
where $t_s \in \left[0,1\right]$ is the sampled time for shape-dominant deformation.~\cref{fig:tta} demonstrates an example of our test-time adaptation for $t = \{0, 0.5, 1\}$ and $t_s = \{0, 0.5, 1\}$.

After the optimisation of the shape-dominant deformation field $\mathbf{\Delta}_{s}(t)$, we obtain the final point-wise correspondences based on the spatial map using
\begin{equation}
    \label{eq:final_pmap}
    \mathbf{\Pi}_{\mathcal{XY}} = \mathrm{NN}\left(\mathbf{Y}, \mathbf{X}(1, 1)\right),
\end{equation}
where $\mathrm{NN}$ denotes nearest neighbour search in $\mathbf{Y}$ for each entry in $\mathbf{X}(1, 1)$. Optimisation based on the Chamfer distance is highly non-convex and prone to local minima for non-rigid shape registration~\cite{groueix20183d,zeng2021corrnet3d,sundararaman2022implicit}. However, for the optimisation in~\cref{eq:tta}, the two shapes are already in the same pose with only minor shape variations.

%% file: sec/5_experiments.tex
\section{Experimental results}
\label{sec:experiments}
In this section we demonstrate the advantages of our method in the context of 3D shape matching and interpolation. 
\subsection{Shape matching}
\label{subsec:matching_results}
\noindent \textbf{Near-isometric shape matching.} We evaluate the matching accuracy of our method on three near-isometric benchmarks: FAUST~\cite{bogo2014faust}, SCAPE \cite{anguelov2005scape} and SHREC'19~\cite{melzi2019shrec}. Instead of the original meshes, we consider the more challenging remeshed versions from~\cite{ren2018continuous,donati2020deep}. The FAUST dataset consists of 10 humans with 10 different poses each, {where the training and testing split is 80/20}. The SCAPE dataset contains 71 different poses of the same person, split into 51 and 20 shapes for training and testing. The SHREC'19 dataset is subject to particular challenges with significant variance in the mesh connectivity and shape geometry. It has a total of 430 pairs for evaluation.

\begin{figure}[ht]
    \centering
    \begin{tabular}{cc}
     \hspace{-1.2cm}
     \input{figs/tikz/faust_pck}&
     \hspace{-1cm}
     \input{figs/tikz/scape_pck}
    \end{tabular}
    \caption{{\textbf{Near-isometric shape matching on FAUST, SCAPE.} Proportion of correct keypoints (PCK) curves and corresponding area under curve (scores in the legend) of our method in comparison to existing state-of-the-art methods.}
    }
    \vspace{-0.3cm}
    \label{fig:iso_pck}
\end{figure}
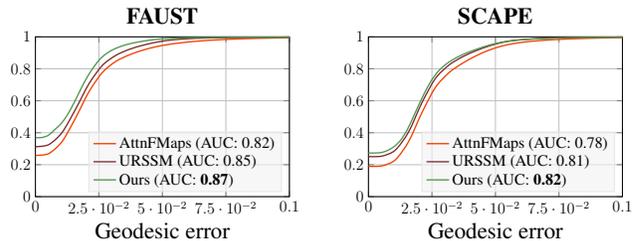

\noindent \textbf{Results.} The mean geodesic error~\cite{kim2011blended} is used as evaluation metric. We compare our method with state-of-the-art axiomatic, supervised and unsupervised methods. The results are summarised in~\cref{tab:near_isometric}. Our method outperforms the previous state-of-the-art methods (see the PCK curves in~\cref{fig:iso_pck}). Meanwhile, our method achieves substantially better cross-dataset generalisation ability than existing learning-based methods.

\begin{table}[ht!]
\small
\setlength{\tabcolsep}{2pt}
\centering
\small
\begin{tabular}{@{}lccc@{}}
\toprule
\multicolumn{1}{l}{Train}  & \multicolumn{1}{c}{\textbf{FAUST}}   & \multicolumn{1}{c}{\textbf{SCAPE}}  & \multicolumn{1}{c}{\textbf{FAUST + SCAPE}} \\ \cmidrule(lr){2-2} \cmidrule(lr){3-3} \cmidrule(lr){4-4}
\multicolumn{1}{l}{Test} & \multicolumn{1}{c}{\textbf{FAUST}} &  \multicolumn{1}{c}{\textbf{SCAPE}} &  \multicolumn{1}{c}{\textbf{SHREC'19}}
\\ \midrule
\multicolumn{4}{c}{Axiomatic Methods} \\
\multicolumn{1}{l}{BCICP~\cite{ren2018continuous}} & \multicolumn{1}{c}{6.1}  & \multicolumn{1}{c}{11.0} & \multicolumn{1}{c}{-}\\
\multicolumn{1}{l}{ZoomOut~\cite{melzi2019zoomout}} & \multicolumn{1}{c}{6.1} & \multicolumn{1}{c}{7.5} &  \multicolumn{1}{c}{-}\\
\multicolumn{1}{l}{Smooth Shells~\cite{eisenberger2020smooth}} & \multicolumn{1}{c}{2.5}  & \multicolumn{1}{c}{4.7} & \multicolumn{1}{c}{-}\\ 
\multicolumn{1}{l}{DiscreteOp~\cite{ren2021discrete}} & \multicolumn{1}{c}{5.6}  & \multicolumn{1}{c}{13.1} & \multicolumn{1}{c}{-}\\ 
\midrule
\multicolumn{4}{c}{Supervised Methods} \\ 
\multicolumn{1}{l}{FMNet~\cite{litany2017deep}} & \multicolumn{1}{c}{11.0} & \multicolumn{1}{c}{17.0} & \multicolumn{1}{c}{-} \\

\multicolumn{1}{l}{GeomFMaps~\cite{donati2020deep}}& \multicolumn{1}{c}{2.6} & \multicolumn{1}{c}{3.0} & \multicolumn{1}{c}{7.9}\\

\midrule
\multicolumn{4}{c}{Unsupervised Methods} \\

\multicolumn{1}{l}{WSupFMNet~\cite{sharma2020weakly}} & \multicolumn{1}{c}{3.8}  & \multicolumn{1}{c}{4.4}  & \multicolumn{1}{c}{-} \\

\multicolumn{1}{l}{Deep Shells~\cite{eisenberger2020deep}} & \multicolumn{1}{c}{1.7}  & \multicolumn{1}{c}{2.5} &  \multicolumn{1}{c}{21.1} \\

\multicolumn{1}{l}{NeuroMorph~\cite{eisenberger2021neuromorph}} & \multicolumn{1}{c}{2.3}  & \multicolumn{1}{c}{5.6} &  \multicolumn{1}{c}{8.5} \\

\multicolumn{1}{l}{DUO-FMNet~\cite{donati2022deep}}  & \multicolumn{1}{c}{2.5}  & \multicolumn{1}{c}{2.6} & \multicolumn{1}{c}{6.4} \\
\multicolumn{1}{l}{AttnFMaps~\cite{li2022learning}}  & \multicolumn{1}{c}{1.9}  & \multicolumn{1}{c}{2.2} & \multicolumn{1}{c}{5.8}\\
\multicolumn{1}{l}{URSSM~\cite{cao2023unsupervised}}  & \multicolumn{1}{c}{1.6}  & \multicolumn{1}{c}{1.9} &  \multicolumn{1}{c}{4.6}\\
\multicolumn{1}{l}{Ours}  & \multicolumn{1}{c}{\textbf{1.4}}  & \multicolumn{1}{c}{\textbf{1.8}} & \multicolumn{1}{c}{\textbf{3.2}} \\\hline
\end{tabular}
\caption{\textbf{Near-isometric shape matching and cross-dataset generalisation on FAUST, SCAPE and SHREC'19.} Our method outperforms prior axiomatic, supervised and unsupervised methods and demonstrates superior cross-dataset generalisation ability.}
\vspace{-0.3cm}
\label{tab:near_isometric}
\end{table}

\noindent \textbf{Non-isometric shape matching.} Regarding non-isometric shape matching, we consider the SMAL dataset~\cite{zuffi20173d} and DT4D-H dataset~\cite{magnet2022smooth}. The SMAL dataset contains 49 four-legged animal shapes of 8 species~\cite{eisenberger2023g}. {Following~\citet{donati2022deep}, we use the last 20 shapes for testing, resulting in a 29/20 split of the dataset.} DT4D-H dataset is based on the large-scale animation dataset DeformingThings4D~\cite{li20214dcomplete}. Nine classes of humanoid shapes are used for evaluation, {resulting in 198/95 shapes for training/testing.}

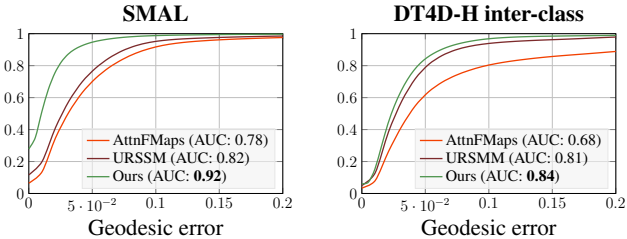
\begin{figure}[ht]
    \centering
    \begin{tabular}{cc}
     \hspace{-1.2cm}
     \input{figs/tikz/smal_pck}&
     \hspace{-1cm}
     \input{figs/tikz/dt4d_inter_pck}
    \end{tabular}
    \caption{{\textbf{Non-isometric matching on SMAL, DT4D-H inter-class datasets.} Our method sets to new state of the art by a large margin based on the combination of spectral and spatial maps.}
    }
    \vspace{-0.3cm}
    \label{fig:noniso_pck}
\end{figure}

\begin{figure}[ht!]
    \centering
    \input{figs/our_matching_results}
    \caption{ 
    \textbf{Non-isometric matching on SMAL and DT4D-H.} Our method obtains more accurate matching results even in the presence of large non-isometric shape deformations.}
    \vspace{-0.4cm}
    \label{fig:non-isometric-qualitative}
\end{figure}
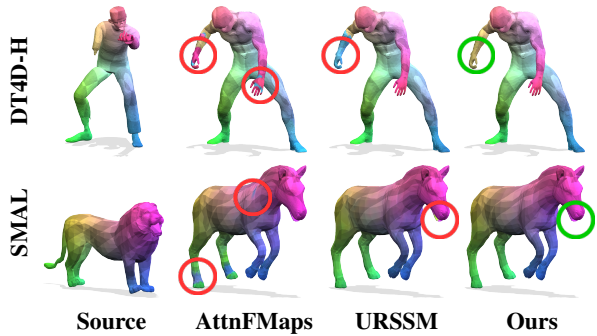

\noindent \textbf{Results.} The results are summarised in~\cref{tab:non-isometry}. Our method substantially outperforms the previous state-of-the-art in the context of non-isometric shape matching, even compared to recent supervised methods. In~\cref{fig:noniso_pck}, we summarise the PCK curves of our method compared to state-of-the-art methods on both non-isometric datasets.~\cref{fig:non-isometric-qualitative} provides qualitative results of our method compared to them.

\begin{table}[ht!]
    
    \small
    \centering
    \begin{tabular}{@{}lccc@{}}
    \toprule
    \multicolumn{1}{l}{\multirow{2}{*}{\textbf{Geo. error ($\times$100)}}}  & \multicolumn{1}{c}{\multirow{2}{*}{\textbf{SMAL}}}   & \multicolumn{2}{c}{\textbf{DT4D-H}}\\ \cmidrule(lr){3-4}
    &  & \multicolumn{1}{c}{\textbf{intra-class}} & \multicolumn{1}{c}{\textbf{inter-class}}
    \\ \midrule
    \multicolumn{4}{c}{Axiomatic Methods} \\
    \multicolumn{1}{l}{ZoomOut~\cite{melzi2019zoomout}}  & \multicolumn{1}{c}{38.4} & \multicolumn{1}{c}{4.0} & \multicolumn{1}{c}{29.0} \\
    \multicolumn{1}{l}{Smooth Shells~\cite{eisenberger2020smooth}}  & \multicolumn{1}{c}{36.1} & \multicolumn{1}{c}{1.1} & \multicolumn{1}{c}{6.3} \\
    \multicolumn{1}{l}{DiscreteOp~\cite{ren2021discrete}}  & \multicolumn{1}{c}{38.1} & \multicolumn{1}{c}{3.6} & \multicolumn{1}{c}{27.6} \\
    \midrule
    \multicolumn{4}{c}{Supervised Methods} \\ 
    \multicolumn{1}{l}{FMNet~\cite{litany2017deep}}  & \multicolumn{1}{c}{42.0} & \multicolumn{1}{c}{9.6} & \multicolumn{1}{c}{38.0} \\
    \multicolumn{1}{l}{GeomFMaps~\cite{donati2020deep}}  & \multicolumn{1}{c}{8.4} & \multicolumn{1}{c}{2.1} & \multicolumn{1}{c}{{4.1}} \\
    \midrule
    \multicolumn{4}{c}{Unsupervised Methods} \\
    \multicolumn{1}{l}{WSupFMNet~\cite{sharma2020weakly}}  & \multicolumn{1}{c}{7.6} & \multicolumn{1}{c}{3.3} & \multicolumn{1}{c}{22.6} \\
    \multicolumn{1}{l}{Deep Shells~\cite{eisenberger2020deep}}  & \multicolumn{1}{c}{29.3} & \multicolumn{1}{c}{3.4} & \multicolumn{1}{c}{31.1} \\
    \multicolumn{1}{l}{NeuroMorph~\cite{eisenberger2021neuromorph}}  & \multicolumn{1}{c}{5.9} & \multicolumn{1}{c}{14.4} & \multicolumn{1}{c}{{25.3}} \\
    \multicolumn{1}{l}{DUO-FMNet~\cite{donati2022deep}}  & \multicolumn{1}{c}{6.7} & \multicolumn{1}{c}{2.6} & \multicolumn{1}{c}{15.8} \\
    \multicolumn{1}{l}{AttnFMaps~\cite{li2022learning}}  & \multicolumn{1}{c}{5.4} & \multicolumn{1}{c}{1.7} & \multicolumn{1}{c}{11.6} \\
    \multicolumn{1}{l}{URSSM~\cite{cao2023unsupervised}}  & \multicolumn{1}{c}{{3.9}} & \multicolumn{1}{c}{\textbf{0.9}} & \multicolumn{1}{c}{{4.1}} \\
    \multicolumn{1}{l}{Ours}  & \multicolumn{1}{c}{\textbf{1.9}} & \multicolumn{1}{c}{\textbf{0.9}} & \multicolumn{1}{c}{\textbf{3.3}} \\
    \hline
    \end{tabular}
    \caption{\textbf{Non-isometric matching on SMAL and DT4D-H.} Our method substantially outperforms previous state-of-the-art methods even in comparison to supervised methods, especially for non-isometric shape matching (i.e. SMAL and DT4D-H inter-class).}
    \vspace{-0.3cm}
    \label{tab:non-isometry}
\end{table}

\subsection{Shape interpolation}
\noindent \textbf{Datasets.} Following NeuroMorph~\cite{eisenberger2021neuromorph}, we evaluate our method on the
FAUST~\cite{bogo2014faust} and MANO~\cite{romero2017emb} datasets. The MANO dataset contains synthetic hands in various poses with a train/test split of 100/20.

\begin{figure}[ht]
    \centering
    \begin{tabular}{cc}
    \multicolumn{2}{c}{\small{\textbf{FAUST}}}
    \\
     \hspace{-1.2cm}
     \input{figs/tikz/faust_chamfer_pck}&
     \hspace{-1cm}
     \input{figs/tikz/faust_conformal_pck}
     \\
     \multicolumn{2}{c}{\small{\textbf{MANO}}}
     \\
    \hspace{-1.2cm}
     \input{figs/tikz/mano_chamfer_pck}&
     \hspace{-1cm}
     \input{figs/tikz/mano_conformal_pck}
    \end{tabular}
    \caption{{\textbf{Interpolation on FAUST and MANO datasets.} Our method achieves state-of-the-art performance on both datasets, even in comparison to methods that requires ground-truth correspondences (shown as dashed lines).}
    }
    \vspace{-0.35cm}
    \label{fig:interp_pck}
\end{figure}
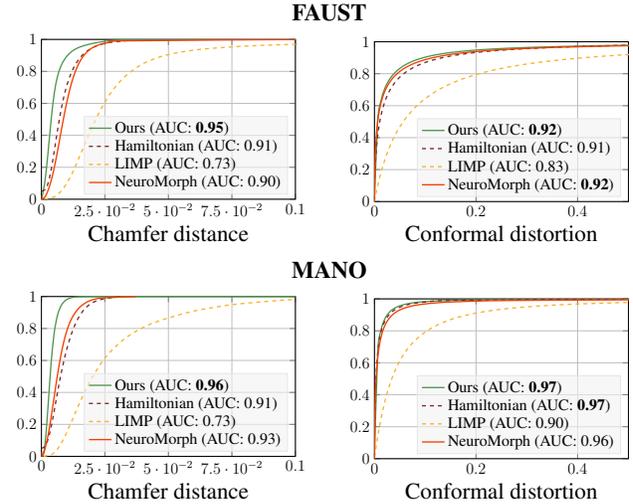

\begin{figure}[!ht]
    \begin{center}  \includegraphics[width=1.0\columnwidth]{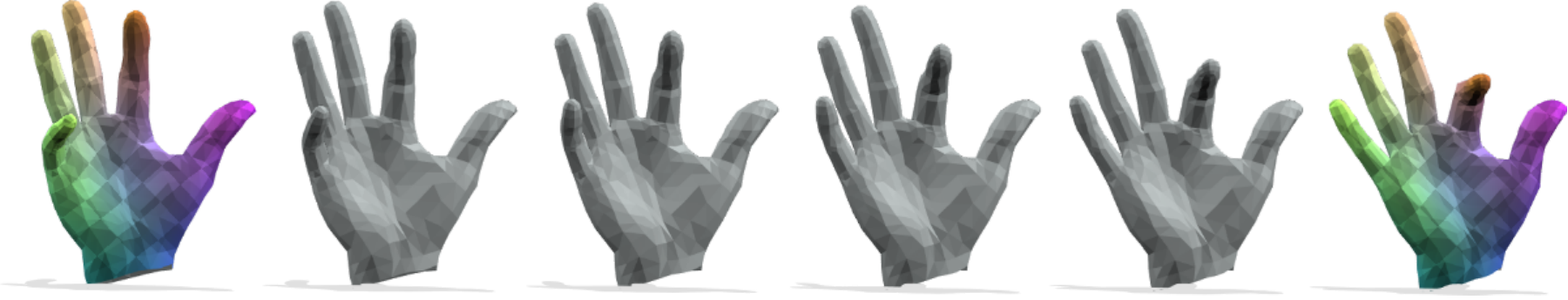}
    \caption{\textbf{Shape interpolation on MANO.} Our method obtains smooth interpolation trajectory between hands in different poses.}
    \vspace{-0.7cm}
    \label{fig:mano_interp}
    \end{center}
\end{figure}

\noindent \textbf{Results.} Two metrics are applied to evaluate the
performance of the shape interpolation. The first conformal distortion metric~\cite{hormann2000mips} measures how much individual triangles of a mesh distort throughout the interpolation sequence, compared to the original shape. The second metric is the Chamfer distance between the target and the final deformed shapes. Besides NeuroMorph, we further evaluate our method against other two baselines (i.e.\ LIMP~\cite{cosmo2020limp} and Hamiltonian~\cite{eisenberger2020hamiltonian}) that require ground-truth correspondences for interpolation.~\cref{fig:interp_pck} summarises the performance of our method against other baselines. Our method achieves state-of-the-art performance on both datasets, even compared to methods relying on ground-truth correspondences.~\cref{fig:mano_interp} shows the interpolation results of our method on the MANO dataset.

\subsection{Statistical shape analysis on medical data}
\label{subsec:medical_data}
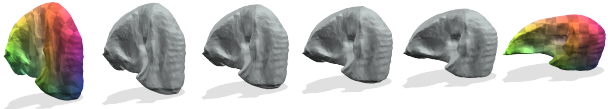
\begin{figure}[ht!]
    \centering
    \input{figs/lung_interpolation}
    \caption{ 
    \textbf{Matching and interpolation on LUNA16 dataset.} Our method obtains accurate correspondences and realistic interpolations between each two lungs despite of large non-isometry.
    }
    \vspace{-0.3cm}
    \label{fig:luna_interp}
\end{figure}

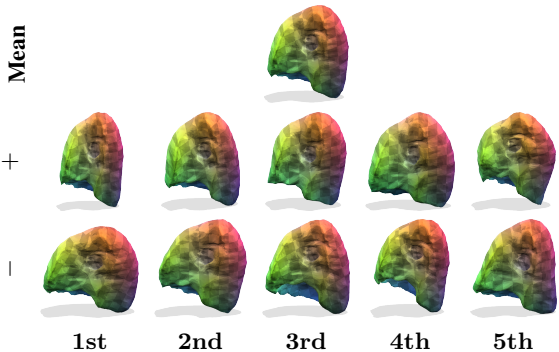
\begin{figure}[ht!]
    \centering
    \input{figs/lung_ssm}
    \caption{ 
    \textbf{Statistical shape analysis on LUNA16 dataset.} The first row visualises the mean shape. While the second row and third row demonstrate the mean shape plus and minus ($\times 2$) the corresponding principal components, respectively.}
    \vspace{-0.3cm}
    \label{fig:luna_ssm}
\end{figure}

To better demonstrate the potential of our method for real-world applications, we conduct statistical shape analysis using medical data.
To this end, we use the real-world LUNA16 dataset~\cite{setio2017validation} that contains chest CT scans with corresponding lung segmentation masks. Based on the provided segmentation masks, we reconstruct 3D lung shapes represented as triangle meshes using Marching Cubes~\cite{lorensen1998marching}. In this way, we get a total of 22 shapes from manual selection. Starting from 22 shapes, our method obtains both accurate point-wise correspondences and smooth interpolation trajectories between them (see~\cref{fig:luna_interp}). Using our obtained correspondences, we build a statistical shape model (SSM) based on a point distribution model (PDM)~\cite{cootes2001statistical}.
The modes of variation of the SSM  are summarised in~\cref{fig:luna_ssm}. We evaluate the quality of the SSM w.r.t.~the two standard metrics: generality and specificity~\cite{davies2002learning} based on the Chamfer distance. We compare our method to the most recent method (i.e.\ S3M~\cite{bastian2023s3m}).~\cref{tab:ssm} summarises the results. Compared to S3M, our method obtains better generalisation and specificity.    

\begin{table}[hbt!]
    \centering
    \small
    \begin{tabular}{@{}lcc@{}}
    \toprule
    \multicolumn{1}{c}{\textbf{Error [mm]}}  & \multicolumn{1}{c}{\textbf{Generalisation} $\downarrow$} & \multicolumn{1}{c}{\textbf{Specificity} $\downarrow$}
    \\ \midrule
    \multicolumn{1}{l}{S3M~\cite{bastian2023s3m}}  & 2.92  & 6.23 \\
    \multicolumn{1}{l}{Ours}  & \textbf{2.44} & \textbf{5.03}  \\ \hline
    \end{tabular} 
    \caption{\textbf{SSM evaluation.} Our method substantially outperforms the state-of-the-art w.r.t.\ generalisation and specificity.}
    \vspace{-4mm}
    \label{tab:ssm}
\end{table}

%% file: figs/tikz/faust_pck.tex
\newcommand{\pckLineWidth}{1pt}
\newcommand{\plotWidth}{\columnwidth}
\newcommand{\plotHeight}{0.7\columnwidth}
\newcommand{\pckTitle}{\textbf{FAUST}}
\definecolor{cPLOT0}{RGB}{28,213,227}
\definecolor{cPLOT1}{RGB}{80,150,80}
\definecolor{cPLOT2}{RGB}{90,130,213}
\definecolor{cPLOT3}{RGB}{247,179,43}
\definecolor{cPLOT4}{RGB}{124,42,43}
\definecolor{cPLOT5}{RGB}{242,64,0}

\pgfplotsset{%
    label style = {font=\LARGE},
    tick label style = {font=\large},
    title style =  {font=\LARGE},
    legend style={  fill= gray!10,
                    fill opacity=0.6, 
                    font=\large,
                    draw=gray!20, %
                    text opacity=1}
}
\begin{tikzpicture}[scale=0.5, transform shape]
	\begin{axis}[
		width=\plotWidth,
		height=\plotHeight,
		grid=major,
		title=\pckTitle,
		legend style={
			at={(0.97,0.03)},
			anchor=south east,
			legend columns=1},
		legend cell align={left},
        xlabel={\LARGE Geodesic error},
		xmin=0,
        xmax=0.1,
        ylabel near ticks,
        xtick={0, 0.025, 0.05, 0.075, 0.1},
	ymin=0,
        ymax=1,
        ytick={0, 0.20, 0.40, 0.60, 0.80, 1.0}
	]

\addplot [color=cPLOT5, smooth, line width=\pckLineWidth]
table[row sep=crcr]{%
0.0 0.25808947368421054 \\
0.005263157894736842 0.26970736842105264 \\
0.010526315789473684 0.34548736842105265 \\
0.015789473684210527 0.4920557894736842 \\
0.021052631578947368 0.6559094736842105 \\
0.02631578947368421 0.7761673684210526 \\
0.031578947368421054 0.8456231578947369 \\
0.03684210526315789 0.8862315789473684 \\
0.042105263157894736 0.9158031578947369 \\
0.04736842105263158 0.9377410526315789 \\
0.05263157894736842 0.9530284210526315 \\
0.05789473684210526 0.9637242105263157 \\
0.06315789473684211 0.9721652631578948 \\
0.06842105263157895 0.9783347368421053 \\
0.07368421052631578 0.98296 \\
0.07894736842105263 0.9864778947368421 \\
0.08421052631578947 0.9891210526315789 \\
0.08947368421052632 0.9914873684210527 \\
0.09473684210526316 0.9931968421052632 \\
0.1 0.9947884210526315 \\
    };
\addlegendentry{\textcolor{black}{AttnFMaps (AUC: {0.82})}}    

\addplot [color=cPLOT4, smooth, line width=\pckLineWidth]
table[row sep=crcr]{%
0.0 0.312412 \\
0.005263157894736842 0.3258105 \\
0.010526315789473684 0.4023205 \\
0.015789473684210527 0.5468495 \\
0.021052631578947368 0.705658 \\
0.02631578947368421 0.8208325 \\
0.031578947368421054 0.8832595 \\
0.03684210526315789 0.921061 \\
0.042105263157894736 0.9465335 \\
0.04736842105263158 0.9657115 \\
0.05263157894736842 0.978877 \\
0.05789473684210526 0.9863015 \\
0.06315789473684211 0.9904765 \\
0.06842105263157895 0.993006 \\
0.07368421052631578 0.994677 \\
0.07894736842105263 0.9958235 \\
0.08421052631578947 0.9966505 \\
0.08947368421052632 0.997221 \\
0.09473684210526316 0.997665 \\
0.1 0.9979685 \\
    };
\addlegendentry{\textcolor{black}{URSSM (AUC: {0.85})}}    

\addplot [color=cPLOT1, smooth, line width=\pckLineWidth]
table[row sep=crcr]{%
0.0  0.3697335 \\
0.002564102564102564  0.3700735 \\
0.005128205128205128  0.384361 \\
0.007692307692307693  0.4204785 \\
0.010256410256410256  0.459689 \\
0.01282051282051282  0.517815 \\
0.015384615384615385  0.5929275 \\
0.017948717948717947  0.6755405 \\
0.020512820512820513  0.7516095 \\
0.023076923076923078  0.813856 \\
0.02564102564102564  0.8644345 \\
0.028205128205128206  0.8981135 \\
0.03076923076923077  0.9218235 \\
0.03333333333333333  0.938483 \\
0.035897435897435895  0.9515735 \\
0.038461538461538464  0.9614475 \\
0.041025641025641026  0.969409 \\
0.04358974358974359  0.9758045 \\
0.046153846153846156  0.981395 \\
0.04871794871794872  0.9858225 \\
0.05128205128205128  0.9889435 \\
0.05384615384615385  0.9913755 \\
0.05641025641025641  0.992925 \\
0.05897435897435897  0.993978 \\
0.06153846153846154  0.994842 \\
0.0641025641025641  0.9954195 \\
0.06666666666666667  0.9958685 \\
0.06923076923076923  0.996193 \\
0.07179487179487179  0.9964825 \\
0.07435897435897436  0.996705 \\
0.07692307692307693  0.9968925 \\
0.07948717948717948  0.9970375 \\
0.08205128205128205  0.9971855 \\
0.08461538461538462  0.99728 \\
0.08717948717948718  0.9973945 \\
0.08974358974358974  0.9974895 \\
0.09230769230769231  0.99756 \\
0.09487179487179487  0.997646 \\
0.09743589743589744  0.9977165 \\
0.1  0.9977885 \\
    };
\addlegendentry{\textcolor{black}{Ours (AUC: \textbf{0.87})}}

\end{axis}
\end{tikzpicture}

%% file: figs/tikz/scape_pck.tex
\newcommand{\pckLineWidth}{1pt}
\newcommand{\plotWidth}{\columnwidth}
\newcommand{\plotHeight}{0.7\columnwidth}
\newcommand{\pckTitle}{\textbf{SCAPE}}
\definecolor{cPLOT0}{RGB}{28,213,227}
\definecolor{cPLOT1}{RGB}{80,150,80}
\definecolor{cPLOT2}{RGB}{90,130,213}
\definecolor{cPLOT3}{RGB}{247,179,43}
\definecolor{cPLOT4}{RGB}{124,42,43}
\definecolor{cPLOT5}{RGB}{242,64,0}

\pgfplotsset{%
    label style = {font=\LARGE},
    tick label style = {font=\large},
    title style =  {font=\LARGE},
    legend style={  fill= gray!10,
                    fill opacity=0.6, 
                    font=\large,
                    draw=gray!20, %
                    text opacity=1}
}
\begin{tikzpicture}[scale=0.5, transform shape]
	\begin{axis}[
		width=\plotWidth,
		height=\plotHeight,
		grid=major,
		title=\pckTitle,
		legend style={
			at={(0.97,0.03)},
			anchor=south east,
			legend columns=1},
		legend cell align={left},
        xlabel={\LARGE Geodesic error},
		xmin=0,
        xmax=0.1,
        ylabel near ticks,
        xtick={0, 0.025, 0.05, 0.075, 0.1},
	ymin=0,
        ymax=1,
        ytick={0, 0.20, 0.40, 0.60, 0.80, 1.0}
	]

\addplot [color=cPLOT5, smooth, line width=\pckLineWidth]
table[row sep=crcr]{%
0.0 0.18994526315789473 \\
0.005263157894736842 0.19249263157894736 \\
0.010526315789473684 0.22945157894736842 \\
0.015789473684210527 0.34391894736842105 \\
0.021052631578947368 0.5259463157894737 \\
0.02631578947368421 0.6822547368421052 \\
0.031578947368421054 0.77422 \\
0.03684210526315789 0.8329821052631579 \\
0.042105263157894736 0.8792747368421052 \\
0.04736842105263158 0.9162526315789473 \\
0.05263157894736842 0.9429273684210526 \\
0.05789473684210526 0.9598747368421052 \\
0.06315789473684211 0.9704663157894737 \\
0.06842105263157895 0.9777010526315789 \\
0.07368421052631578 0.98298 \\
0.07894736842105263 0.9869557894736842 \\
0.08421052631578947 0.9898021052631579 \\
0.08947368421052632 0.9920073684210526 \\
0.09473684210526316 0.9938705263157894 \\
0.1 0.995238947368421 \\
    };
\addlegendentry{\textcolor{black}{AttnFMaps (AUC: {0.78})}}

\addplot [color=cPLOT4, smooth, line width=\pckLineWidth]
table[row sep=crcr]{%
0.0 0.2509255 \\
0.005263157894736842 0.2536675 \\
0.010526315789473684 0.290562 \\
0.015789473684210527 0.404467 \\
0.021052631578947368 0.588076 \\
0.02631578947368421 0.738729 \\
0.031578947368421054 0.8219865 \\
0.03684210526315789 0.87358 \\
0.042105263157894736 0.911547 \\
0.04736842105263158 0.942448 \\
0.05263157894736842 0.964943 \\
0.05789473684210526 0.978441 \\
0.06315789473684211 0.985795 \\
0.06842105263157895 0.9904385 \\
0.07368421052631578 0.993671 \\
0.07894736842105263 0.995846 \\
0.08421052631578947 0.997343 \\
0.08947368421052632 0.9983675 \\
0.09473684210526316 0.999008 \\
0.1 0.9994695 \\
    };
\addlegendentry{\textcolor{black}{URSSM (AUC: {0.81})}}

\addplot [color=cPLOT1, smooth, line width=\pckLineWidth]
table[row sep=crcr]{%
0.0  0.2731925 \\
0.002564102564102564  0.273208 \\
0.005128205128205128  0.276346 \\
0.007692307692307693  0.2904325 \\
0.010256410256410256  0.3152935 \\
0.01282051282051282  0.3563665 \\
0.015384615384615385  0.4221145 \\
0.017948717948717947  0.5089775 \\
0.020512820512820513  0.5979155 \\
0.023076923076923078  0.680033 \\
0.02564102564102564  0.747472 \\
0.028205128205128206  0.796993 \\
0.03076923076923077  0.831469 \\
0.03333333333333333  0.858105 \\
0.035897435897435895  0.879837 \\
0.038461538461538464  0.898152 \\
0.041025641025641026  0.9138255 \\
0.04358974358974359  0.928218 \\
0.046153846153846156  0.941629 \\
0.04871794871794872  0.953267 \\
0.05128205128205128  0.9628985 \\
0.05384615384615385  0.970467 \\
0.05641025641025641  0.9760465 \\
0.05897435897435897  0.9801995 \\
0.06153846153846154  0.983204 \\
0.0641025641025641  0.9857025 \\
0.06666666666666667  0.987673 \\
0.06923076923076923  0.9893865 \\
0.07179487179487179  0.9908635 \\
0.07435897435897436  0.9920265 \\
0.07692307692307693  0.9930515 \\
0.07948717948717948  0.993885 \\
0.08205128205128205  0.99461 \\
0.08461538461538462  0.995209 \\
0.08717948717948718  0.995746 \\
0.08974358974358974  0.996196 \\
0.09230769230769231  0.996597 \\
0.09487179487179487  0.9969595 \\
0.09743589743589744  0.997263 \\
0.1  0.9975295 \\
    };
\addlegendentry{\textcolor{black}{Ours (AUC: \textbf{0.82})}}    

\end{axis}
\end{tikzpicture}

%% file: figs/tikz/smal_pck.tex
\newcommand{\pckLineWidth}{1pt}
\newcommand{\plotWidth}{\columnwidth}
\newcommand{\plotHeight}{0.7\columnwidth}
\newcommand{\pckTitle}{\textbf{SMAL}}
\definecolor{cPLOT0}{RGB}{28,213,227}
\definecolor{cPLOT1}{RGB}{80,150,80}
\definecolor{cPLOT2}{RGB}{90,130,213}
\definecolor{cPLOT3}{RGB}{247,179,43}
\definecolor{cPLOT4}{RGB}{124,42,43}
\definecolor{cPLOT5}{RGB}{242,64,0}

\pgfplotsset{%
    label style = {font=\LARGE},
    tick label style = {font=\large},
    title style =  {font=\LARGE},
    legend style={  fill= gray!10,
                    fill opacity=0.6, 
                    font=\large,
                    draw=gray!20, %
                    text opacity=1}
}
\begin{tikzpicture}[scale=0.5, transform shape]
	\begin{axis}[
		width=\plotWidth,
		height=\plotHeight,
		grid=major,
		title=\pckTitle,
		legend style={
			at={(0.97,0.03)},
			anchor=south east,
			legend columns=1},
		legend cell align={left},
        xlabel={\LARGE Geodesic error},
		xmin=0,
        xmax=0.2,
        ylabel near ticks,
        xtick={0, 0.05, 0.1, 0.15, 0.2},
	ymin=0,
        ymax=1,
        ytick={0, 0.20, 0.40, 0.60, 0.80, 1.0}
	]

\addplot [color=cPLOT5, smooth, line width=\pckLineWidth]
table[row sep=crcr]{%
0.0 0.06488340934619913 \\
0.010526315789473684 0.13704104694753083 \\
0.021052631578947368 0.3482007280994979 \\
0.031578947368421054 0.5091134238269884 \\
0.042105263157894736 0.6313163984788405 \\
0.05263157894736842 0.7207291821737424 \\
0.06315789473684211 0.788068912316791 \\
0.07368421052631578 0.8386001001475146 \\
0.08421052631578947 0.877629210593983 \\
0.09473684210526316 0.9058721630509805 \\
0.10526315789473684 0.9262048152007687 \\
0.11578947368421053 0.9395975152589625 \\
0.12631578947368421 0.9487271792234507 \\
0.1368421052631579 0.9551447402254672 \\
0.14736842105263157 0.9603266974327049 \\
0.15789473684210525 0.9645085328389114 \\
0.16842105263157894 0.9678350543367934 \\
0.17894736842105263 0.9708245929815539 \\
0.18947368421052632 0.9734906822211095 \\
0.2 0.9757561814023359 \\
    };
\addlegendentry{\textcolor{black}{AttnFMaps (AUC: 0.78)}}

\addplot [color=cPLOT4, smooth, line width=\pckLineWidth]
table[row sep=crcr]{%
0.0 0.11681666238107483 \\
0.010526315789473684 0.2009649010028285 \\
0.021052631578947368 0.4125514271020828 \\
0.031578947368421054 0.5778947030084854 \\
0.042105263157894736 0.6992298791463101 \\
0.05263157894736842 0.7843635896117254 \\
0.06315789473684211 0.8472923630753407 \\
0.07368421052631578 0.893088840318848 \\
0.08421052631578947 0.9245872975057855 \\
0.09473684210526316 0.944853432759064 \\
0.10526315789473684 0.9571213679609154 \\
0.11578947368421053 0.9645358704037027 \\
0.12631578947368421 0.9692600925687838 \\
0.1368421052631579 0.9726851375674981 \\
0.14736842105263157 0.9756203394188737 \\
0.15789473684210525 0.9781679094883003 \\
0.16842105263157894 0.9803272049370018 \\
0.17894736842105263 0.9822512213936745 \\
0.18947368421052632 0.9838518899460016 \\
0.2 0.98525777834919 \\
    };
\addlegendentry{\textcolor{black}{URSSM (AUC: {0.82})}}

\addplot [color=cPLOT1, smooth, line width=\pckLineWidth]
table[row sep=crcr]{%
0.0  0.27914631010542557 \\
0.005128205128205128  0.350073926459244 \\
0.010256410256410256  0.5110452558498328 \\
0.015384615384615385  0.6512342504499872 \\
0.020512820512820513  0.7535754692723065 \\
0.02564102564102564  0.8234244021599383 \\
0.03076923076923077  0.8702121367960916 \\
0.035897435897435895  0.9010613268192338 \\
0.041025641025641026  0.9220480843404474 \\
0.046153846153846156  0.9376536384674724 \\
0.05128205128205128  0.9493147338647467 \\
0.05641025641025641  0.9579339161738236 \\
0.06153846153846154  0.9648470043713037 \\
0.06666666666666667  0.970458986886089 \\
0.07179487179487179  0.9750430701979943 \\
0.07692307692307693  0.978640395988686 \\
0.08205128205128205  0.9815691694523013 \\
0.08717948717948718  0.9839856004114168 \\
0.09230769230769231  0.9857771920802263 \\
0.09743589743589744  0.9872717922345076 \\
0.10256410256410256  0.9885028284906145 \\
0.1076923076923077  0.9894632296220108 \\
0.11282051282051282  0.9903207765492414 \\
0.11794871794871795  0.9910131139110311 \\
0.12307692307692308  0.991636024685009 \\
0.1282051282051282  0.9922486500385703 \\
0.13333333333333333  0.9928002057084083 \\
0.13846153846153847  0.9933003342761635 \\
0.14358974358974358  0.9937676780663409 \\
0.14871794871794872  0.9942356646952945 \\
0.15384615384615385  0.9945924402159938 \\
0.15897435897435896  0.9949498585754692 \\
0.1641025641025641  0.9952609925430702 \\
0.16923076923076924  0.995514271020828 \\
0.17435897435897435  0.9957302648495757 \\
0.1794871794871795  0.9959552584211879 \\
0.18461538461538463  0.9961326819233736 \\
0.18974358974358974  0.996304319876575 \\
0.19487179487179487  0.9964695294420159 \\
0.2  0.9966302391360247 \\
    };
\addlegendentry{\textcolor{black}{Ours (AUC: \textbf{0.92})}}  

\end{axis}
\end{tikzpicture}

%% file: figs/tikz/dt4d_inter_pck.tex
\newcommand{\pckLineWidth}{1pt}
\newcommand{\plotWidth}{\columnwidth}
\newcommand{\plotHeight}{0.7\columnwidth}
\newcommand{\pckTitle}{\textbf{DT4D-H inter-class}}
\definecolor{cPLOT0}{RGB}{28,213,227}
\definecolor{cPLOT1}{RGB}{80,150,80}
\definecolor{cPLOT2}{RGB}{90,130,213}
\definecolor{cPLOT3}{RGB}{247,179,43}
\definecolor{cPLOT4}{RGB}{124,42,43}
\definecolor{cPLOT5}{RGB}{242,64,0}

\pgfplotsset{%
    label style = {font=\LARGE},
    tick label style = {font=\large},
    title style =  {font=\LARGE},
    legend style={  fill= gray!10,
                    fill opacity=0.6, 
                    font=\large,
                    draw=gray!20, %
                    text opacity=1}
}
\begin{tikzpicture}[scale=0.5, transform shape]
	\begin{axis}[
		width=\plotWidth,
		height=\plotHeight,
		grid=major,
		title=\pckTitle,
		legend style={
			at={(0.97,0.03)},
			anchor=south east,
			legend columns=1},
		legend cell align={left},
        xlabel={\LARGE Geodesic error},
		xmin=0,
        xmax=0.2,
        ylabel near ticks,
        xtick={0, 0.05, 0.1, 0.15, 0.2},
	ymin=0,
        ymax=1,
        ytick={0, 0.20, 0.40, 0.60, 0.80, 1.0}
	]

\addplot [color=cPLOT5, smooth, line width=\pckLineWidth]
table[row sep=crcr]{%
0.0 0.03456135231465077 \\
0.010526315789473684 0.08081747504012339 \\
0.021052631578947368 0.26366498530545884 \\
0.031578947368421054 0.4222855743377035 \\
0.042105263157894736 0.5487909831794402 \\
0.05263157894736842 0.6365265648123059 \\
0.06315789473684211 0.6959819288409029 \\
0.07368421052631578 0.7381741251016112 \\
0.08421052631578947 0.7694500489818038 \\
0.09473684210526316 0.7939294870458762 \\
0.10526315789473684 0.8120624465889905 \\
0.11578947368421053 0.8260296600454384 \\
0.12631578947368421 0.8372674406486441 \\
0.1368421052631579 0.8469622944327491 \\
0.14736842105263157 0.8553124413781603 \\
0.15789473684210525 0.8624852533505638 \\
0.16842105263157894 0.8689187527356859 \\
0.17894736842105263 0.8752861787940055 \\
0.18947368421052632 0.8819866811180357 \\
0.2 0.8887820205515142 \\
    };
\addlegendentry{\textcolor{black}{AttnFMaps (AUC: 0.68)}}

\addplot [color=cPLOT4, smooth, line width=\pckLineWidth]
table[row sep=crcr]{%
0.0 0.050371427975905124 \\
0.010526315789473684 0.12944817308293557 \\
0.021052631578947368 0.3864891510515455 \\
0.031578947368421054 0.5803202576234445 \\
0.042105263157894736 0.7183634866706964 \\
0.05263157894736842 0.8076292806970007 \\
0.06315789473684211 0.8629943514600746 \\
0.07368421052631578 0.897993205077433 \\
0.08421052631578947 0.9198287721199742 \\
0.09473684210526316 0.9337159472247119 \\
0.10526315789473684 0.942982262334035 \\
0.11578947368421053 0.9494709965191654 \\
0.12631578947368421 0.9543339516851825 \\
0.1368421052631579 0.95824395022615 \\
0.14736842105263157 0.9615212497655127 \\
0.15789473684210525 0.9646511870271172 \\
0.16842105263157894 0.9679835129332806 \\
0.17894736842105263 0.9716687162598745 \\
0.18947368421052632 0.9755318173291369 \\
0.2 0.9789688809221085 \\
    };
\addlegendentry{\textcolor{black}{URSMM (AUC: {0.81})}}  

\addplot [color=cPLOT1, smooth, line width=\pckLineWidth]
table[row sep=crcr]{%
0.0  0.0606067490672614 \\
0.005128205128205128  0.0705998707714113 \\
0.010256410256410256  0.15187006273839548 \\
0.015384615384615385  0.293468537007316 \\
0.020512820512820513  0.43549325718573484 \\
0.02564102564102564  0.5486407028367759 \\
0.03076923076923077  0.6389126039560623 \\
0.035897435897435895  0.7125051587218876 \\
0.041025641025641026  0.7709200241782521 \\
0.046153846153846156  0.8160542551639327 \\
0.05128205128205128  0.8499745711486754 \\
0.05641025641025641  0.8759373241344811 \\
0.06153846153846154  0.8967822081414011 \\
0.06666666666666667  0.9137993205077433 \\
0.07179487179487179  0.9273997957354566 \\
0.07692307692307693  0.9383027283906872 \\
0.08205128205128205  0.9474153240094212 \\
0.08717948717948718  0.9548898430497947 \\
0.09230769230769231  0.961058528044688 \\
0.09743589743589744  0.9660736394522376 \\
0.10256410256410256  0.9700842070158617 \\
0.1076923076923077  0.9732954332284219 \\
0.11282051282051282  0.9758325864476728 \\
0.11794871794871795  0.9778802342789253 \\
0.12307692307692308  0.9796044979886196 \\
0.1282051282051282  0.9809855764220355 \\
0.13333333333333333  0.9821223711361694 \\
0.13846153846153847  0.9830418742314025 \\
0.14358974358974358  0.983758884465473 \\
0.14871794871794872  0.9843996081455697 \\
0.15384615384615385  0.9850262625841549 \\
0.15897435897435896  0.9856060195510349 \\
0.1641025641025641  0.9862023469579173 \\
0.16923076923076924  0.9867311420055443 \\
0.17435897435897435  0.987226483523355 \\
0.1794871794871795  0.9876773245513475 \\
0.18461538461538463  0.98811138670613 \\
0.18974358974358974  0.9885532651061967 \\
0.19487179487179487  0.988999312170415 \\
0.2  0.9893872063697188 \\
    };
\addlegendentry{\textcolor{black}{Ours (AUC: \textbf{0.84})}}  
\end{axis}
\end{tikzpicture}

%% file: figs/our_matching_results.tex
\def\heightIQ{2.0cm}
\def\widthIQ{1.6cm}
\def\hspaceColsIQ{-0.3cm}
\def\firstRow{Strafing009-StandingReactLargeFromLeft016}
\def\secondRow{shape_030-shape_034}
\begin{tabular}{lccccc}%
    \setlength{\tabcolsep}{0pt} 
    \rotatedCentering{90}{\heightIQ}{\textbf{\small{DT4D-H}}}&
    \hspace{\hspaceColsIQ}
    \includegraphics[height=\heightIQ, width=\widthIQ]{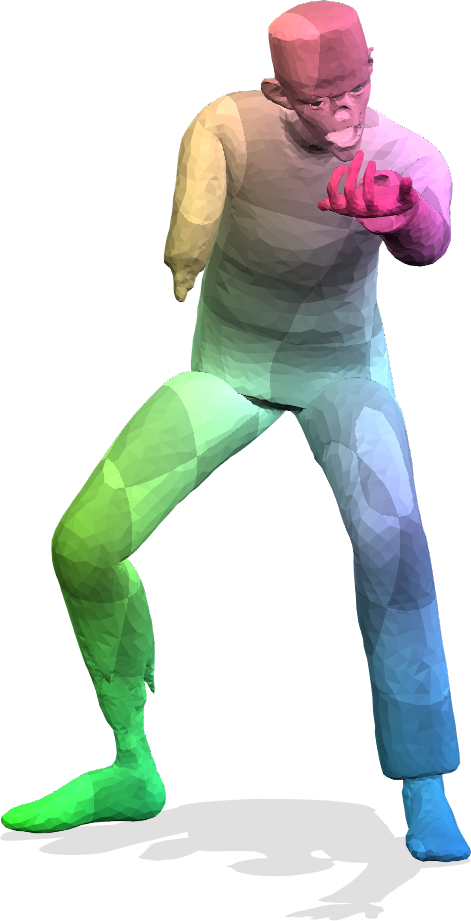}&
    \hspace{\hspaceColsIQ}
    \begin{overpic}[height=\heightIQ, width=\widthIQ]{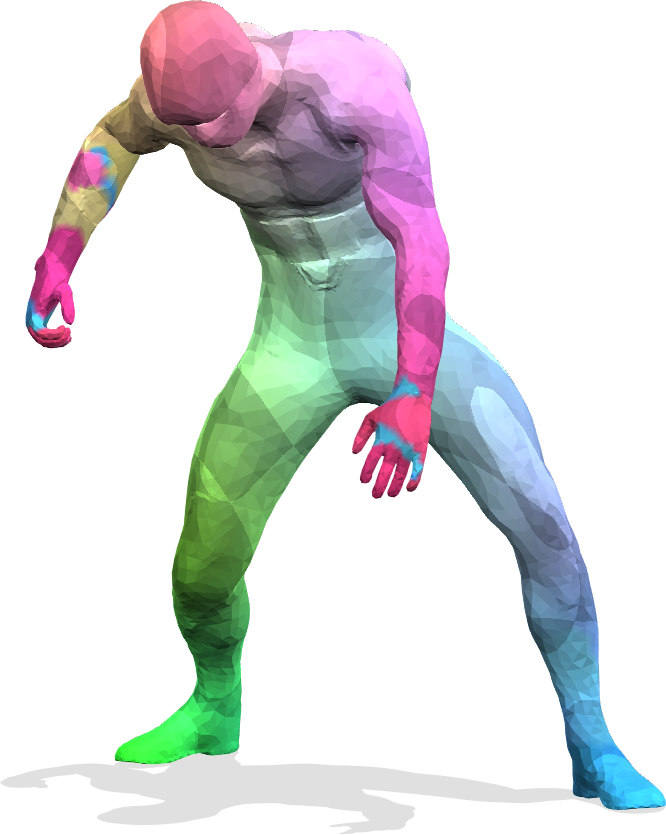}
        \put(-5,55){\includegraphics[height=0.5cm]{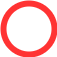}}
        \put(36,35){\includegraphics[height=0.5cm]{figs/red_circle.png}}
    \end{overpic}
    &
    \hspace{\hspaceColsIQ}
    \begin{overpic}[height=\heightIQ, width=\widthIQ]{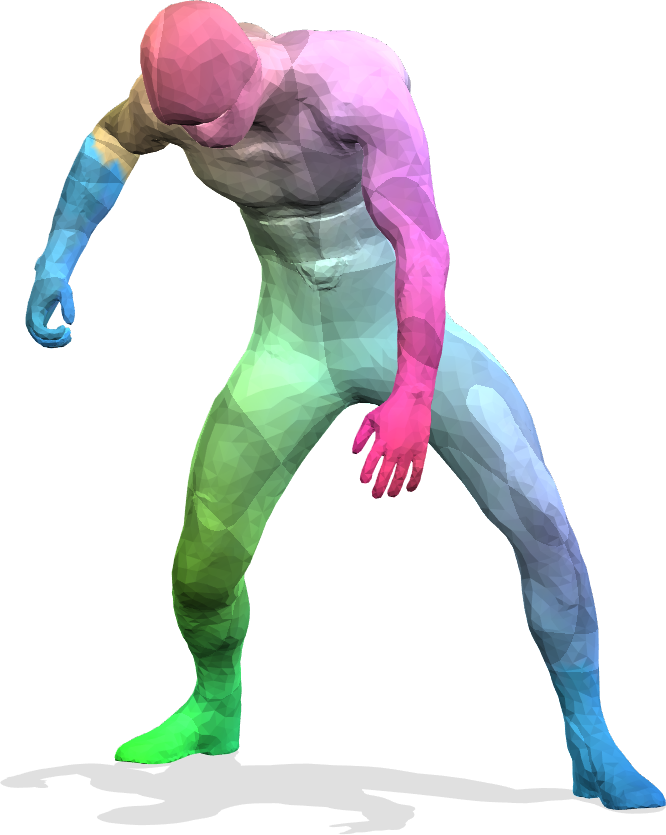}
        \put(-5,55){\includegraphics[height=0.5cm]{figs/red_circle.png}}
    \end{overpic}
    &
    \hspace{\hspaceColsIQ}
    \begin{overpic}[height=\heightIQ, width=\widthIQ]{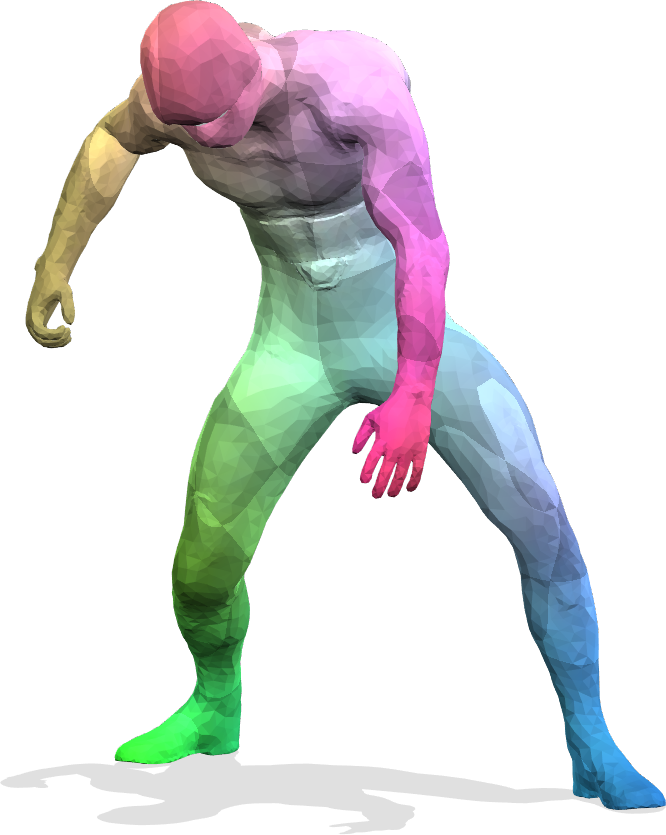}
        \put(-5,55){\includegraphics[height=0.5cm]{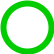}}
    \end{overpic} \\

    \setlength{\tabcolsep}{0pt} 
    \rotatedCentering{90}{\heightIQ}{\textbf{\small{SMAL}}}&
    \hspace{\hspaceColsIQ}
    \includegraphics[height=\heightIQ, width=\widthIQ]{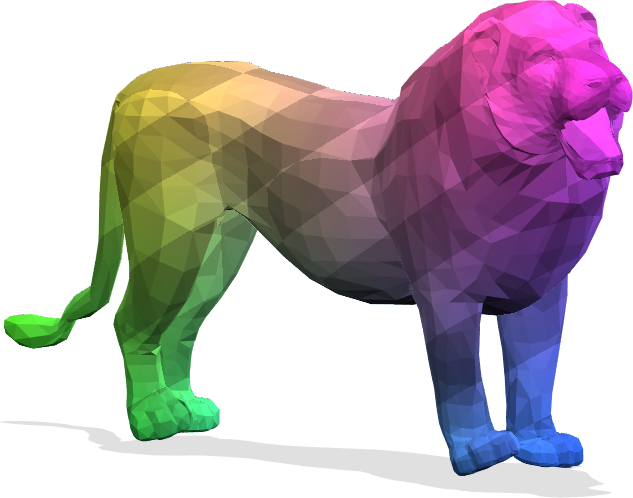}&
    \hspace{\hspaceColsIQ}
    \begin{overpic}[height=\heightIQ, width=\widthIQ]{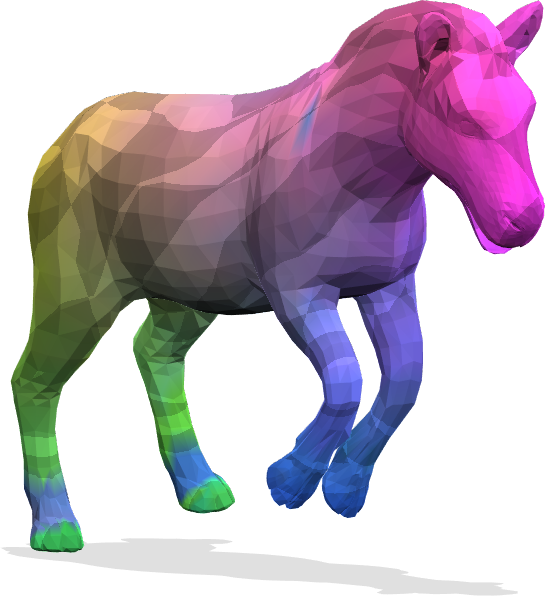}
        \put(-6,2){\includegraphics[height=0.5cm]{figs/red_circle.png}}
        \put(36,62){\includegraphics[height=0.5cm]{figs/red_circle.png}}
    \end{overpic}&
    \hspace{\hspaceColsIQ}
    \begin{overpic}[height=\heightIQ, width=\widthIQ]{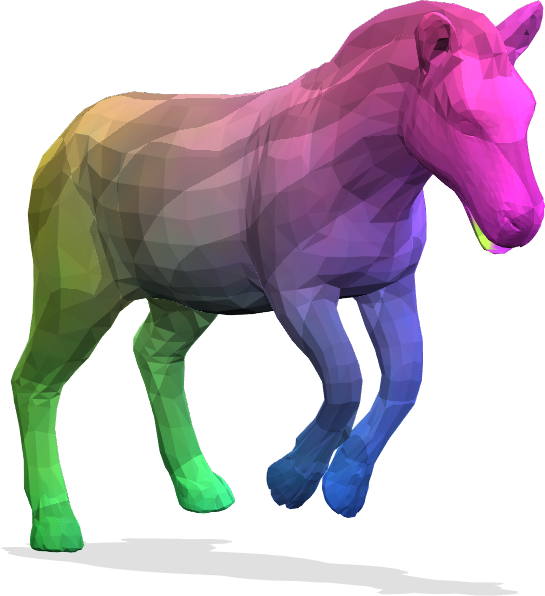}
        \put(70,45){\includegraphics[height=0.5cm]{figs/red_circle.png}}
    \end{overpic}&
    \hspace{\hspaceColsIQ}
    \begin{overpic}[height=\heightIQ, width=\widthIQ]{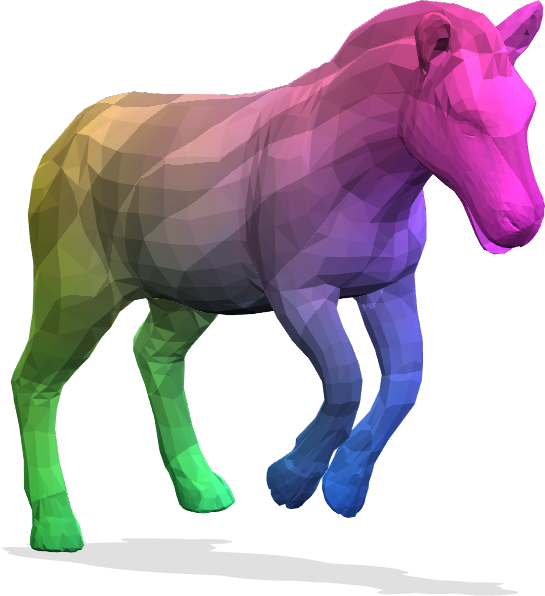}
        \put(70,45){\includegraphics[height=0.5cm]{figs/green_circle.png}}
    \end{overpic} \\

    & \textbf{\small{Source}} & \textbf{\small{AttnFMaps}} & \textbf{\small{URSSM}} & \textbf{\small{Ours}}
    \\
\end{tabular}

%% file: figs/tikz/faust_chamfer_pck.tex
\newcommand{\pckLineWidth}{1pt}
\newcommand{\plotWidth}{\columnwidth}
\newcommand{\plotHeight}{0.7\columnwidth}
\newcommand{\pckTitle}{\textbf{FAUST}}
\definecolor{cPLOT0}{RGB}{28,213,227}
\definecolor{cPLOT1}{RGB}{80,150,80}
\definecolor{cPLOT2}{RGB}{90,130,213}
\definecolor{cPLOT3}{RGB}{247,179,43}
\definecolor{cPLOT4}{RGB}{124,42,43}
\definecolor{cPLOT5}{RGB}{242,64,0}

\pgfplotsset{%
    label style = {font=\LARGE},
    tick label style = {font=\large},
    title style =  {font=\large},
    legend style={  fill= gray!10,
                    fill opacity=0.6, 
                    font=\large,
                    draw=gray!20, %
                    text opacity=1}
}
\begin{tikzpicture}[scale=0.5, transform shape]
	\begin{axis}[
		width=\plotWidth,
		height=\plotHeight,
		grid=major,
		legend style={
			at={(0.97,0.03)},
			anchor=south east,
			legend columns=1},
		legend cell align={left},
        xlabel={\LARGE Chamfer distance},
		xmin=0,
        xmax=0.1,
        ylabel near ticks,
        xtick={0, 0.025, 0.05, 0.075, 0.1},
	ymin=0,
        ymax=1,
        ytick={0, 0.20, 0.40, 0.60, 0.80, 1.0}
	]

\addplot [color=cPLOT1, smooth, line width=\pckLineWidth]
table[row sep=crcr]{%
0.0  0.0 \\
0.0012658227848101266  0.09748370130389569 \\
0.002531645569620253  0.2798826093912487 \\
0.00379746835443038  0.49249310055195583 \\
0.005063291139240506  0.663889388848892 \\
0.006329113924050633  0.7683405327573795 \\
0.00759493670886076  0.8298246140308775 \\
0.008860759493670886  0.8689859811215103 \\
0.010126582278481013  0.8969807415406768 \\
0.01139240506329114  0.9173776097912167 \\
0.012658227848101266  0.9326858851291897 \\
0.013924050632911392  0.9438849892008639 \\
0.01518987341772152  0.9524818014558836 \\
0.016455696202531647  0.9594602431805456 \\
0.017721518987341773  0.9653502719782417 \\
0.0189873417721519  0.9704458643308536 \\
0.020253164556962026  0.9750134989200864 \\
0.021518987341772152  0.9790756739460843 \\
0.02278481012658228  0.9827593792496601 \\
0.024050632911392405  0.9858166346692264 \\
0.02531645569620253  0.9882904367650588 \\
0.026582278481012658  0.9903807695384369 \\
0.027848101265822784  0.9923371130309575 \\
0.02911392405063291  0.9942824574034077 \\
0.03037974683544304  0.9957458403327734 \\
0.03164556962025317  0.9967287616990641 \\
0.03291139240506329  0.9975271978241741 \\
0.03417721518987342  0.9981476481881449 \\
0.035443037974683546  0.9987850971922246 \\
0.03670886075949367  0.9991135709143268 \\
0.0379746835443038  0.9994295456363491 \\
0.039240506329113925  0.9995880329573634 \\
0.04050632911392405  0.9997375209983201 \\
0.04177215189873418  0.9997895168386529 \\
0.043037974683544304  0.9998360131189504 \\
0.04430379746835443  0.9998765098792096 \\
0.04556962025316456  0.99991250699944 \\
0.04683544303797468  0.9999905007599392 \\
0.04810126582278481  1.0 \\
0.049367088607594936  1.0 \\
0.05063291139240506  1.0 \\
0.05189873417721519  1.0 \\
0.053164556962025315  1.0 \\
0.05443037974683544  1.0 \\
0.05569620253164557  1.0 \\
0.056962025316455694  1.0 \\
0.05822784810126582  1.0 \\
0.05949367088607595  1.0 \\
0.06075949367088608  1.0 \\
0.06202531645569621  1.0 \\
0.06329113924050633  1.0 \\
0.06455696202531645  1.0 \\
0.06582278481012659  1.0 \\
0.0670886075949367  1.0 \\
0.06835443037974684  1.0 \\
0.06962025316455696  1.0 \\
0.07088607594936709  1.0 \\
0.07215189873417721  1.0 \\
0.07341772151898734  1.0 \\
0.07468354430379746  1.0 \\
0.0759493670886076  1.0 \\
0.07721518987341772  1.0 \\
0.07848101265822785  1.0 \\
0.07974683544303797  1.0 \\
0.0810126582278481  1.0 \\
0.08227848101265824  1.0 \\
0.08354430379746836  1.0 \\
0.08481012658227849  1.0 \\
0.08607594936708861  1.0 \\
0.08734177215189874  1.0 \\
0.08860759493670886  1.0 \\
0.089873417721519  1.0 \\
0.09113924050632911  1.0 \\
0.09240506329113925  1.0 \\
0.09367088607594937  1.0 \\
0.0949367088607595  1.0 \\
0.09620253164556962  1.0 \\
0.09746835443037975  1.0 \\
0.09873417721518987  1.0 \\
0.1  1.0 \\
    };
\addlegendentry{\textcolor{black}{Ours (AUC: \textbf{0.95})}}  

\addplot [color=cPLOT4, smooth, dashed, line width=\pckLineWidth]
table[row sep=crcr]{%
0.0  0.0 \\
0.0  0.010010726203245479 \\
0.0  0.020020261928883246 \\
0.0  0.03003098813212873 \\
0.0  0.04004052385776649 \\
0.0002477052733368539  0.05005005958340426 \\
0.0014799347472772334  0.060060785786649744 \\
0.001884615042575857  0.07007032151228752 \\
0.0021737316310315364  0.08008104771553298 \\
0.0024140727168728343  0.09009058344117075 \\
0.0026159630238047084  0.10010011916680853 \\
0.002795220262207872  0.110110845370054 \\
0.0029606184544238375  0.12012038109569177 \\
0.003114710188177397  0.13013110729893726 \\
0.0032607149994709453  0.14014064302457505 \\
0.003400009210097229  0.1501501787502128 \\
0.003530523204748936  0.16016090495345828 \\
0.003655420008048448  0.17017044067909606 \\
0.003774689777081115  0.1801811668823415 \\
0.003891624795938346  0.1901907026079793 \\
0.004005351278168124  0.20020023833361705 \\
0.004116699043841574  0.21021096453686255 \\
0.004226195108495239  0.2202205002625003 \\
0.004331730119200906  0.2302312264657458 \\
0.004439588070585294  0.24024076219138354 \\
0.004543088031681464  0.2502502979170213 \\
0.004645726636780129  0.2602610241202668 \\
0.004746507906169984  0.2702705598459046 \\
0.004848069855100197  0.2802812860491501 \\
0.0049473868975384245  0.2902908217747878 \\
0.00504545150194757  0.3003003575004256 \\
0.005143847626416395  0.31031108370367105 \\
0.0052417850311138004  0.3203206194293089 \\
0.005338097730197052  0.3303313456325543 \\
0.005436428470343091  0.3403408813581921 \\
0.00553407363540724  0.35035041708382986 \\
0.005631567630728537  0.36036114328707536 \\
0.005728130228822547  0.37037067901271314 \\
0.0058277701372560185  0.3803814052159586 \\
0.005925767770852153  0.3903909409415964 \\
0.006026158110828497  0.4004004766672341 \\
0.006125450364893862  0.4104112028704796 \\
0.006227398542049307  0.4204207385961174 \\
0.006328073439540923  0.43043146479936284 \\
0.006430094519102327  0.4404410005250006 \\
0.006533396700561118  0.45045053625063836 \\
0.0066396403858936  0.46046126245388386 \\
0.006742536257574515  0.47047079817952164 \\
0.0068475555065195415  0.4804815243827671 \\
0.006955797064092762  0.4904910601084049 \\
0.007064601164201066  0.5005005958340426 \\
0.007176470731065376  0.5105113220372881 \\
0.0072906592794186355  0.5205208577629259 \\
0.0074054382322739426  0.5305315839661714 \\
0.007521389235398599  0.5405411196918092 \\
0.00764229335066913  0.5505506554174469 \\
0.007762363614257336  0.5605613816206924 \\
0.007886962460414787  0.5705709173463301 \\
0.008011654552471513  0.5805816435495756 \\
0.00813995695944579  0.5905911792752134 \\
0.008272906779545867  0.6006007150008512 \\
0.008405007124854997  0.6106114412040966 \\
0.008541469344409297  0.6206209769297345 \\
0.008681452734485647  0.63063170313298 \\
0.008828222844626319  0.6406412388586178 \\
0.008979716839148309  0.6506507745842555 \\
0.00913649569200244  0.660661500787501 \\
0.009295534255438326  0.6706710365131388 \\
0.009460404467159907  0.6806817627163843 \\
0.009631448124361947  0.6906912984420219 \\
0.0098049327772737  0.7007008341676597 \\
0.009988571925494226  0.7107115603709052 \\
0.010180375917247004  0.7207210960965429 \\
0.010379311302803403  0.7307318222997884 \\
0.010584164472927196  0.7407413580254263 \\
0.010796125079460784  0.750750893751064 \\
0.011018636412920283  0.7607616199543095 \\
0.011251738926407353  0.7707711556799473 \\
0.01149983946295262  0.7807818818831928 \\
0.011760096563168012  0.7907914176088304 \\
0.012030636266638722  0.8008009533344682 \\
0.012324222194167109  0.8108116795377137 \\
0.012627789676544753  0.8208212152633514 \\
0.01295108736115272  0.8308319414665969 \\
0.013300314567360881  0.8408414771922348 \\
0.013670450877425995  0.8508510129178726 \\
0.014079783232366182  0.8608617391211181 \\
0.014513400340832268  0.8708712748467557 \\
0.014996040329641198  0.8808820010500013 \\
0.01551246799689365  0.890891536775639 \\
0.01609596432975187  0.9009010725012767 \\
0.01674515949265104  0.9109117987045222 \\
0.017482466271730737  0.9209213344301601 \\
0.018327189277174798  0.9309320606334055 \\
0.019305860487063307  0.9409415963590433 \\
0.02048422077263801  0.9509511320846811 \\
0.021966734787698356  0.9609618582879266 \\
0.0239557102894321  0.9709713940135642 \\
0.026876879586014268  0.9809821202168097 \\
0.033017364753337046  0.9909916559424475 \\
    };
\addlegendentry{\textcolor{black}{Hamiltonian (AUC: {0.91})}}    

\addplot [color=cPLOT3, smooth, dashed, line width=\pckLineWidth]
table[row sep=crcr]{%
0.001025913450089836  0.0 \\
0.004416869672846353  0.010010726203245479 \\
0.0055105640466302916  0.020020261928883246 \\
0.0063964869901485585  0.03003098813212873 \\
0.007079190537010273  0.04004052385776649 \\
0.007696820098385925  0.05005005958340426 \\
0.00820174185160551  0.060060785786649744 \\
0.008704555963049027  0.07007032151228752 \\
0.00916634113488486  0.08008104771553298 \\
0.009590220345668027  0.09009058344117075 \\
0.009951950643762166  0.10010011916680853 \\
0.010340299324553513  0.110110845370054 \\
0.010716258310768641  0.12012038109569177 \\
0.011081813207382931  0.13013110729893726 \\
0.011399792999564512  0.14014064302457505 \\
0.011728504056096262  0.1501501787502128 \\
0.012047607969760876  0.16016090495345828 \\
0.012347707853470441  0.17017044067909606 \\
0.01264576712486717  0.1801811668823415 \\
0.012921844840748173  0.1901907026079793 \\
0.013200714901438346  0.20020023833361705 \\
0.013462678571288671  0.21021096453686255 \\
0.013737252259671056  0.2202205002625003 \\
0.014007479753301483  0.2302312264657458 \\
0.014309427626786799  0.24024076219138354 \\
0.01456680033835703  0.2502502979170213 \\
0.014831048862063684  0.2602610241202668 \\
0.015088381694411547  0.2702705598459046 \\
0.015371505290423988  0.2802812860491501 \\
0.015646380641387556  0.2902908217747878 \\
0.015909846275315512  0.3003003575004256 \\
0.016169713990546238  0.31031108370367105 \\
0.01642138513955509  0.3203206194293089 \\
0.016685567655044326  0.3303313456325543 \\
0.01694496276824488  0.3403408813581921 \\
0.017222496538122196  0.35035041708382986 \\
0.017479950656861552  0.36036114328707536 \\
0.017746291677923774  0.37037067901271314 \\
0.018008619095287098  0.3803814052159586 \\
0.018281970044503233  0.3903909409415964 \\
0.01853341076440009  0.4004004766672341 \\
0.018808230867821258  0.4104112028704796 \\
0.01908423875730445  0.4204207385961174 \\
0.019359709176250958  0.43043146479936284 \\
0.01963287051883784  0.4404410005250006 \\
0.01991936907709576  0.45045053625063836 \\
0.020231769311869274  0.46046126245388386 \\
0.020520653535170096  0.47047079817952164 \\
0.020822449668073173  0.4804815243827671 \\
0.021117000198903675  0.4904910601084049 \\
0.021417706761071586  0.5005005958340426 \\
0.02173680173266776  0.5105113220372881 \\
0.02203701781818839  0.5205208577629259 \\
0.022349893326231836  0.5305315839661714 \\
0.02265651103034529  0.5405411196918092 \\
0.022976905190765114  0.5505506554174469 \\
0.02332056487269663  0.5605613816206924 \\
0.02368091319786037  0.5705709173463301 \\
0.024052926993726145  0.5805816435495756 \\
0.024425480835496605  0.5905911792752134 \\
0.024782216808532008  0.6006007150008512 \\
0.025145418528374593  0.6106114412040966 \\
0.025499977435441773  0.6206209769297345 \\
0.025879528757243492  0.63063170313298 \\
0.026260460872712887  0.6406412388586178 \\
0.026650327467441948  0.6506507745842555 \\
0.02709917237138645  0.660661500787501 \\
0.027532876533826002  0.6706710365131388 \\
0.02799449871570244  0.6806817627163843 \\
0.02847616255953357  0.6906912984420219 \\
0.02899494670273092  0.7007008341676597 \\
0.029480545839604293  0.7107115603709052 \\
0.030014157159529053  0.7207210960965429 \\
0.030590257715972746  0.7307318222997884 \\
0.031156696698929516  0.7407413580254263 \\
0.03179194814524408  0.750750893751064 \\
0.03241727604774242  0.7607616199543095 \\
0.03308832035666098  0.7707711556799473 \\
0.03386007689708783  0.7807818818831928 \\
0.034596518207277854  0.7907914176088304 \\
0.0354394618092003  0.8008009533344682 \\
0.03630946524363816  0.8108116795377137 \\
0.03725109975668613  0.8208212152633514 \\
0.03818423194518236  0.8308319414665969 \\
0.03919174843543207  0.8408414771922348 \\
0.04033192758480478  0.8508510129178726 \\
0.04171459300639239  0.8608617391211181 \\
0.04318542397266428  0.8708712748467557 \\
0.04478854318625512  0.8808820010500013 \\
0.046712050368638916  0.890891536775639 \\
0.04890128150654824  0.9009010725012767 \\
0.05162854577358792  0.9109117987045222 \\
0.055057029491332704  0.9209213344301601 \\
0.059159006081302576  0.9309320606334055 \\
0.06454645664273528  0.9409415963590433 \\
0.0729670293646135  0.9509511320846811 \\
0.08401393188871326  0.9609618582879266 \\
0.10337528689998361  0.9709713940135642 \\
0.13758329153515575  0.9809821202168097 \\
0.1912326621854664  0.9909916559424475 \\
0.35303353001019167  1.0 \\
    };
\addlegendentry{\textcolor{black}{LIMP (AUC: {0.73})}}    

\addplot [color=cPLOT5, smooth, line width=\pckLineWidth]
table[row sep=crcr]{%
2.5927718727024864e-05  0.0 \\
0.0012040686164023159  0.010010726203245479 \\
0.0016106758507056231  0.020020261928883246 \\
0.0019349927271389894  0.03003098813212873 \\
0.0022228963556631797  0.04004052385776649 \\
0.0024792722984521  0.05005005958340426 \\
0.0027178968896458913  0.060060785786649744 \\
0.002942192086125098  0.07007032151228752 \\
0.0031482191581694236  0.08008104771553298 \\
0.0033501646620401676  0.09009058344117075 \\
0.0035404245767542662  0.10010011916680853 \\
0.0037180400538414735  0.110110845370054 \\
0.0038915590973971972  0.12012038109569177 \\
0.004058339684740129  0.13013110729893726 \\
0.0042201375802065665  0.14014064302457505 \\
0.004375612861166528  0.1501501787502128 \\
0.0045247407262974226  0.16016090495345828 \\
0.004674162863623861  0.17017044067909606 \\
0.0048140861188884125  0.1801811668823415 \\
0.004953209543473962  0.1901907026079793 \\
0.005089190918021323  0.20020023833361705 \\
0.0052227501270803054  0.21021096453686255 \\
0.0053549027734895276  0.2202205002625003 \\
0.005483756565909667  0.2302312264657458 \\
0.005611087141497629  0.24024076219138354 \\
0.005736527668785503  0.2502502979170213 \\
0.00585614617998561  0.2602610241202668 \\
0.005978127394718459  0.2702705598459046 \\
0.006097211721884896  0.2802812860491501 \\
0.006216348639329474  0.2902908217747878 \\
0.006334035531058976  0.3003003575004256 \\
0.006452881293759899  0.31031108370367105 \\
0.006568420242511262  0.3203206194293089 \\
0.0066863450800901765  0.3303313456325543 \\
0.006800071710588085  0.3403408813581921 \\
0.006914758828437578  0.35035041708382986 \\
0.007027333275055937  0.36036114328707536 \\
0.00714203778076052  0.37037067901271314 \\
0.007254328715455975  0.3803814052159586 \\
0.007366474299026267  0.3903909409415964 \\
0.0074802486933675935  0.4004004766672341 \\
0.007595900015811076  0.4104112028704796 \\
0.007710836049443904  0.4204207385961174 \\
0.007824807802335122  0.43043146479936284 \\
0.0079412985290901  0.4404410005250006 \\
0.008055049700763539  0.45045053625063836 \\
0.008170529723884995  0.46046126245388386 \\
0.008287859107398597  0.47047079817952164 \\
0.008404811395025896  0.4804815243827671 \\
0.008523638274784093  0.4904910601084049 \\
0.008644627685499423  0.5005005958340426 \\
0.008765708523394568  0.5105113220372881 \\
0.008887444219946677  0.5205208577629259 \\
0.009009552893600399  0.5305315839661714 \\
0.009134654895946255  0.5405411196918092 \\
0.009259158845911787  0.5505506554174469 \\
0.009386037789429836  0.5605613816206924 \\
0.009514069905110124  0.5705709173463301 \\
0.009645790863736294  0.5805816435495756 \\
0.009780296547554603  0.5905911792752134 \\
0.009913320035690384  0.6006007150008512 \\
0.010050168663436467  0.6106114412040966 \\
0.010191854890591396  0.6206209769297345 \\
0.010332570130400003  0.63063170313298 \\
0.01047568359743251  0.6406412388586178 \\
0.010623442146241264  0.6506507745842555 \\
0.010772527270974894  0.660661500787501 \\
0.010920714700788382  0.6706710365131388 \\
0.011076307142681282  0.6806817627163843 \\
0.011240097546514708  0.6906912984420219 \\
0.01140232493993714  0.7007008341676597 \\
0.011571266691252372  0.7107115603709052 \\
0.011746051790914196  0.7207210960965429 \\
0.01193095377319805  0.7307318222997884 \\
0.012119507967862611  0.7407413580254263 \\
0.012313723435670648  0.750750893751064 \\
0.012512263499728091  0.7607616199543095 \\
0.012716873088013921  0.7707711556799473 \\
0.012931346917643641  0.7807818818831928 \\
0.013158268760308683  0.7907914176088304 \\
0.013395662501653824  0.8008009533344682 \\
0.013650447220324458  0.8108116795377137 \\
0.013907140284632889  0.8208212152633514 \\
0.014182798022159614  0.8308319414665969 \\
0.014475547385376578  0.8408414771922348 \\
0.014779457134051108  0.8508510129178726 \\
0.015107862876815915  0.8608617391211181 \\
0.015471014441952029  0.8708712748467557 \\
0.0158569011139686  0.8808820010500013 \\
0.016289814172387427  0.890891536775639 \\
0.01676790199420777  0.9009010725012767 \\
0.01729582919891046  0.9109117987045222 \\
0.017902786438366056  0.9209213344301601 \\
0.018588524603776038  0.9309320606334055 \\
0.019406937658304847  0.9409415963590433 \\
0.020435025929078747  0.9509511320846811 \\
0.0217773787973479  0.9609618582879266 \\
0.023644137631489696  0.9709713940135642 \\
0.0267839981072391  0.9809821202168097 \\
0.033340404495382135  0.9909916559424475 \\
0.1040703972735706  1.0 \\
    };
\addlegendentry{\textcolor{black}{NeuroMorph (AUC: {0.90})}}    

\end{axis}
\end{tikzpicture}

%% file: figs/tikz/faust_conformal_pck.tex
\newcommand{\pckLineWidth}{1pt}
\newcommand{\plotWidth}{\columnwidth}
\newcommand{\plotHeight}{0.7\columnwidth}
\newcommand{\pckTitle}{\textbf{FAUST}}
\definecolor{cPLOT0}{RGB}{28,213,227}
\definecolor{cPLOT1}{RGB}{80,150,80}
\definecolor{cPLOT2}{RGB}{90,130,213}
\definecolor{cPLOT3}{RGB}{247,179,43}
\definecolor{cPLOT4}{RGB}{124,42,43}
\definecolor{cPLOT5}{RGB}{242,64,0}

\pgfplotsset{%
    label style = {font=\LARGE},
    tick label style = {font=\large},
    title style =  {font=\large},
    legend style={  fill= gray!10,
                    fill opacity=0.6, 
                    font=\large,
                    draw=gray!20, %
                    text opacity=1}
}
\begin{tikzpicture}[scale=0.5, transform shape]
	\begin{axis}[
		width=\plotWidth,
		height=\plotHeight,
		grid=major,
		legend style={
			at={(0.97,0.03)},
			anchor=south east,
			legend columns=1},
		legend cell align={left},
        xlabel={\LARGE Conformal distortion},
		xmin=0,
        xmax=0.5,
        ylabel near ticks,
        xtick={0, 0.2, 0.4},
	ymin=0,
        ymax=1,
        ytick={0, 0.20, 0.40, 0.60, 0.80, 1.0}
	]

\addplot [color=cPLOT1, smooth, line width=\pckLineWidth]
table[row sep=crcr]{%
0.0  0.0 \\
0.006329113924050633  0.5068308733794814 \\
0.012658227848101266  0.6293226332426376 \\
0.0189873417721519  0.6967614011483675 \\
0.02531645569620253  0.741423192921735 \\
0.03164556962025317  0.773880985665413 \\
0.0379746835443038  0.7987646671935019 \\
0.04430379746835443  0.8186813842929738 \\
0.05063291139240506  0.8349710657410371 \\
0.056962025316455694  0.8486144316181178 \\
0.06329113924050633  0.8602169569262164 \\
0.06962025316455696  0.8702203142505602 \\
0.0759493670886076  0.8788998104393406 \\
0.08227848101265822  0.8865044001580505 \\
0.08860759493670886  0.8932562482494398 \\
0.09493670886075949  0.8992508227632843 \\
0.10126582278481013  0.9046005971911012 \\
0.10759493670886076  0.9094325809258963 \\
0.11392405063291139  0.9137976965128841 \\
0.12025316455696203  0.9177418023767606 \\
0.12658227848101267  0.9213475499659891 \\
0.13291139240506328  0.924677959447023 \\
0.13924050632911392  0.9277382825004001 \\
0.14556962025316456  0.930523642565621 \\
0.1518987341772152  0.9331093449903969 \\
0.15822784810126583  0.9355120826164373 \\
0.16455696202531644  0.9377422962848111 \\
0.17088607594936708  0.9398170226972631 \\
0.17721518987341772  0.9417547052556818 \\
0.18354430379746836  0.9435851972631242 \\
0.18987341772151897  0.9453069982394366 \\
0.1962025316455696  0.9469127308238636 \\
0.20253164556962025  0.9484262463988476 \\
0.2088607594936709  0.9498586110055217 \\
0.21518987341772153  0.9512080115637004 \\
0.22151898734177214  0.95249101587508 \\
0.22784810126582278  0.9537130944402209 \\
0.23417721518987342  0.9548544596770967 \\
0.24050632911392406  0.955946465368918 \\
0.24683544303797467  0.956990049315781 \\
0.25316455696202533  0.9579811477172695 \\
0.25949367088607594  0.9589373286951824 \\
0.26582278481012656  0.9598443376880602 \\
0.2721518987341772  0.9607029249359795 \\
0.27848101265822783  0.9615356289012484 \\
0.2848101265822785  0.962332977802897 \\
0.2911392405063291  0.9630960032210307 \\
0.2974683544303797  0.9638257054757522 \\
0.3037974683544304  0.9645293056278009 \\
0.310126582278481  0.965209804637484 \\
0.31645569620253167  0.9658591061639725 \\
0.3227848101265823  0.9664918398887644 \\
0.3291139240506329  0.967095439290573 \\
0.33544303797468356  0.9676804702504802 \\
0.34177215189873417  0.9682492460087228 \\
0.3481012658227848  0.9687893250840269 \\
0.35443037974683544  0.9693129613976472 \\
0.36075949367088606  0.9698227182698463 \\
0.3670886075949367  0.9703181893205826 \\
0.37341772151898733  0.9707822128080986 \\
0.37974683544303794  0.9712420161951825 \\
0.3860759493670886  0.9716845015404929 \\
0.3924050632911392  0.9721140765044815 \\
0.3987341772151899  0.9725310536871798 \\
0.4050632911392405  0.9729408098091389 \\
0.4113924050632911  0.9733302156690141 \\
0.4177215189873418  0.973716933168614 \\
0.4240506329113924  0.9740902713868438 \\
0.43037974683544306  0.9744547317641645 \\
0.43670886075949367  0.9748106269006082 \\
0.4430379746835443  0.9751572690761043 \\
0.44936708860759494  0.9754914072503201 \\
0.45569620253164556  0.975816229943582 \\
0.4620253164556962  0.976137520256482 \\
0.46835443037974683  0.976444618527929 \\
0.47468354430379744  0.9767456210987516 \\
0.4810126582278481  0.9770358702284732 \\
0.4873417721518987  0.9773213053177017 \\
0.49367088607594933  0.9776008635263285 \\
0.5  0.9778758577744878 \\
    };
\addlegendentry{\textcolor{black}{Ours (AUC: \textbf{0.92})}}

\addplot [color=cPLOT4, smooth, dashed, line width=\pckLineWidth]
table[row sep=crcr]{%
0.0  0.0 \\
0.0  0.010010023902667243 \\
0.0  0.020020047805334486 \\
0.0  0.030030036660616436 \\
0.0  0.04004006056328368 \\
0.0  0.050050084465950925 \\
0.0  0.06006007332123287 \\
0.0  0.07007009722390012 \\
0.0  0.08008008607918207 \\
0.0  0.0900901099818493 \\
0.0  0.10010013388451655 \\
3.895244345031301e-05  0.11011012273979849 \\
0.00013584426041735043  0.12012014664246574 \\
0.00023888350586398488  0.1301301354977477 \\
0.0003481399982945277  0.14014015940041494 \\
0.00046326475050051386  0.15015018330308216 \\
0.0005848558056689157  0.16016017215836414 \\
0.000712465702492171  0.17017019606103137 \\
0.0008464642426688762  0.18018018491631335 \\
0.0009873283351655715  0.19019020881898058 \\
0.001134026992707149  0.2002002327216478 \\
0.0012882453884266454  0.21021022157692976 \\
0.0014487980912458731  0.22022024547959698 \\
0.0016169607740672554  0.23023023433487896 \\
0.001792330019998456  0.2402402582375462 \\
0.0019755651016186704  0.25025028214021344 \\
0.002167032058630891  0.2602602709954954 \\
0.002367237055174609  0.2702702948981626 \\
0.0025760378885091306  0.2802802837534446 \\
0.0027933619459243886  0.2902903076561118 \\
0.0030206677718087604  0.30030033155877905 \\
0.003258638562368077  0.31031032041406104 \\
0.003506776426783098  0.3203203443167283 \\
0.003765481968105888  0.3303303331720102 \\
0.0040360666371488385  0.34034035707467747 \\
0.004317711145136016  0.3503503809773447 \\
0.004613004994761027  0.3603603698326267 \\
0.00492146418050865  0.3703703937352939 \\
0.005243850323146226  0.3803803825905759 \\
0.005579577270451441  0.3903904064932431 \\
0.0059310767175975165  0.40040043039591033 \\
0.006297818186358661  0.4104104192511923 \\
0.006682762334261572  0.4204204431538595 \\
0.007082508238734686  0.4304304320091415 \\
0.007503725749036239  0.44044045591180875 \\
0.007943985265022047  0.45045047981447595 \\
0.008403735465237538  0.4604604686697579 \\
0.008886544393329032  0.4704704925724251 \\
0.00939072887393877  0.4804804814277071 \\
0.009921690639837166  0.49049050533037436 \\
0.010477227374045661  0.5005005292330416 \\
0.011059342249455373  0.5105105180883236 \\
0.011672222523399167  0.5205205419909908 \\
0.01231545441496662  0.5305305308462728 \\
0.012992264237856599  0.54054055474894 \\
0.01370579549359574  0.5505505786516072 \\
0.014456144664954262  0.5605605675068892 \\
0.0152462121407555  0.5705705914095565 \\
0.016080749477784373  0.5805806153122236 \\
0.016962878713069784  0.5905906041675056 \\
0.01789188071254655  0.6006006280701729 \\
0.018876396394009287  0.6106106169254548 \\
0.019918770529019802  0.6206206408281221 \\
0.021021916636343896  0.6306306647307892 \\
0.022194473703097017  0.6406406535860713 \\
0.023444891467194307  0.6506506774887384 \\
0.024771365354363972  0.6606606663440204 \\
0.026187156578658666  0.6706706902466877 \\
0.027688525885522175  0.6806807141493549 \\
0.02929269682503044  0.690690703004637 \\
0.031013224506024173  0.7007007269073041 \\
0.0328561763459394  0.710710715762586 \\
0.03484006123364436  0.7207207396652534 \\
0.03697216825830685  0.7307307635679205 \\
0.03927629014363676  0.7407407524232025 \\
0.04177313062523247  0.7507507763258697 \\
0.0444736284264069  0.7607607651811518 \\
0.04739931277324816  0.7707707890838189 \\
0.050586282139549965  0.7807808129864862 \\
0.05409162243620713  0.7907908018417682 \\
0.057927298074015976  0.8008008257444353 \\
0.06214869979423199  0.8108108145997173 \\
0.06682059544777363  0.8208208385023846 \\
0.07203314314595843  0.8308308624050518 \\
0.07785089731945627  0.8408408512603337 \\
0.0843883713988065  0.8508508751630011 \\
0.09180331168217482  0.860860864018283 \\
0.10025521075233135  0.8708708879209501 \\
0.10997536229726546  0.8808809118236175 \\
0.12130006585830921  0.8908909006788994 \\
0.1346511076777266  0.9009009245815666 \\
0.1506459214726914  0.9109109134368487 \\
0.17007385836054523  0.9209209373395159 \\
0.1943545798411903  0.9309309612421831 \\
0.2254313258807512  0.9409409500974649 \\
0.26665274385112747  0.9509509740001323 \\
0.324365982574625  0.9609609628554142 \\
0.4111421627012892  0.9709709867580814 \\
0.5613204487331918  0.9809810106607487 \\
0.9232151316461761  0.9909909995160306 \\
2.0  1.0 \\
    };
\addlegendentry{\textcolor{black}{Hamiltonian (AUC: {0.91})}}    

\addplot [color=cPLOT3, smooth, dashed, line width=\pckLineWidth]
table[row sep=crcr]{%
5.760504517837717e-09  0.0 \\
0.000586758119517139  0.01001006315605227 \\
0.0011837544035939729  0.020020060111485813 \\
0.0017970627999246425  0.03003005706691936 \\
0.002416530543534634  0.04004005402235291 \\
0.0030563546018611554  0.05005005097778645 \\
0.003699576004701388  0.06006011413383872 \\
0.004363729525673321  0.07007011108927226 \\
0.005043959501404238  0.08008010804470582 \\
0.005734678308988883  0.09009010500013936 \\
0.006442326713232305  0.1001001019555729 \\
0.0071613940932606646  0.11011016511162516 \\
0.007899750575418363  0.12012016206705871 \\
0.008658730013374694  0.13013015902249225 \\
0.009436522932434244  0.1401401559779258 \\
0.010230063541324608  0.15015015293335934 \\
0.011036322240990604  0.16016021608941164 \\
0.011860273597355864  0.17017021304484514 \\
0.012704666285259236  0.18018021000027873 \\
0.013570894145947587  0.19019020695571226 \\
0.014467117455922995  0.2002002039111458 \\
0.015380514224658803  0.21021026706719806 \\
0.016312048111159605  0.22022026402263162 \\
0.01727226342476973  0.23023026097806515 \\
0.0182498085433056  0.2402402579334987 \\
0.01925065003504134  0.25025025488893227 \\
0.020276080755141646  0.2602603180449845 \\
0.021339032770570476  0.27027031500041804 \\
0.022431920020002494  0.2802803119558516 \\
0.023548721418365393  0.29029030891128516 \\
0.024710563433071098  0.3003003058667187 \\
0.025894149148000256  0.310310369022771 \\
0.02710244719682997  0.3203203659782045 \\
0.02834360433868266  0.33033036293363804 \\
0.02962372779966005  0.3403403598890716 \\
0.03095492891828666  0.3503503568445051 \\
0.032312794341886164  0.36036042000055746 \\
0.033710118157645574  0.37037041695599093 \\
0.0351581425123344  0.3803804139114245 \\
0.03665363683306522  0.39039041086685805 \\
0.03817941241696943  0.4004004078222916 \\
0.0397678205711558  0.4104104709783439 \\
0.041417069348192115  0.4204204679337774 \\
0.04311741226982635  0.43043046488921094 \\
0.04486544355428501  0.44044046184464447 \\
0.0466483715112842  0.45045045880007806 \\
0.048514367400737335  0.4604605219561303 \\
0.05046940358758656  0.4704705189115639 \\
0.052474194364644955  0.4804805158669974 \\
0.054572256848144374  0.4904905128224309 \\
0.056730028187410575  0.5005005097778645 \\
0.05897540825081782  0.5105105729339168 \\
0.0613136855919989  0.5205205698893502 \\
0.06376281463596101  0.5305305668447838 \\
0.06628694986553585  0.5405405638002174 \\
0.06891612435123706  0.5505505607556509 \\
0.07161953317846503  0.5605606239117032 \\
0.07449380502472058  0.5705706208671368 \\
0.07746532542430629  0.5805806178225703 \\
0.08057165789481369  0.5905906147780038 \\
0.08385609615548706  0.6006006117334374 \\
0.08724339670938663  0.6106106748894896 \\
0.09081578069507223  0.6206206718449232 \\
0.0945756113144296  0.6306306688003568 \\
0.09852449287114862  0.6406406657557904 \\
0.10268270447250893  0.6506506627112238 \\
0.10702425853053255  0.6606607258672761 \\
0.11162184279235587  0.6706707228227097 \\
0.11649258297044307  0.6806807197781432 \\
0.12157522214793115  0.6906907167335767 \\
0.12700606971503925  0.7007007136890102 \\
0.1327784027010095  0.7107107768450625 \\
0.13894076869947725  0.720720773800496 \\
0.14547300039515942  0.7307307707559296 \\
0.15245164912284181  0.7407407677113632 \\
0.15997233541135847  0.7507507646667968 \\
0.16799915161544865  0.7607607616222303 \\
0.17658302178948349  0.7707708247782825 \\
0.1858110728671405  0.7807808217337161 \\
0.19590354457891834  0.7907908186891497 \\
0.2067700762723974  0.8008008156445832 \\
0.21879271177239445  0.8108108126000166 \\
0.23192743957263007  0.8208208757560691 \\
0.24641205205965377  0.8308308727115025 \\
0.26231047911337235  0.840840869666936 \\
0.28016289951767304  0.8508508666223696 \\
0.30024541090074486  0.8608608635778032 \\
0.3230643376104356  0.8708709267338555 \\
0.3494756389179674  0.8808809236892889 \\
0.37939425374739866  0.8908909206447225 \\
0.4142981907154324  0.9009009176001561 \\
0.45576719718565917  0.9109109145555897 \\
0.5062694229602176  0.9209209777116419 \\
0.5685125981960639  0.9309309746670755 \\
0.6491057863746601  0.940940971622509 \\
0.756137873906825  0.9509509685779425 \\
0.9113417238669479  0.9609609655333761 \\
1.158294539295691  0.9709710286894284 \\
1.618134617867288  0.9809810256448618 \\
2.0  0.9909910226002954 \\
    };
\addlegendentry{\textcolor{black}{LIMP (AUC: {0.83})}}    

\addplot [color=cPLOT5, smooth, line width=\pckLineWidth]
table[row sep=crcr]{%
8.389515748774555e-11  0.0 \\
1.4338294515714444e-05  0.010010054963808907 \\
3.057639172732163e-05  0.020020035451921134 \\
4.8768722535585946e-05  0.030030090415730043 \\
6.901925146882703e-05  0.04004007090384227 \\
9.130521400499968e-05  0.050050051391954495 \\
0.00011582445162237053  0.0600601063557634 \\
0.00014271030152985807  0.07007008684387563 \\
0.00017193125360279724  0.08008014180768454 \\
0.00020383182259247067  0.09009012229579677 \\
0.00023816548379407274  0.10010010278390899 \\
0.0002751776702468067  0.1101101577477179 \\
0.0003151621647490721  0.12012013823583013 \\
0.00035822347049070387  0.13013019319963903 \\
0.0004042920054202925  0.14014017368775125 \\
0.00045349660341471583  0.15015015417586347 \\
0.0005061457792322032  0.16016020913967238 \\
0.0005622844486089961  0.17017018962778463 \\
0.0006221399987422549  0.18018024459159354 \\
0.0006862381866039823  0.19019022507970576 \\
0.0007538695614712098  0.20020020556781798 \\
0.0008260115765530428  0.2102102605316269 \\
0.0009023798819899341  0.2202202410197391 \\
0.0009833275421061716  0.23023029598354802 \\
0.001069113276687017  0.24024027647166027 \\
0.0011600687324474278  0.2502502569597725 \\
0.0012569425308847303  0.2602603119235814 \\
0.0013589248682412734  0.27027029241169365 \\
0.0014674813445635147  0.2802803473755025 \\
0.0015822948122920398  0.29029032786361475 \\
0.0017038675303465032  0.30030030835172694 \\
0.001832600374373583  0.31031036331553585 \\
0.001968394074757884  0.32032034380364816 \\
0.002111730297285441  0.33033039876745707 \\
0.002263133786763216  0.34034037925556926 \\
0.002422719066913359  0.3503503597436815 \\
0.0025911253780083188  0.3603604147074904 \\
0.0027702560083075587  0.3703703951956026 \\
0.0029589356212031954  0.3803804501594115 \\
0.0031585118078905736  0.3903904306475237 \\
0.003369377006642682  0.40040041113563596 \\
0.003593659628199375  0.41041046609944487 \\
0.0038293485170429074  0.42042044658755706 \\
0.004078569924916908  0.430430501551366 \\
0.004342396354448374  0.4404404820394782 \\
0.004623197720575711  0.4504504625275904 \\
0.004918191436842956  0.4604605174913993 \\
0.005231274832076593  0.4704704979795116 \\
0.005562725071281527  0.48048055294332054 \\
0.00591324025333153  0.49049053343143273 \\
0.006285100018624323  0.500500513919545 \\
0.006681791959609406  0.5105105688833539 \\
0.00709986123051387  0.5205205493714661 \\
0.007545734129800729  0.530530604335275 \\
0.00801867551563049  0.5405405848233873 \\
0.008520637092714977  0.5505505653114995 \\
0.009056297092582887  0.5605606202753084 \\
0.00962601575488855  0.5705706007634206 \\
0.010233293918605613  0.5805805812515328 \\
0.01087848642435807  0.5905906362153417 \\
0.011567832246025554  0.6006006167034539 \\
0.01230413709565996  0.6106106716672628 \\
0.013091902597708026  0.6206206521553751 \\
0.013941033787269007  0.6306306326434873 \\
0.014842912751587799  0.6406406876072963 \\
0.015812619723260733  0.6506506680954084 \\
0.016848386524551916  0.6606607230592174 \\
0.017964500296892182  0.6706707035473296 \\
0.01916570435227416  0.6806806840354418 \\
0.020456096470840433  0.6906907389992507 \\
0.021857865087150107  0.700700719487363 \\
0.02337968013508027  0.7107107744511718 \\
0.025015846829404716  0.7207207549392841 \\
0.02679774373870085  0.7307307354273963 \\
0.02872833692291723  0.7407407903912052 \\
0.030830636150521906  0.7507507708793174 \\
0.03312477403522615  0.7607608258431263 \\
0.035644555261881344  0.7707708063312385 \\
0.0383985656567783  0.7807807868193508 \\
0.04143957176556696  0.7907908417831596 \\
0.0448193680486693  0.8008008222712719 \\
0.04853766237842347  0.8108108772350807 \\
0.05269909443145033  0.820820857723193 \\
0.057377262362410406  0.8308308382113053 \\
0.06262802820124319  0.8408408931751141 \\
0.0686250370624209  0.8508508736632264 \\
0.07542813502634131  0.8608609286270353 \\
0.08325425346536808  0.8708709091151475 \\
0.09235211796511301  0.8808808896032597 \\
0.10308256634813355  0.8908909445670687 \\
0.1159032784297127  0.9009009250551808 \\
0.13157358648231182  0.9109109800189898 \\
0.15083459262943813  0.9209209605071021 \\
0.17551459054310675  0.9309309409952142 \\
0.2080200544328058  0.9409409959590233 \\
0.2531992722053382  0.9509509764471353 \\
0.3200749975970361  0.9609610314109444 \\
0.4289660322236427  0.9709710118990565 \\
0.6425138846380235  0.9809809923871687 \\
1.2546067115003348  0.9909910473509776 \\
2.0  1.0 \\
    };
\addlegendentry{\textcolor{black}{NeuroMorph (AUC: \textbf{0.92})}}

\end{axis}
\end{tikzpicture}

%% file: figs/tikz/mano_chamfer_pck.tex
\newcommand{\pckLineWidth}{1pt}
\newcommand{\plotWidth}{\columnwidth}
\newcommand{\plotHeight}{0.7\columnwidth}
\newcommand{\pckTitle}{\textbf{FAUST}}
\definecolor{cPLOT0}{RGB}{28,213,227}
\definecolor{cPLOT1}{RGB}{80,150,80}
\definecolor{cPLOT2}{RGB}{90,130,213}
\definecolor{cPLOT3}{RGB}{247,179,43}
\definecolor{cPLOT4}{RGB}{124,42,43}
\definecolor{cPLOT5}{RGB}{242,64,0}

\pgfplotsset{%
    label style = {font=\LARGE},
    tick label style = {font=\large},
    title style =  {font=\large},
    legend style={  fill= gray!10,
                    fill opacity=0.6, 
                    font=\large,
                    draw=gray!20, %
                    text opacity=1}
}
\begin{tikzpicture}[scale=0.5, transform shape]
	\begin{axis}[
		width=\plotWidth,
		height=\plotHeight,
		grid=major,
		legend style={
			at={(0.97,0.03)},
			anchor=south east,
			legend columns=1},
		legend cell align={left},
        xlabel={\LARGE Chamfer distance},
		xmin=0,
        xmax=0.1,
        ylabel near ticks,
        xtick={0, 0.025, 0.05, 0.075, 0.1},
	ymin=0,
        ymax=1,
        ytick={0, 0.20, 0.40, 0.60, 0.80, 1.0}
	]

\addplot [color=cPLOT1, smooth, line width=\pckLineWidth]
table[row sep=crcr]{%
0.0  0.0 \\
0.0012658227848101266  0.05342223650385604 \\
0.002531645569620253  0.29586118251928023 \\
0.00379746835443038  0.580896529562982 \\
0.005063291139240506  0.7763592544987147 \\
0.006329113924050633  0.8832808483290489 \\
0.00759493670886076  0.9378984575835475 \\
0.008860759493670886  0.9667609254498715 \\
0.010126582278481013  0.9826895886889461 \\
0.01139240506329114  0.9909511568123394 \\
0.012658227848101266  0.9954016709511568 \\
0.013924050632911392  0.9976542416452442 \\
0.01518987341772152  0.998788560411311 \\
0.016455696202531647  0.9993894601542417 \\
0.017721518987341773  0.9996818766066838 \\
0.0189873417721519  0.9998650385604113 \\
0.020253164556962026  0.9999389460154242 \\
0.021518987341772152  0.9999678663239074 \\
0.02278481012658228  0.9999775064267352 \\
0.024050632911392405  0.999987146529563 \\
0.02531645569620253  0.9999903598971722 \\
0.026582278481012658  0.9999967866323908 \\
0.027848101265822784  1.0 \\
0.02911392405063291  1.0 \\
0.03037974683544304  1.0 \\
0.03164556962025317  1.0 \\
0.03291139240506329  1.0 \\
0.03417721518987342  1.0 \\
0.035443037974683546  1.0 \\
0.03670886075949367  1.0 \\
0.0379746835443038  1.0 \\
0.039240506329113925  1.0 \\
0.04050632911392405  1.0 \\
0.04177215189873418  1.0 \\
0.043037974683544304  1.0 \\
0.04430379746835443  1.0 \\
0.04556962025316456  1.0 \\
0.04683544303797468  1.0 \\
0.04810126582278481  1.0 \\
0.049367088607594936  1.0 \\
0.05063291139240506  1.0 \\
0.05189873417721519  1.0 \\
0.053164556962025315  1.0 \\
0.05443037974683544  1.0 \\
0.05569620253164557  1.0 \\
0.056962025316455694  1.0 \\
0.05822784810126582  1.0 \\
0.05949367088607595  1.0 \\
0.06075949367088608  1.0 \\
0.06202531645569621  1.0 \\
0.06329113924050633  1.0 \\
0.06455696202531645  1.0 \\
0.06582278481012659  1.0 \\
0.0670886075949367  1.0 \\
0.06835443037974684  1.0 \\
0.06962025316455696  1.0 \\
0.07088607594936709  1.0 \\
0.07215189873417721  1.0 \\
0.07341772151898734  1.0 \\
0.07468354430379746  1.0 \\
0.0759493670886076  1.0 \\
0.07721518987341772  1.0 \\
0.07848101265822785  1.0 \\
0.07974683544303797  1.0 \\
0.0810126582278481  1.0 \\
0.08227848101265824  1.0 \\
0.08354430379746836  1.0 \\
0.08481012658227849  1.0 \\
0.08607594936708861  1.0 \\
0.08734177215189874  1.0 \\
0.08860759493670886  1.0 \\
0.089873417721519  1.0 \\
0.09113924050632911  1.0 \\
0.09240506329113925  1.0 \\
0.09367088607594937  1.0 \\
0.0949367088607595  1.0 \\
0.09620253164556962  1.0 \\
0.09746835443037975  1.0 \\
0.09873417721518987  1.0 \\
0.1  1.0 \\
    };
\addlegendentry{\textcolor{black}{Ours (AUC: \textbf{0.96})}}

\addplot [color=cPLOT4, smooth, dashed, line width=\pckLineWidth]
table[row sep=crcr]{%
0.0  0.0 \\
0.0  0.010012885645519427 \\
0.0  0.020022557913103832 \\
0.0  0.03003223018068824 \\
0.0  0.040041902448272645 \\
0.0001852445339356179  0.050051574715857056 \\
0.0013119840795107042  0.060061246983441466 \\
0.0017121956805050491  0.07007091925102588 \\
0.0020040858882453072  0.08008059151861029 \\
0.00225458084535447  0.09009026378619468 \\
0.002467547999394986  0.10010314943171411 \\
0.0026666235786483522  0.11011282169929851 \\
0.0028455652170523127  0.12012249396688293 \\
0.003019638380822274  0.13013216623446733 \\
0.0031780319292240315  0.14014183850205175 \\
0.003332752718878611  0.15015151076963615 \\
0.0034787213223930326  0.16016118303722057 \\
0.0036221396488795733  0.17017085530480497 \\
0.003761426432748307  0.18018052757238937 \\
0.0038946095192136615  0.1901901998399738 \\
0.0040253965847312926  0.2002030854854932 \\
0.004151757667614904  0.2102127577530776 \\
0.004275582869752418  0.220222430020662 \\
0.004399102873072394  0.23023210228824642 \\
0.00451946366992937  0.24024177455583082 \\
0.004638486873778128  0.2502514468234152 \\
0.0047566096508313005  0.26026111909099964 \\
0.004870349582424193  0.27027079135858406 \\
0.004982419732084153  0.28028046362616843 \\
0.005095734251256682  0.29029334927168793 \\
0.005211196565628143  0.3003030215392723 \\
0.005324417017541187  0.3103126938068567 \\
0.005436666892901182  0.32032236607444114 \\
0.005547896110546463  0.3303320383420255 \\
0.005660592287714757  0.34034171060960994 \\
0.005772889049121083  0.35035138287719436 \\
0.005886409671420394  0.36036105514477873 \\
0.00599756040676589  0.37037072741236315 \\
0.00610808746870214  0.3803803996799476 \\
0.006223687337380609  0.39039328532546697 \\
0.006337849342167502  0.4004029575930514 \\
0.006450535811458825  0.4104126298606358 \\
0.006565610792239524  0.4204223021282202 \\
0.0066826818248304195  0.4304319743958046 \\
0.006799630847290478  0.44044164666338903 \\
0.0069211420339385455  0.4504513189309734 \\
0.007043856215681452  0.4604609911985578 \\
0.007166117463864317  0.47047066346614225 \\
0.007288849601819427  0.48048354911166163 \\
0.007414520408520047  0.49049322137924606 \\
0.007538254053423152  0.5005028936468304 \\
0.007664667665935923  0.5105125659144149 \\
0.007793143346097223  0.5205222381819993 \\
0.007923148852353842  0.5305319104495837 \\
0.008052493655619608  0.5405415827171681 \\
0.008183835229777588  0.5505512549847524 \\
0.00832107156911697  0.5605609272523369 \\
0.008457028206631231  0.5705705995199213 \\
0.008597379680457554  0.5805834851654408 \\
0.008738318400268119  0.5905931574330252 \\
0.008885920576463691  0.6006028297006096 \\
0.009037241923901271  0.610612501968194 \\
0.009185080855128473  0.6206221742357784 \\
0.009339283294720111  0.6306318465033628 \\
0.009491710314655795  0.6406415187709471 \\
0.009651513725788596  0.6506511910385316 \\
0.00981314082781335  0.660660863306116 \\
0.009981145453058041  0.6706737489516353 \\
0.010150514993566295  0.6806834212192199 \\
0.010332900961780737  0.6906930934868042 \\
0.010507957290540012  0.7007027657543887 \\
0.010695818305336611  0.710712438021973 \\
0.010891466197161954  0.7207221102895575 \\
0.01108939402652584  0.7307317825571418 \\
0.011283432635138231  0.7407414548247263 \\
0.011492558222168436  0.7507511270923106 \\
0.011704581348408188  0.7607607993598952 \\
0.011926227218837432  0.7707736850054145 \\
0.012151523417463903  0.780783357272999 \\
0.012390439520070094  0.7907930295405834 \\
0.012631941392818348  0.8008027018081678 \\
0.012886471002315944  0.8108123740757521 \\
0.0131442912598551  0.8208220463433367 \\
0.013427966529607316  0.830831718610921 \\
0.013722102532110338  0.8408413908785054 \\
0.014037715155304175  0.8508510631460898 \\
0.014362615939939021  0.8608639487916092 \\
0.014723083905527427  0.8708736210591936 \\
0.015112910127592557  0.8808832933267781 \\
0.015511693585332822  0.8908929655943625 \\
0.015934100242621464  0.9009026378619468 \\
0.01641135014781215  0.9109123101295313 \\
0.016964261082404602  0.9209219823971156 \\
0.017569164549748176  0.9309316546647001 \\
0.01825608045897499  0.9409413269322845 \\
0.019060362791154223  0.9509509991998689 \\
0.02009015765963302  0.9609638848453883 \\
0.021391711712695882  0.9709735571129727 \\
0.02330645071354966  0.9809832293805572 \\
0.026706517189657275  0.9909929016481415 \\
0.03201730890733827  1.0 \\
    };
\addlegendentry{\textcolor{black}{Hamiltonian (AUC: {0.91})}}    

\addplot [color=cPLOT3, smooth, dashed, line width=\pckLineWidth]
table[row sep=crcr]{%
0.0007600421194840968  0.0 \\
0.0038725288902248843  0.010012885645519427 \\
0.004722770051097494  0.020022557913103832 \\
0.00547513850683237  0.03003223018068824 \\
0.006049357954139746  0.040041902448272645 \\
0.006549215410936905  0.050051574715857056 \\
0.0069288109048760895  0.060061246983441466 \\
0.007348759031807762  0.07007091925102588 \\
0.007753920292934763  0.08008059151861029 \\
0.00811681922321464  0.09009026378619468 \\
0.008440057273448556  0.10010314943171411 \\
0.008785191881445416  0.11011282169929851 \\
0.009088138445294217  0.12012249396688293 \\
0.009378555254789148  0.13013216623446733 \\
0.009670143070324237  0.14014183850205175 \\
0.009959522332719366  0.15015151076963615 \\
0.010231755663243724  0.16016118303722057 \\
0.010491434223718555  0.17017085530480497 \\
0.010766786648316885  0.18018052757238937 \\
0.011060821573107523  0.1901901998399738 \\
0.011317676098298764  0.2002030854854932 \\
0.01160823874962781  0.2102127577530776 \\
0.011931149854511776  0.220222430020662 \\
0.012218725684404297  0.23023210228824642 \\
0.012449685949593375  0.24024177455583082 \\
0.012708504385611075  0.2502514468234152 \\
0.012976889109340495  0.26026111909099964 \\
0.013272593095152994  0.27027079135858406 \\
0.013572714446319929  0.28028046362616843 \\
0.013833557390985358  0.29029334927168793 \\
0.014095099842996612  0.3003030215392723 \\
0.014391258696444744  0.3103126938068567 \\
0.014674918218735986  0.32032236607444114 \\
0.014924361382289444  0.3303320383420255 \\
0.015188599621877245  0.34034171060960994 \\
0.015484732483483506  0.35035138287719436 \\
0.01579015118166796  0.36036105514477873 \\
0.016061157460461065  0.37037072741236315 \\
0.016342313672337815  0.3803803996799476 \\
0.01666274015958204  0.39039328532546697 \\
0.01691371886908407  0.4004029575930514 \\
0.01722080918205749  0.4104126298606358 \\
0.017528092871634583  0.4204223021282202 \\
0.01781924173497105  0.4304319743958046 \\
0.018147037940175702  0.44044164666338903 \\
0.018496518055099654  0.4504513189309734 \\
0.018808120746037795  0.4604609911985578 \\
0.019106304412015065  0.47047066346614225 \\
0.019438665156547248  0.48048354911166163 \\
0.01982898607832172  0.49049322137924606 \\
0.020285487577701176  0.5005028936468304 \\
0.020640279773738293  0.5105125659144149 \\
0.021019831411567123  0.5205222381819993 \\
0.021408024826068506  0.5305319104495837 \\
0.021815443451191473  0.5405415827171681 \\
0.022216225137358023  0.5505512549847524 \\
0.022623121318079302  0.5605609272523369 \\
0.02299709419992691  0.5705705995199213 \\
0.023399046619261694  0.5805834851654408 \\
0.02384907355601884  0.5905931574330252 \\
0.024302672101973143  0.6006028297006096 \\
0.02479288942894479  0.610612501968194 \\
0.025307833861016715  0.6206221742357784 \\
0.025834717450125114  0.6306318465033628 \\
0.026324723611026637  0.6406415187709471 \\
0.02694588505060352  0.6506511910385316 \\
0.02753646242444908  0.660660863306116 \\
0.028144503901583306  0.6706737489516353 \\
0.028824438359550272  0.6806834212192199 \\
0.02949240291077296  0.6906930934868042 \\
0.030095699861258665  0.7007027657543887 \\
0.030856214424116  0.710712438021973 \\
0.03156988646343775  0.7207221102895575 \\
0.03234887517914275  0.7307317825571418 \\
0.03324816220581518  0.7407414548247263 \\
0.034053654687979315  0.7507511270923106 \\
0.035001843304408836  0.7607607993598952 \\
0.036024596938328615  0.7707736850054145 \\
0.03706816024255155  0.780783357272999 \\
0.03813850168298628  0.7907930295405834 \\
0.039458846237380704  0.8008027018081678 \\
0.040601218393743595  0.8108123740757521 \\
0.04194949072220913  0.8208220463433367 \\
0.043367100846982774  0.830831718610921 \\
0.045162003765109884  0.8408413908785054 \\
0.04704366528379902  0.8508510631460898 \\
0.04892736778360308  0.8608639487916092 \\
0.05114887160574684  0.8708736210591936 \\
0.05316717965633835  0.8808832933267781 \\
0.0553467340418208  0.8908929655943625 \\
0.05795837206530116  0.9009026378619468 \\
0.06060568292188692  0.9109123101295313 \\
0.0638939792989377  0.9209219823971156 \\
0.06748864547070021  0.9309316546647001 \\
0.07131743993497561  0.9409413269322845 \\
0.07607642716785958  0.9509509991998689 \\
0.08246749234231578  0.9609638848453883 \\
0.08999074809315667  0.9709735571129727 \\
0.09909606257932778  0.9809832293805572 \\
0.11171014775210374  0.9909929016481415 \\
0.16992338820178426  1.0 \\
    };
\addlegendentry{\textcolor{black}{LIMP (AUC: {0.73})}}    

\addplot [color=cPLOT5, smooth, line width=\pckLineWidth]
table[row sep=crcr]{%
5.8360457254484795e-05  0.0 \\
0.0006990915235069268  0.010012885645519427 \\
0.0009289382427887994  0.020022557913103832 \\
0.001119252603217226  0.03003223018068824 \\
0.0012909318399290028  0.040041902448272645 \\
0.0014558421561747408  0.050051574715857056 \\
0.001610508809805396  0.060061246983441466 \\
0.0017609307549336128  0.07007091925102588 \\
0.0019061715337069516  0.08008059151861029 \\
0.002053567121692846  0.09009026378619468 \\
0.0021975983288224257  0.10010314943171411 \\
0.0023350408723237207  0.11011282169929851 \\
0.0024648009559328525  0.12012249396688293 \\
0.0025918967283051035  0.13013216623446733 \\
0.002714861615630919  0.14014183850205175 \\
0.002832045883464542  0.15015151076963615 \\
0.002950657490235394  0.16016118303722057 \\
0.003059828204489424  0.17017085530480497 \\
0.003169290575995927  0.18018052757238937 \\
0.0032780140762791263  0.1901901998399738 \\
0.0033846729930627015  0.2002030854854932 \\
0.003484237965309625  0.2102127577530776 \\
0.0035846962273915596  0.220222430020662 \\
0.003684933992344568  0.23023210228824642 \\
0.003781705688545865  0.24024177455583082 \\
0.0038772750789632495  0.2502514468234152 \\
0.0039695375047460076  0.26026111909099964 \\
0.004065338953516193  0.27027079135858406 \\
0.004157078030138468  0.28028046362616843 \\
0.004250640892122209  0.29029334927168793 \\
0.004345244238210644  0.3003030215392723 \\
0.004440449689460266  0.3103126938068567 \\
0.004534834100243978  0.32032236607444114 \\
0.004628714158331521  0.3303320383420255 \\
0.004720169998770729  0.34034171060960994 \\
0.004814568406010037  0.35035138287719436 \\
0.004909071700667321  0.36036105514477873 \\
0.005000230194521393  0.37037072741236315 \\
0.005095570838497732  0.3803803996799476 \\
0.005186781038489907  0.39039328532546697 \\
0.005282408827779248  0.4004029575930514 \\
0.005375993009673676  0.4104126298606358 \\
0.005470348395143252  0.4204223021282202 \\
0.005566938398226419  0.4304319743958046 \\
0.005660819047234441  0.44044164666338903 \\
0.005755187508343908  0.4504513189309734 \\
0.005852075704780208  0.4604609911985578 \\
0.0059494633942805296  0.47047066346614225 \\
0.006045166790049659  0.48048354911166163 \\
0.00614499929033537  0.49049322137924606 \\
0.006242874437339352  0.5005028936468304 \\
0.006343224291667043  0.5105125659144149 \\
0.006444750180395236  0.5205222381819993 \\
0.006547449757689047  0.5305319104495837 \\
0.006651515046061376  0.5405415827171681 \\
0.00675545752630483  0.5505512549847524 \\
0.006862937234318392  0.5605609272523369 \\
0.0069726592880097514  0.5705705995199213 \\
0.007078952972401391  0.5805834851654408 \\
0.007190722796195133  0.5905931574330252 \\
0.007305778066401171  0.6006028297006096 \\
0.007422427707572893  0.610612501968194 \\
0.007539535371479813  0.6206221742357784 \\
0.007658300991908243  0.6306318465033628 \\
0.007778561987697027  0.6406415187709471 \\
0.007904832760808014  0.6506511910385316 \\
0.008032945099951923  0.660660863306116 \\
0.00816196285248828  0.6706737489516353 \\
0.00829812584142357  0.6806834212192199 \\
0.008434759999014798  0.6906930934868042 \\
0.008574846295183547  0.7007027657543887 \\
0.008718367347669372  0.710712438021973 \\
0.008869892662106365  0.7207221102895575 \\
0.009024439516070213  0.7307317825571418 \\
0.009190209490194149  0.7407414548247263 \\
0.009359380337958276  0.7507511270923106 \\
0.009536545443376206  0.7607607993598952 \\
0.009717373355427076  0.7707736850054145 \\
0.009909501684828131  0.780783357272999 \\
0.010103787675596576  0.7907930295405834 \\
0.010317896168698434  0.8008027018081678 \\
0.010536981564076515  0.8108123740757521 \\
0.010769275425505159  0.8208220463433367 \\
0.01101468595360853  0.830831718610921 \\
0.01127734601215385  0.8408413908785054 \\
0.011559814006437828  0.8508510631460898 \\
0.01186338789287585  0.8608639487916092 \\
0.01219217372776399  0.8708736210591936 \\
0.01256093477073669  0.8808832933267781 \\
0.012960220704459966  0.8908929655943625 \\
0.013391189978855203  0.9009026378619468 \\
0.013867881942289634  0.9109123101295313 \\
0.014399342070120967  0.9209219823971156 \\
0.014972748488910987  0.9309316546647001 \\
0.015663087348933225  0.9409413269322845 \\
0.016484715936591062  0.9509509991998689 \\
0.017407678032780868  0.9609638848453883 \\
0.01862291828233124  0.9709735571129727 \\
0.020399791852890023  0.9809832293805572 \\
0.023936125218029267  0.9909929016481415 \\
0.0370243012619249  1.0 \\
    };
\addlegendentry{\textcolor{black}{NeuroMorph (AUC: {0.93})}} 

\end{axis}
\end{tikzpicture}

%% file: figs/tikz/mano_conformal_pck.tex
\newcommand{\pckLineWidth}{1pt}
\newcommand{\plotWidth}{\columnwidth}
\newcommand{\plotHeight}{0.7\columnwidth}
\newcommand{\pckTitle}{\textbf{FAUST}}
\definecolor{cPLOT0}{RGB}{28,213,227}
\definecolor{cPLOT1}{RGB}{80,150,80}
\definecolor{cPLOT2}{RGB}{90,130,213}
\definecolor{cPLOT3}{RGB}{247,179,43}
\definecolor{cPLOT4}{RGB}{124,42,43}
\definecolor{cPLOT5}{RGB}{242,64,0}

\pgfplotsset{%
    label style = {font=\LARGE},
    tick label style = {font=\large},
    title style =  {font=\large},
    legend style={  fill= gray!10,
                    fill opacity=0.6, 
                    font=\large,
                    draw=gray!20, %
                    text opacity=1}
}
\begin{tikzpicture}[scale=0.5, transform shape]
	\begin{axis}[
		width=\plotWidth,
		height=\plotHeight,
		grid=major,
		legend style={
			at={(0.97,0.03)},
			anchor=south east,
			legend columns=1},
		legend cell align={left},
        xlabel={\LARGE Conformal distortion},
		xmin=0,
        xmax=0.5,
        ylabel near ticks,
        xtick={0, 0.2, 0.4},
	ymin=0,
        ymax=1,
        ytick={0, 0.20, 0.40, 0.60, 0.80, 1.0}
	]

\addplot [color=cPLOT1, smooth, line width=\pckLineWidth]
table[row sep=crcr]{%
0.0  0.0 \\
0.006329113924050633  0.6650048764629389 \\
0.012658227848101266  0.8017085907022107 \\
0.0189873417721519  0.8652592652795839 \\
0.02531645569620253  0.901065100780234 \\
0.03164556962025317  0.9234891092327698 \\
0.0379746835443038  0.938700422626788 \\
0.04430379746835443  0.9494276251625487 \\
0.05063291139240506  0.9573786979843953 \\
0.056962025316455694  0.9634635078023407 \\
0.06329113924050633  0.9681711232119636 \\
0.06962025316455696  0.9719765929778934 \\
0.0759493670886076  0.9750853381014304 \\
0.08227848101265822  0.9776326804291288 \\
0.08860759493670886  0.9798118498049415 \\
0.09493670886075949  0.9816013085175552 \\
0.10126582278481013  0.9831445058517555 \\
0.10759493670886076  0.9844857363459038 \\
0.11392405063291139  0.9856682786085825 \\
0.12025316455696203  0.9867079811443433 \\
0.12658227848101267  0.9875975292587776 \\
0.13291139240506328  0.9884322171651495 \\
0.13924050632911392  0.9891417425227568 \\
0.14556962025316456  0.9897752763328999 \\
0.1518987341772152  0.9903724398569571 \\
0.15822784810126583  0.9909114921976593 \\
0.16455696202531644  0.9913802421976593 \\
0.17088607594936708  0.9918002275682705 \\
0.17721518987341772  0.9921944083224967 \\
0.18354430379746836  0.9925725373862159 \\
0.18987341772151897  0.9929126706762028 \\
0.1962025316455696  0.9932265929778934 \\
0.20253164556962025  0.9935163361508452 \\
0.2088607594936709  0.9937792587776333 \\
0.21518987341772153  0.9940200341352405 \\
0.22151898734177214  0.9942534947984395 \\
0.22784810126582278  0.9944621667750325 \\
0.23417721518987342  0.9946669782184655 \\
0.24050632911392406  0.9948508615084526 \\
0.24683544303797467  0.9950237727568271 \\
0.25316455696202533  0.9951865247074122 \\
0.25949367088607594  0.9953547626788036 \\
0.26582278481012656  0.9955004470091028 \\
0.2721518987341772  0.9956453185955787 \\
0.27848101265822783  0.9957777958387516 \\
0.2848101265822785  0.9959058029908973 \\
0.2911392405063291  0.9960258858907672 \\
0.2974683544303797  0.9961354031209363 \\
0.3037974683544304  0.99624613946684 \\
0.310126582278481  0.9963524057217165 \\
0.31645569620253167  0.9964373374512354 \\
0.3227848101265823  0.9965257233420026 \\
0.3291139240506329  0.9966161410923277 \\
0.33544303797468356  0.9966961963589077 \\
0.34177215189873417  0.9967762516254877 \\
0.3481012658227848  0.996849195383615 \\
0.35443037974683544  0.996920716840052 \\
0.36075949367088606  0.9969932542262678 \\
0.3670886075949367  0.9970659947984395 \\
0.37341772151898733  0.9971259346553967 \\
0.37974683544303794  0.9971866872561769 \\
0.3860759493670886  0.9972397187906372 \\
0.3924050632911392  0.9972974236020806 \\
0.3987341772151899  0.997356144343303 \\
0.4050632911392405  0.9974071440182055 \\
0.4113924050632911  0.9974613946684006 \\
0.4177215189873418  0.9975065019505852 \\
0.4240506329113924  0.9975593302990897 \\
0.43037974683544306  0.9976046407672302 \\
0.43670886075949367  0.9976467002600781 \\
0.4430379746835443  0.9976901820546163 \\
0.44936708860759494  0.9977275682704811 \\
0.45569620253164556  0.9977688150195059 \\
0.4620253164556962  0.9978088426527958 \\
0.46835443037974683  0.9978456193107932 \\
0.47468354430379744  0.9978799577373212 \\
0.4810126582278481  0.9979134834200261 \\
0.4873417721518987  0.9979498537061119 \\
0.49367088607594933  0.9979823634590377 \\
0.5  0.9980160923276983 \\
    };
\addlegendentry{\textcolor{black}{Ours (AUC: \textbf{0.97})}}    

\addplot [color=cPLOT4, smooth, dashed, line width=\pckLineWidth]
table[row sep=crcr]{%
0.0  0.0 \\
0.0  0.010010040733768141 \\
0.0  0.020020081467536283 \\
0.0  0.030030122201304424 \\
0.0  0.04004006731814305 \\
0.0  0.050050108051911193 \\
0.0  0.060060148785679335 \\
0.0  0.07007009390251796 \\
0.0  0.0800801346362861 \\
0.0  0.09009017537005425 \\
0.0  0.10010012048689289 \\
1.482390978075898e-05  0.11011016122066103 \\
5.0935584054023765e-05  0.12012020195442917 \\
8.79697570713489e-05  0.1301301470712678 \\
0.00012686160895327702  0.14014018780503593 \\
0.0001666923224210137  0.15015022853880408 \\
0.00020767119219957663  0.1601601736556427 \\
0.000250451951356645  0.17017021438941085 \\
0.0002945670290142033  0.18018025512317898 \\
0.00034041136817908324  0.19019020024001762 \\
0.000387723786976224  0.20020024097378578 \\
0.00043660388004269636  0.2102102817075539 \\
0.00048732087419054435  0.22022022682439255 \\
0.0005395458703336686  0.23023026755816065 \\
0.000593956771417048  0.2402403082919288 \\
0.0006499703836919223  0.25025025340876744 \\
0.0007079922492227552  0.2602602941425356 \\
0.000768001645354488  0.27027033487630375 \\
0.0008303850498473153  0.28028037561007185 \\
0.0008949878436801838  0.29029032072691047 \\
0.0009619352767808564  0.3003003614606786 \\
0.0010311039761282004  0.3103104021944468 \\
0.0011025218023277717  0.3203203473112854 \\
0.0011763732555333916  0.33033038804505355 \\
0.001253194431273563  0.3403404287788217 \\
0.0013326163595641205  0.35035037389566037 \\
0.0014151362709290182  0.36036041462942847 \\
0.0015009116072443262  0.37037045536319657 \\
0.0015889624118427647  0.38038040048003524 \\
0.0016810582465126343  0.39039044121380334 \\
0.0017765976404692907  0.40040048194757155 \\
0.0018758681763307193  0.4104104270644101 \\
0.0019792440132246013  0.4204204677981783 \\
0.002086324348292301  0.4304305085319464 \\
0.0021972116979361188  0.4404404536487851 \\
0.0023126700171273384  0.4504504943825532 \\
0.002433391295326892  0.4604605351163213 \\
0.0025598299290791712  0.47047048023315996 \\
0.0026903972625101245  0.4804805209669281 \\
0.002826907804959866  0.49049056170069627 \\
0.002969350537224713  0.5005005068175349 \\
0.0031179265462965232  0.510510547551303 \\
0.003273047518346317  0.5205205882850712 \\
0.003434945894031216  0.5305305334019098 \\
0.003603832331043577  0.540540574135678 \\
0.0037810751558760456  0.5505506148694461 \\
0.003965718437214783  0.5605606556032142 \\
0.004159090727869952  0.5705706007200528 \\
0.004364145348851167  0.5805806414538209 \\
0.004578558152556055  0.5905906821875891 \\
0.004802706577491733  0.6006006273044278 \\
0.005039107436356716  0.610610668038196 \\
0.005287602402785652  0.6206207087719641 \\
0.0055478579622958435  0.6306306538888027 \\
0.005824348201750418  0.6406406946225708 \\
0.006115657270828123  0.650650735356339 \\
0.006423589289203946  0.6606606804731776 \\
0.006750241038743887  0.6706707212069457 \\
0.007096235768585046  0.6806807619407138 \\
0.007463138127257096  0.6906907070575525 \\
0.007850854785317728  0.7007007477913207 \\
0.008264324249664234  0.7107107885250887 \\
0.008708440581280818  0.7207207336419275 \\
0.009180804200061664  0.7307307743756956 \\
0.00968993667714777  0.7407408151094637 \\
0.010232300149323947  0.7507507602263023 \\
0.010817344375713134  0.7607608009600705 \\
0.01144974542126924  0.7707708416938387 \\
0.012135507392177303  0.7807807868106772 \\
0.012878755739645698  0.7907908275444454 \\
0.013686735171070552  0.8008008682782135 \\
0.014563845984482881  0.8108108133950522 \\
0.0155274188789841  0.8208208541288202 \\
0.01659100944235874  0.8308308948625884 \\
0.017769311273570485  0.8408409355963566 \\
0.01907591268629787  0.8508508807131951 \\
0.020548713013043063  0.8608609214469634 \\
0.022207334952425127  0.8708709621807315 \\
0.024092907593901456  0.8808809072975702 \\
0.026277506409395635  0.8908909480313383 \\
0.02878542784297231  0.9009009887651064 \\
0.03175707192814805  0.9109109338819451 \\
0.035319118113002014  0.9209209746157131 \\
0.039695572359799945  0.9309310153494813 \\
0.045179788792320076  0.9409409604663199 \\
0.052285803183745115  0.9509510012000881 \\
0.0620852402628472  0.9609610419338562 \\
0.0766109123414644  0.9709709870506948 \\
0.10138446649637664  0.980981027784463 \\
0.1599049271590422  0.990991068518231 \\
0.5  1.0 \\
    };
\addlegendentry{\textcolor{black}{Hamiltonian (AUC: \textbf{0.97})}}    

\addplot [color=cPLOT3, smooth, dashed, line width=\pckLineWidth]
table[row sep=crcr]{%
0.00042696517027884795  0.010010115953279141 \\
0.0008551290838614278  0.02002005129678719 \\
0.0012900231622547942  0.030030167250066332 \\
0.0017250534589470234  0.04004010259357438 \\
0.0021692222511187076  0.05005021854685352 \\
0.0026213204889500297  0.06006015389036156 \\
0.003079076223060806  0.0700700892338696 \\
0.0035525704485768195  0.08008020518714876 \\
0.004025732289019146  0.09009014053065678 \\
0.0045072147648737015  0.10010025648393593 \\
0.004995814980099045  0.11011019182744398 \\
0.005500833573497843  0.12012012717095204 \\
0.006006556826641418  0.13013024312423116 \\
0.006526078712701278  0.1401401784677392 \\
0.007046114178535845  0.15015029442101835 \\
0.00757895410705034  0.1601602297645264 \\
0.008125941853659224  0.17017034571780554 \\
0.008674727525545212  0.18018028106131356 \\
0.00924352410268714  0.19019021640482162 \\
0.009817243991484315  0.20020033235810078 \\
0.010406030576864289  0.21021026770160883 \\
0.011001461625498532  0.22022038365488797 \\
0.011608317969650006  0.23023031899839602 \\
0.012226681859182964  0.24024025434190407 \\
0.012857131878939665  0.2502503702951832 \\
0.013507773385851962  0.26026030563869124 \\
0.014168701689470264  0.2702704215919704 \\
0.01483970279904101  0.2802803569354784 \\
0.01552020402853449  0.29029029227898645 \\
0.01621794938544907  0.3003004082322656 \\
0.01693106507313491  0.31031034357577364 \\
0.017654865329419067  0.3203204595290528 \\
0.018400773122845937  0.33033039487256083 \\
0.01915356095087084  0.34034051082584 \\
0.019917797150368344  0.3503504461693481 \\
0.020698619664496487  0.3603603815128561 \\
0.021499503413838994  0.3703704974661352 \\
0.022334988407484158  0.38038043280964323 \\
0.0232010818247681  0.3903905487629224 \\
0.02407469537394702  0.4004004841064305 \\
0.024975295733120628  0.4104104194499385 \\
0.02590981654186164  0.42042053540321767 \\
0.026852821973942564  0.43043047074672564 \\
0.027820223816217737  0.4404405867000048 \\
0.028809544505954854  0.4504505220435129 \\
0.029836627152212802  0.46046063799679204 \\
0.0308899096848414  0.47047057334030007 \\
0.03198210186766426  0.48048050868380815 \\
0.03309232994516398  0.4904906246370872 \\
0.034232034287449586  0.5005005599805953 \\
0.035421006607034844  0.5105106759338744 \\
0.03663613866757798  0.5205206112773825 \\
0.03788673027988532  0.5305305466208905 \\
0.03918646230151834  0.5405406625741697 \\
0.040521966497325757  0.5505505979176777 \\
0.041912826855894814  0.5605607138709569 \\
0.043350056930535175  0.5705706492144649 \\
0.044838903010234965  0.5805805845579729 \\
0.046378623512392636  0.5905907005112521 \\
0.04797097998516575  0.6006006358547602 \\
0.04964849343537514  0.6106107518080393 \\
0.05140029213822572  0.6206206871515473 \\
0.05322853514674497  0.6306308031048264 \\
0.05512041382729116  0.6406407384483345 \\
0.057119557647606456  0.6506506737918426 \\
0.05920527579081991  0.6606607897451217 \\
0.06135612946990543  0.6706707250886298 \\
0.06364303904240165  0.6806808410419088 \\
0.06603596590081251  0.6906907763854169 \\
0.06854201072419963  0.700700711728925 \\
0.0711944831366651  0.7107108276822041 \\
0.07402771369416294  0.7207207630257122 \\
0.07702934143545104  0.7307308789789912 \\
0.0802117403890068  0.7407408143224994 \\
0.0835261983151292  0.7507509302757785 \\
0.08706558922516905  0.7607608656192865 \\
0.09083293431970807  0.7707708009627946 \\
0.09485313906401327  0.7807809169160737 \\
0.09920306454914263  0.7907908522595818 \\
0.10387699715101739  0.800800968212861 \\
0.1089925047864777  0.8108109035563689 \\
0.1143909167640853  0.820820838899877 \\
0.12030275156220771  0.8308309548531561 \\
0.12670684287950662  0.8408408901966642 \\
0.13388392462948673  0.8508510061499434 \\
0.14168921899967257  0.8608609414934513 \\
0.15035269878986313  0.8708708768369594 \\
0.16028695802192017  0.8808809927902385 \\
0.17149278536595114  0.8908909281337466 \\
0.1847192174685066  0.9009010440870258 \\
0.19986691368449172  0.9109109794305339 \\
0.21781611316070615  0.920921095383813 \\
0.23952173946530575  0.9309310307273209 \\
0.2670319655364102  0.940940966070829 \\
0.3032399883203847  0.9509510820241082 \\
0.3524484502081613  0.9609610173676163 \\
0.4279376056138813  0.9709711333208954 \\
0.5605922040770368  0.9809810686644034 \\
0.9072214014203572  0.9909910040079114 \\
1.0  1.0 \\
    };
\addlegendentry{\textcolor{black}{LIMP (AUC: {0.90})}}    

\addplot [color=cPLOT5, smooth, line width=\pckLineWidth]
table[row sep=crcr]{%
2.15508677570142e-10  0.0 \\
1.0836969841498956e-05  0.010010161331713533 \\
2.2997394324342222e-05  0.020020119477429998 \\
3.6137143296421925e-05  0.030030077623146464 \\
5.041030945633906e-05  0.040040238954859995 \\
6.595475753545799e-05  0.05005019710057646 \\
8.272024936655243e-05  0.06006015524629293 \\
0.00010058839370188366  0.07007011339200939 \\
0.0001197681337652412  0.08008027472372292 \\
0.000140296674261009  0.09009023286943937 \\
0.00016194304840162486  0.10010019101515585 \\
0.0001851002784764688  0.1101101491608723 \\
0.00020979204379365513  0.12012031049258586 \\
0.00023595529728908103  0.1301302686383023 \\
0.000263806613398998  0.14014022678401877 \\
0.0002930403915945234  0.15015018492973523 \\
0.00032437000101701583  0.16016034626144876 \\
0.00035731892808481546  0.17017030440716524 \\
0.0003926109519108323  0.1801802625528817 \\
0.00042965296933807903  0.19019022069859814 \\
0.00046889825637954626  0.2002003820303117 \\
0.0005098896633432659  0.21021034017602816 \\
0.0005537401692949472  0.2202202983217446 \\
0.0005996234331057338  0.2302302564674611 \\
0.0006483559485895184  0.2402404177991746 \\
0.0006998420522739578  0.2502503759448911 \\
0.0007541747382578734  0.26026033409060756 \\
0.0008113127346620174  0.270270292236324 \\
0.0008710976020013916  0.28028045356803755 \\
0.0009340743757599412  0.290290411713754 \\
0.0010003225054203923  0.30030036985947045 \\
0.0010693780528878705  0.31031032800518693 \\
0.0011421039985684445  0.3203204893369005 \\
0.0012187321359387936  0.330330447482617 \\
0.0012986536720191566  0.3403404056283334 \\
0.0013826598237319133  0.3503503637740499 \\
0.0014705049825254335  0.3603605251057634 \\
0.001562367896793404  0.37037048325147987 \\
0.0016590001646656916  0.3803804413971963 \\
0.0017606128630944795  0.39039039954291277 \\
0.00186637783963981  0.40040056087462633 \\
0.001976722517674645  0.41041051902034276 \\
0.002093608535292546  0.42042047716605924 \\
0.002215871154770177  0.4304304353117757 \\
0.0023443437029539993  0.4404405966434892 \\
0.002477138680389146  0.45045055478920565 \\
0.0026167254379918957  0.4604605129349222 \\
0.002763473949407569  0.47047047108063866 \\
0.0029178330727132715  0.48048063241235217 \\
0.0030790826311640983  0.49049059055806865 \\
0.003248092500221844  0.5005005487037851 \\
0.0034257812367373573  0.5105107100354986 \\
0.0036123996076677044  0.5205206681812151 \\
0.0038061188736971726  0.5305306263269316 \\
0.004010140249887861  0.540540584472648 \\
0.0042258295021657235  0.5505507458043616 \\
0.004450747083162376  0.560560703950078 \\
0.004688990564931217  0.5705706620957944 \\
0.004941216378886004  0.5805806202415109 \\
0.005203412311395005  0.5905907815732244 \\
0.005479203096225759  0.6006007397189409 \\
0.005767233913042879  0.6106106978646574 \\
0.006074642442185407  0.6206206560103739 \\
0.006397634293309018  0.6306308173420874 \\
0.006738471847379479  0.6406407754878038 \\
0.0071002558061779904  0.6506507336335203 \\
0.007483988683030773  0.6606606917792368 \\
0.007890335574960083  0.6706708531109503 \\
0.00832052293200558  0.6806808112566668 \\
0.008779926866127673  0.6906907694023832 \\
0.009271021596280881  0.7007007275480998 \\
0.00978818523806133  0.7107108888798132 \\
0.010342499608361422  0.7207208470255297 \\
0.010932606658749044  0.7307308051712462 \\
0.011563428991732749  0.7407407633169626 \\
0.012246414918714343  0.7507509246486762 \\
0.0129778444346341  0.7607608827943926 \\
0.013765926246294845  0.7707708409401092 \\
0.014620931088336953  0.7807807990858255 \\
0.015548803526712972  0.7907909604175392 \\
0.01656278991799347  0.8008009185632555 \\
0.017663578591663585  0.810810876708972 \\
0.01886511087006415  0.8208208348546884 \\
0.0201925179446798  0.830830996186402 \\
0.021676912239538024  0.8408409543321185 \\
0.023331689489198837  0.850850912477835 \\
0.025197033177907356  0.8608608706235514 \\
0.027284143416410167  0.8708710319552649 \\
0.029691854631849954  0.8808809901009815 \\
0.032470525515854116  0.8908909482466979 \\
0.03572724856394114  0.9009009063924144 \\
0.03960472725846404  0.9109110677241279 \\
0.04430023388586779  0.9209210258698444 \\
0.05012779961968245  0.9309309840155607 \\
0.057665880747115844  0.9409409421612773 \\
0.06787067490062544  0.9509511034929907 \\
0.08266634433382637  0.9609610616387073 \\
0.10660387763943113  0.9709710197844237 \\
0.1538802083493424  0.9809811811161373 \\
0.30309193511813265  0.9909911392618537 \\
1.0  1.0 \\
    };
\addlegendentry{\textcolor{black}{NeuroMorph (AUC: {0.96})}}

\end{axis}
\end{tikzpicture}

%% file: figs/lung_interpolation.tex
\def\heightIQ{2cm}
\def\widthIQ{8cm}
\def\pathIQ{figs/ours/}
\def\firstRowIQ{000-007}
\def\secondRowIQ{001-004}
\begin{tabular}{c}%
    \setlength{\tabcolsep}{0pt} 
    \includegraphics[height=\heightIQ, width=\widthIQ]{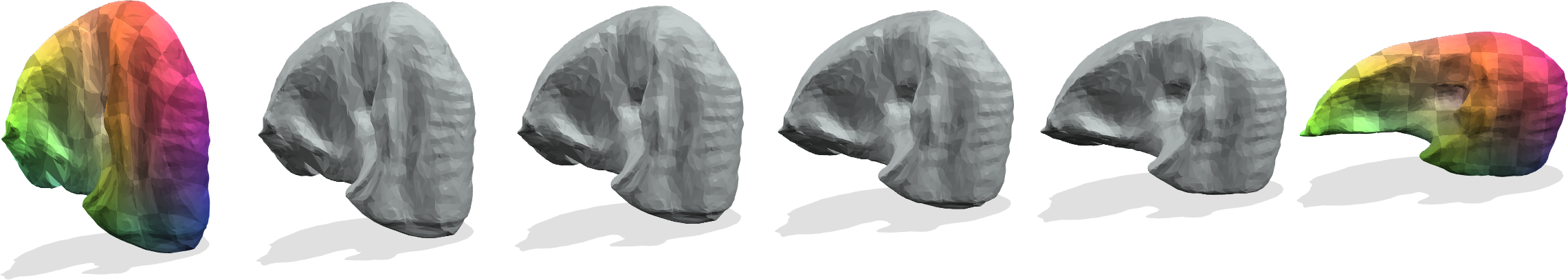}
    \\
\end{tabular}

%% file: figs/lung_ssm.tex
\def\plusEnd{_plus}
\def\minusEnd{_minus}
\def\columnZero{mean_shape}
\def\columnOneT{component_1}
\def\columnTwoT{component_2}
\def\columnThreeT{component_3}
\def\columnFourT{component_4}
\def\columnFiveT{component_5}
\def\hspaceColsT{-0.3cm}
\def\heightT{1.3cm}
\def\widthT{1.3cm}
\begin{tabular}{lccccc}%
        \setlength{\tabcolsep}{0pt} 
        \rotatedCentering{90}{\heightT}{\textbf{\small{Mean}}}&
        \multicolumn{5}{c}{\includegraphics[height=\heightT, width=\widthT]{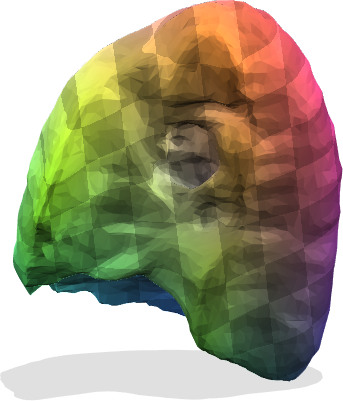}}\\
        \rotatedCentering{90}{\heightT}{\small{$\mathbf{+}$}}&
        \hspace{\hspaceColsT}
        \includegraphics[height=\heightT, width=\widthT]{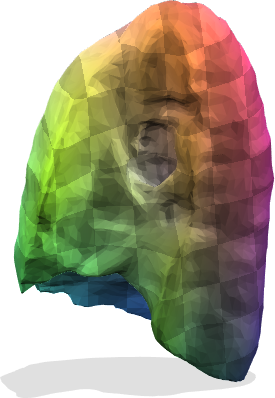}&
        \hspace{\hspaceColsT}
        \includegraphics[height=\heightT, width=\widthT]{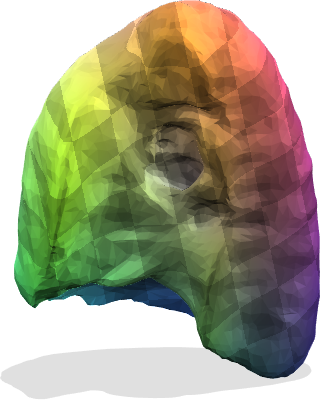}&
        \hspace{\hspaceColsT}
        \includegraphics[height=\heightT, width=\widthT]{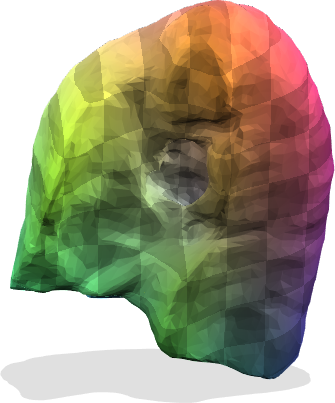}&
        \hspace{\hspaceColsT}
        \includegraphics[height=\heightT, width=\widthT]{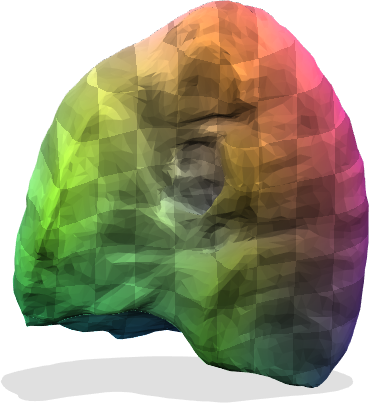}&
        \hspace{\hspaceColsT}
        \includegraphics[height=\heightT, width=\widthT]{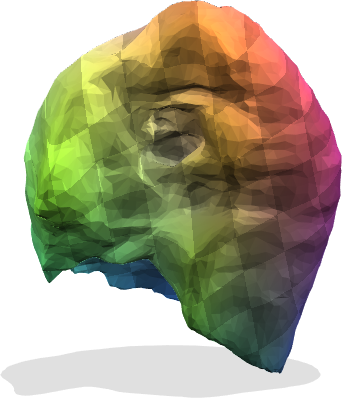}\\
        \rotatedCentering{90}{\heightT}{\small{$\mathbf{-}$}}&
        \hspace{\hspaceColsT}
        \includegraphics[height=\heightT, width=\widthT]{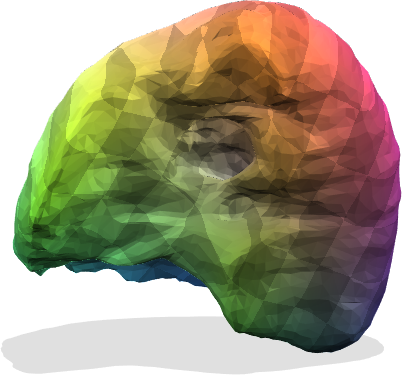}&
        \hspace{\hspaceColsT}
        \includegraphics[height=\heightT, width=\widthT]{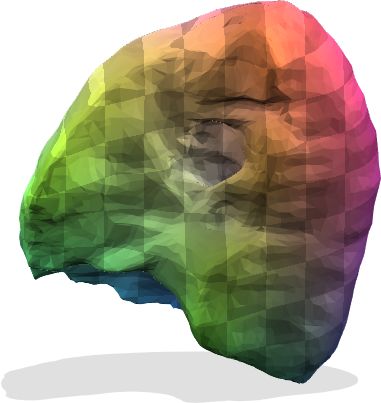}&
        \hspace{\hspaceColsT}
        \includegraphics[height=\heightT, width=\widthT]{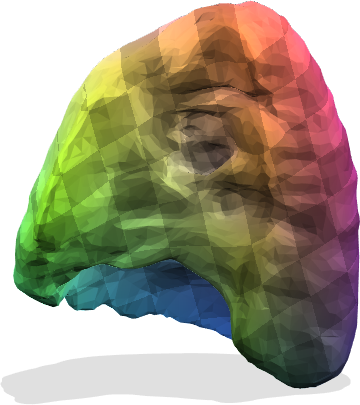}&
        \hspace{\hspaceColsT}
        \includegraphics[height=\heightT, width=\widthT]{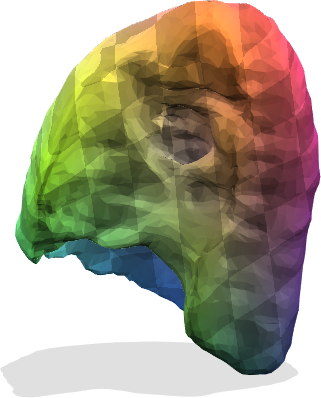}&
        \hspace{\hspaceColsT}
        \includegraphics[height=\heightT, width=\widthT]{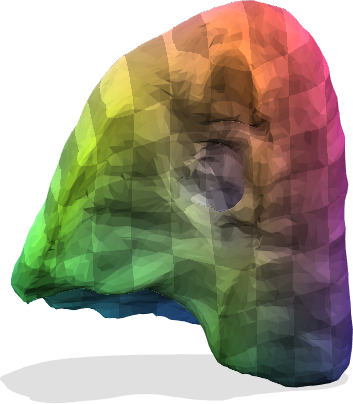}\\
        &
        \hspace{\hspaceColsT}
        \small{$\mathbf{1{st}}$}&
        \hspace{\hspaceColsT}
        \small{$\mathbf{2{nd}}$}&
        \hspace{\hspaceColsT}
        \small{$\mathbf{3{rd}}$}&
        \hspace{\hspaceColsT}
        \small{$\mathbf{4{th}}$}&
        \hspace{\hspaceColsT}
        \small{$\mathbf{5{th}}$}\\
    \end{tabular}

%% file: sec/6_ablation.tex
\section{Ablation study}
\label{sec:ablation}
In this section we conduct ablative experiments to analyse individual components of our approach. To this end, we consider only using the spectral loss $L_{\mathrm{spectral}}$ in~\cref{eq:l_fmap} for shape matching or use the feature similarity to obtain the final point-wise correspondences instead of test-time adaptation. For the experiment, we consider the challenging non-isometric SMAL and DT4D-H dataset.

\begin{figure}[ht!]
    \centering
    \input{figs/ablation}
    \caption{ 
    \textbf{Qualitative comparison on the SMAL dataset.} By harmonising spectral and spatial maps, our method obtains more accurate and smooth correspondences.}
    \vspace{-0.4cm}
    \label{fig:ablation}
\end{figure}
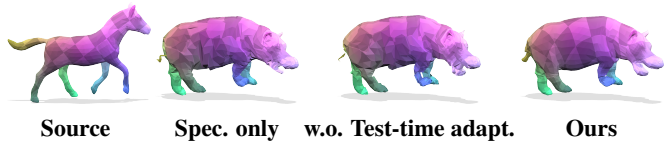

\noindent \textbf{Results.}~\cref{tab:ablation} summarises the {quantitative} results. By comparing the first and the second rows, we can conclude that the spatial regularisation (i.e.\ $L_{\mathrm{spatial}}$ in~\cref{eq:spatial}) plays an important role in obtaining more accurate point-wise correspondences. By comparing the second and the third rows, we notice that the test-time adaptation can further boost the matching performance. Together with both of them, our method achieves the most accurate shape matching results.~\cref{fig:ablation} shows the qualitative comparison.

\begin{table}[hbt!]
    \centering
    \small
    \begin{tabular}{@{}lcc@{}}
    \toprule
    \multicolumn{1}{c}{\textbf{Geo.~error ($\times$100)}}  & \multicolumn{1}{c}{\textbf{SMAL}} & \multicolumn{1}{c}{\textbf{DT4D-H inter}}
    \\ \midrule
    \multicolumn{1}{l}{Spectral only}  & 3.9 & 4.1 \\
    \multicolumn{1}{l}{w.o. Test-time adaptation}  & 2.6 & 3.4  \\
    \multicolumn{1}{l}{Ours}  & \textbf{1.9} & \textbf{3.3}  \\ \hline
    \end{tabular} 
    \caption{Ablation study on the SMAL and DT4D-H datasets. The \textbf{best} result in each column is highlighted.}
    \vspace{-3.5mm}
    \label{tab:ablation}
\end{table}

%% file: figs/ablation.tex
\def\heightIQ{2.1cm}
\def\widthIQ{1.9cm}
\def\hspaceColsIQ{-0.44cm}
\def\pathIQ{figs/ours/}
\def\firstColIQ{001-027_M}
\def\secondColIQ{001-027-spectral_N}
\def\thirdColIQ{001-027_N}
\def\fourthColIQ{001-027_nn_N}
\begin{tabular}{ccccc}%
    \setlength{\tabcolsep}{0pt} 
    \hspace{\hspaceColsIQ}
    \includegraphics[height=\heightIQ, width=\widthIQ]{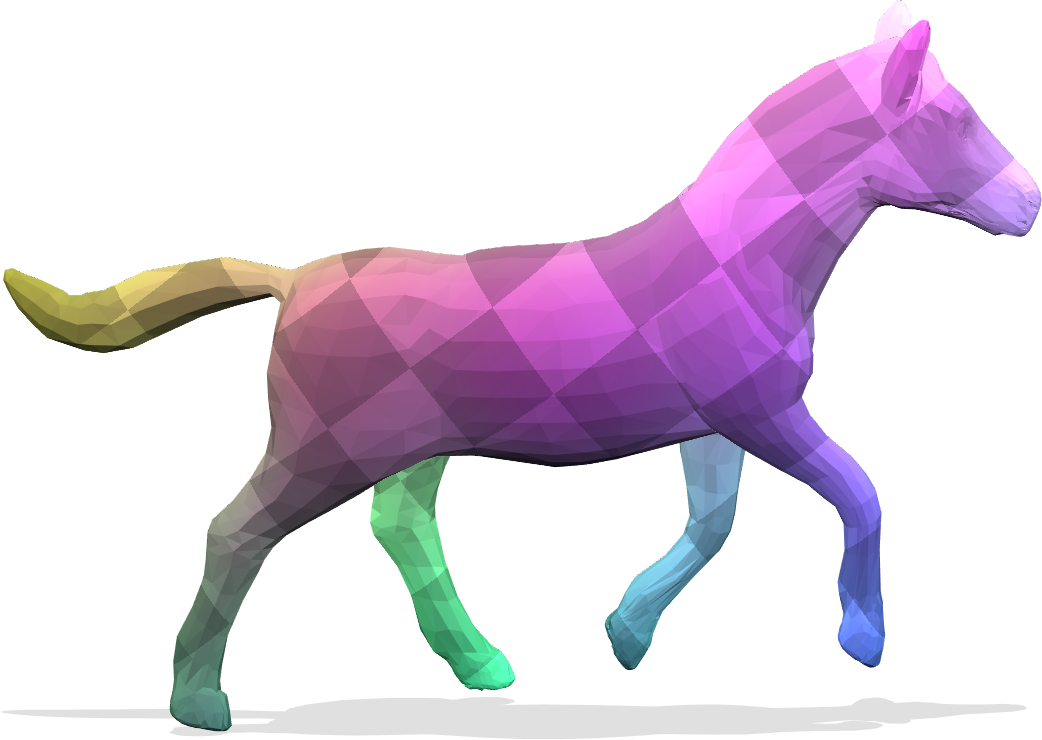}&
    \hspace{\hspaceColsIQ}
    \includegraphics[height=\heightIQ, width=\widthIQ]{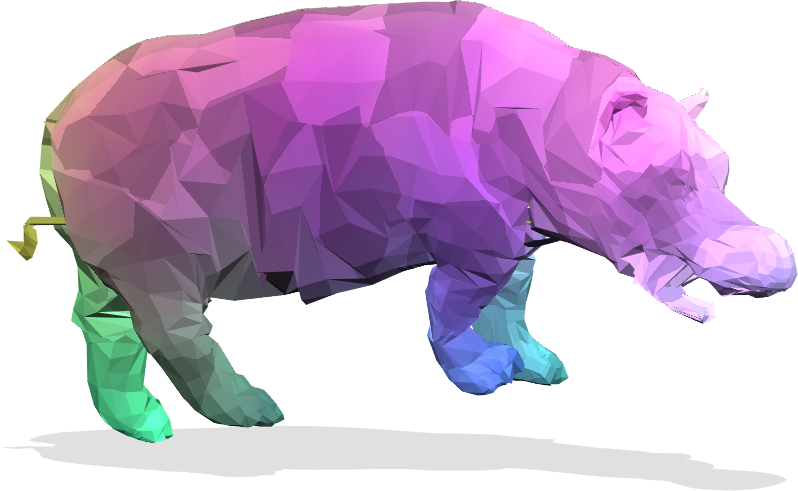}&
    \hspace{\hspaceColsIQ}
    \includegraphics[height=\heightIQ, width=\widthIQ]{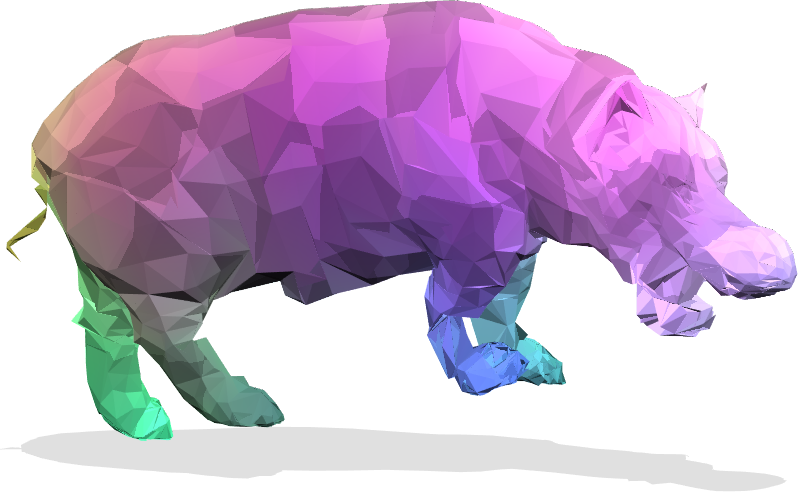}&
    \hspace{\hspaceColsIQ}
    \includegraphics[height=\heightIQ, width=\widthIQ]{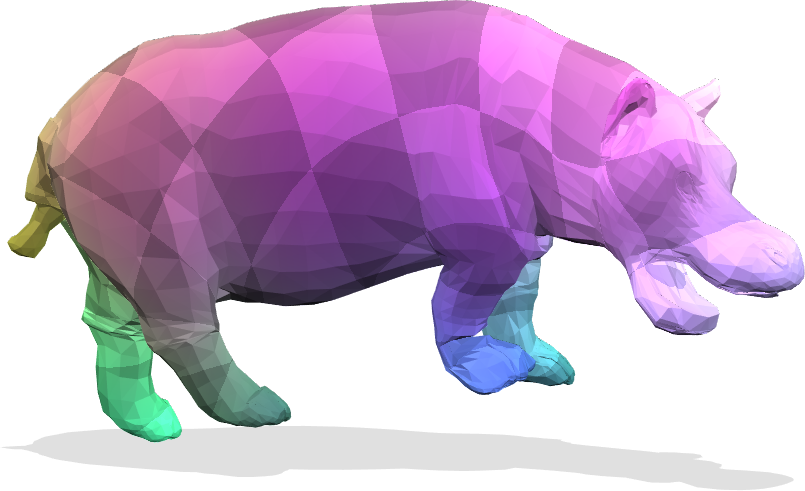}
    \\
    \hspace{\hspaceColsIQ} \textbf{\small{Source}}&
    \hspace{\hspaceColsIQ} \textbf{\small{Spec. only}}&
    \hspace{\hspaceColsIQ} \textbf{\small{w.o. Test-time adapt.}}&
    \hspace{\hspaceColsIQ} \textbf{\small{Ours}}
\end{tabular}

%% file: sec/7_conclusion.tex
\section{Discussion and limitations}
\label{sec:limitation}
For the first time we combine spectral and spatial maps to enable unsupervised learning for joint 3d shape matching and interpolation and thereby set the new state of the art in both tasks simultaneously. Yet, some limitations still give rise to future investigations.
Compared to prior works~\cite{eisenberger2019divergence,eisenberger2020hamiltonian}, our method captures pose and shape-dominant deformations without any prior knowledge. On the other hand, this separation between pose and shape is not always perfect: in practice, the interpolation module slightly alters the shape in some cases. In future work, we plan to investigate whether a clearer separation can be achieved by including additional assumptions about the input shape classes~\cite{zhou2020unsupervised}.
Another interesting direction is to integrate {end-to-end} statistical shape analysis into our framework for improved shape analysis and generation. 

\section{Conclusion}
In this work we propose the first unsupervised learning framework that harmonises spectral and spatial maps to enable joint 3D shape matching and interpolation. Our method allows to compute accurate shape correspondences and realistic interpolation trajectories between different 3D shapes without relying on any shape-specific prior knowledge.  We experimentally demonstrate that our method sets the new state of the art  in both shape matching and interpolation on numerous benchmarks. Moreover, we showcase the application of our method to real-world statistical shape models of medical data. Overall, we believe that our method will be a valuable contribution for bringing shape matching to practical applications in challenging real-world settings.
\label{sec:conclusion}

%% file: sec/8_acknowledgement.tex
\section{Acknowledgement}
This work is supported by the Deutsche Forschungsgemeinschaft (DFG, German Research Foundation) – 458610525, by the ERC Advanced Grant `SIMULACRON' and by the Collaborative Research Center `Discretization in Geometry and Dynamics' of the German Research Foundation.

%% file: sec/X_suppl.tex
\clearpage
\setcounter{page}{1}
\maketitlesupplementary

In this supplementary document we first provide implementation details of our method. Next, we provide more ablative experiments to demonstrate the advantages of our method. Afterwards, we explain the preparation process of the medical data for our statistical shape analysis. Eventually, we show more qualitative results of our method.

\section{Implementation details}
Firstly, we provide definitions of our spectral regularisation $L_{\mathrm{struct}}$ in~\cref{eq:l_struct}. The $L_{\mathrm{bij}}$ is the bijectivity loss to encourage the functional map from $\mathcal{X}$ through $\mathcal{Y}$ back to $\mathcal{X}$ is an identity map, and vice versa. It can be expressed in the form  
\begin{equation}
    \label{eq:bij}
    {L}_{\mathrm{bij}}=\left\|\mathbf{C}_{\mathcal{XY}}\mathbf{C}_{\mathcal{YX}}-\mathbf{I}\right\|^{2}_{F}+\left\|\mathbf{C}_{\mathcal{YX}}\mathbf{C}_{\mathcal{XY}}-\mathbf{I}\right\|^{2}_{F}.
\end{equation}
The ${L}_{\mathrm{orth}}$ is the orthogonality loss to prompt a locally area-preserving matching, see~\cite{roufosse2019unsupervised} for more details. It can be expressed in the form
\begin{equation}
    \label{eq:orth}    {L}_{\mathrm{orth}}=\left\|\mathbf{C}_{\mathcal{XY}}^{\top}\mathbf{C}_{\mathcal{YX}}-\mathbf{I}\right\|^{2}_{F}+\left\|\mathbf{C}_{\mathcal{XY}}^{\top}\mathbf{C}_{\mathcal{YX}}-\mathbf{I}\right\|^{2}_{F}.
\end{equation}
In the following we explain each component in our framework in detail. Following prior works~\cite{cao2023unsupervised,li2022learning}, we use DiffusionNet~\cite{sharp2020diffusionnet} as our feature extractor and wave kernel signature (WKS)~\cite{aubry2011wave} with 128 dimensions as input features, since it is agnostic to shape discretisation and orientation. The dimension of the output features is 256 for non-isometric datasets and 384 for near-isometric datasets. In the context of functional map computation, we use the regularised functional map solver~\cite{ren2019structured} based on the resolvent mask $\mathbf{M}$, in which the regularisation term $E_{\mathrm{reg}}$ in~\cref{eq:fmap} can be expressed in the form
    \begin{equation}
        \label{eq:fmap_reg}
        E_{\mathrm{reg}}=\sum_{i j} \mathbf{C}_{i j}^{2} \mathbf{M}_{i j},
    \end{equation}
where 
    \begin{multline}
        \label{eq:resolvent_mask}
        \mathbf{M}_{i j}=\left(\frac{\mathbf{\Lambda}_{\mathcal{Y}}(i)^{\gamma}}{\mathbf{\Lambda}_{\mathcal{Y}}(i)^{2 \gamma}+1}-\frac{\mathbf{\Lambda}_{\mathcal{X}}(j)^{\gamma}}{\mathbf{\Lambda}_{\mathcal{X}}(j)^{2 \gamma}+1}\right)^{2}\\
        +\left(\frac{1}{\mathbf{\Lambda}_{\mathcal{Y}}(i)^{2 \gamma}+1}-\frac{1}{\mathbf{\Lambda}_{\mathcal{X}}(j)^{2 \gamma}+1}\right)^{2}.
    \end{multline}
The regularisation strength $\lambda$ in~\cref{eq:fmap} is 100. The number of eigenfunctions $\mathbf{\Phi}$ and eigenvalues $\mathbf{\Lambda}$ used for functional map computation is 200 for non-isometric datasets and 300 for near-isometric datasets. In terms of our spectral loss $L_{\mathrm{spectral}}$, we empirically set $\lambda_{\mathrm{bij}} = 1, \lambda_{\mathrm{orth}} = 1$ in~\cref{eq:l_struct}, $\lambda_{\mathrm{struct}}=1, \lambda_{\mathrm{couple}}=1$ in~\cref{eq:l_fmap}. For our spatial loss $L_{\mathrm{spatial}}$, we empirically set $\lambda_{\mathrm{align}} = 5, \lambda_{\mathrm{arap}} = 100, \lambda_{\mathrm{sym}} = 1, \lambda_{\mathrm{var}} = 1$ in~\cref{eq:spatial}. We use Adam~\cite{kingma2015adam} optimiser with learning rate equal to $10^{-3}$. In the context of test-time adaptation, the shape-dominant deformation field $\mathbf{\Delta}_{s}(t)$ is initialised as all zeros. The Dirichlet energy in~\cref{eq:tta} can be expressed in the form
\begin{equation}
    L_{\mathrm{D}} = \left\|\mathbf{\Delta}\right\|_{\mathbf{L}}^{2},
\end{equation}
where $\left\|\mathbf{X}\right\|_{L}=\mathrm{Trace}(\mathbf{X}^{\mathrm{T}}\mathbf{L}\mathbf{X})$. We empirically set $\lambda_{\mathrm{D}}=0.1$. The number of iterations for optimisation is $2\times10^3$.

\section{Additional evaluations}
\subsection{Smoothness of point-wise maps} In~\cref{subsec:matching_results} we demonstrate that our method substantially outperforms existing state-of-the-art shape matching methods in terms of matching accuracy based on mean geodesic error. In this experiment, we evaluate the matching smoothness based on the conformal distortion metric~\cite{hormann2000mips} that is also used in~\cite{eisenberger2020smooth,ehm2023geometrically}. 
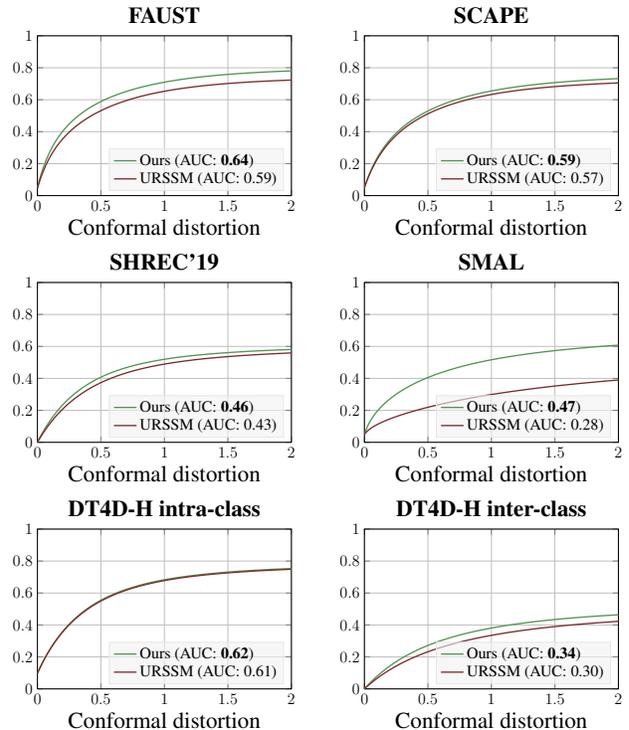
\begin{figure}[bht!]
    \centering
    \begin{tabular}{cc}
     \hspace{-1.2cm}
     \input{figs/tikz/faust_matching_conformal_pck}&
     \hspace{-1cm}
     \input{figs/tikz/scape_matching_conformal_pck} \\
     \hspace{-1.2cm}
     \input{figs/tikz/shrec19_matching_conformal_pck}&
     \hspace{-1cm}
     \input{figs/tikz/smal_matching_conformal_pck} \\
     \hspace{-1.2cm}
     \input{figs/tikz/dt4d_intra_matching_conformal_pck}&
     \hspace{-1cm}
     \input{figs/tikz/dt4d_inter_matching_conformal_pck}
    \end{tabular}
    \caption{{\textbf{Matching smoothness on all datasets in our matching experiments.} Our method obtains smoother point-wise correspondences based on the combination of spectral and spatial maps.}
    }
    \label{fig:matching_conformal}
\end{figure}
We compare our method with the state-of-the-art matching method (i.e.\ URSSM~\cite{cao2023unsupervised}) based on deep functional map framework. \cref{fig:matching_conformal} summarises the results on all datasets in our matching experiments. Our method obtains smoother point-wise correspondences by harmonising spectral and spatial maps.

\subsection{Matching on topologically noisy data}
We evaluate the performance of our method on topologically noisy data. Such topological noise presents a great challenge to shape matching methods, especially for matching methods with spatial regularisation~\cite{eisenberger2023g}. To this end, we use the TOPKIDS dataset~\cite{lahner2016shrec}, which contains synthetic shapes of children with topological merging. Due to the small amount of training data, we only consider axiomatic and unsupervised methods for comparison, similar to~\cite{cao2023unsupervised,eisenberger2023g}.~\cref{tab:topkids} summarises the matching results. We observe that our method obtains comparable but slightly worse performance compared to state-of-the-art, due to the incorporation of explicit spatial regularisation. Meanwhile, our method outperforms existing shape matching methods with explicit spatial regularisation. Moreover, our spatial regularisation ensures that our method does not suffer from symmetry flip as shown in~\cref{fig:topkids}.

\begin{table}[hbt!]
\setlength{\tabcolsep}{12pt}
    \centering
    \small
    \begin{tabular}{@{}lcc@{}}
    \toprule
    \multicolumn{1}{c}{\textbf{Geo. error ($\times$100)}}      & \multicolumn{1}{c}{\textbf{TOPKIDS}} & \multicolumn{1}{c}{\textbf{Spatial reg.}}
    \\ \midrule
    \multicolumn{3}{c}{Axiomatic Methods} \\
    \multicolumn{1}{l}{ZoomOut~\cite{melzi2019zoomout}}  & \multicolumn{1}{c}{33.7} & \multicolumn{1}{c}{\xmark}  \\ 
    \multicolumn{1}{l}{Smooth Shells~\cite{eisenberger2020smooth}}  & \multicolumn{1}{c}{11.8} & \multicolumn{1}{c}{\cmark}\\
    \multicolumn{1}{l}{DiscreteOp~\cite{ren2021discrete}}  & \multicolumn{1}{c}{35.5} & \multicolumn{1}{c}{\xmark}
    \\\midrule
    \multicolumn{3}{c}{Unsupervised Methods} \\
    \multicolumn{1}{l}{WSupFMNet~\cite{sharma2020weakly}}  & \multicolumn{1}{c}{47.9} & \multicolumn{1}{c}{\xmark}\\
    \multicolumn{1}{l}{Deep Shells~\cite{eisenberger2020deep}}  & \multicolumn{1}{c}{13.7} & \multicolumn{1}{c}{\cmark} \\
    \multicolumn{1}{l}{NeuroMorph~\cite{eisenberger2021neuromorph}}  & \multicolumn{1}{c}{13.8} & \multicolumn{1}{c}{\cmark} \\
    \multicolumn{1}{l}{AttnFMaps~\cite{li2022learning}}  & \multicolumn{1}{c}{23.4} & \multicolumn{1}{c}{\xmark} \\
    \multicolumn{1}{l}{URSSM~\cite{cao2023unsupervised}}  & \multicolumn{1}{c}{\textbf{9.2}} & \multicolumn{1}{c}{\xmark}
    \\ 
    \multicolumn{1}{l}{Ours}  & \multicolumn{1}{c}{{9.4}} & \multicolumn{1}{c}{\cmark}
    \\ 
    \hline
    \end{tabular}
    \caption{\textbf{Matching on TOPKIDS dataset.} Our method achieves comparable but slightly worse performance compared to the state-of-the-art method, while outperforming existing shape matching methods with explicit spatial regularisation.}
    \vspace{-4mm}
    \label{tab:topkids}
\end{table}
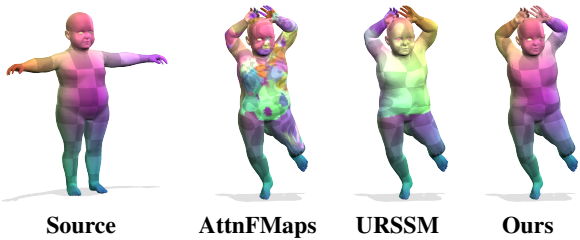
\begin{figure}[hbt!]
    \centering
    \input{figs/topkids_results}
    \caption{ 
    \textbf{Qualitative comparison on TOPKIDS dataset.} By incorporating spatial regularisation, our method does not suffer from front-back flips that occurs for functional map methods.
    }
    \vspace{-4mm}
    \label{fig:topkids}
\end{figure}

\section{Preparation of lung shapes}

To generate the lung meshes used in our statistical shape analysis experiment (see~\cref{subsec:medical_data}), we randomly select 22 CT images from the 9th subset of the LUNA dataset~\cite{setio2017validation}.  Each CT image is accompanied by segmentation masks of the lung structures. We first use the Marching Cubes algorithm~\cite{lorensen1998marching} to extract triangular meshes for the left lungs. After extraction, we remove outliers by clustering triangle meshes and retaining only the biggest triangle mesh in terms of the number of triangles. Next, we employ a Laplacian smoothing algorithm to smooth the triangle mesh. Finally, we apply the Quadric Error Metric Decimation~\cite{garland1997simplification}, resulting in simplified mesh structures comprising approximately 8000 triangles per mesh. 

\section{More qualitative results}
In this section we provide more qualitative results obtained by our method for shape matching and interpolation. 
\subsection{Shape matching}

\begin{figure}[!hbt]
    \centering
    \input{figs/shrec19_results}
    \caption{\textbf{Qualitative shape matching results of our method on the SHREC'19 dataset.} The top-left shape is the source shape that is matched to the other shapes. Our method obtains accurate matchings for human shapes with diverse poses and appearances.}
    \vspace{-2mm}
    \label{fig:shrec19_qualitative}
\end{figure}
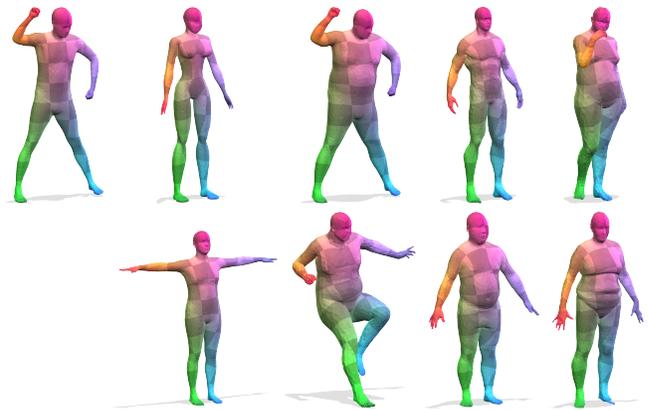

\begin{figure}[!hbt]
    \centering
    \input{figs/smal_results}
    \caption{\textbf{Qualitative shape matching results of our method on the SMAL dataset.} The top-left shape is the source shape that is matched to the other shapes. Our method obtains accurate correspondences for shapes in different classes.}
    \vspace{-2mm}
    \label{fig:smal_qualitative}
\end{figure}
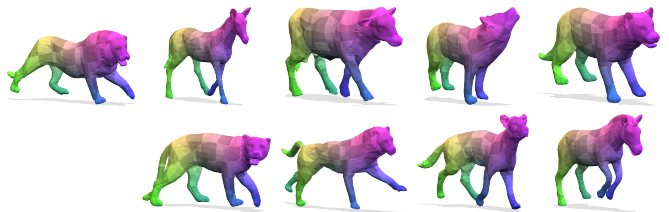

\begin{figure}[!bht]
    \centering
    \input{figs/dt4d_results}
    \caption{\textbf{Qualitative shape matching results of our method on the DT4D-H inter-class dataset.} The top-left shape is the source shape that is matched to the other shapes. Our method obtains accurate correspondences for non-isometric deformed shapes.}
    \vspace{-2mm}
    \label{fig:dt4d_qualitative}
\end{figure}
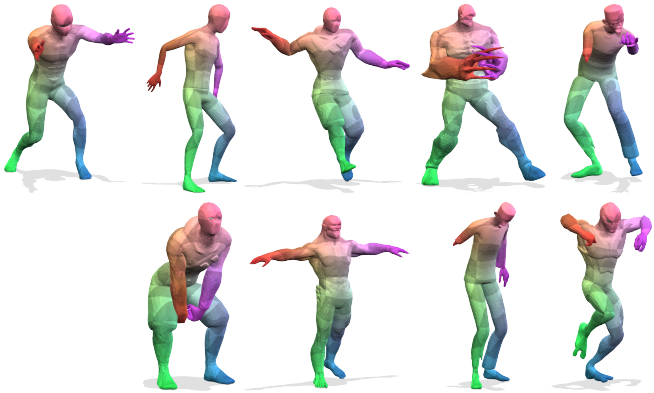

\subsection{Shape interpolation}
In this subsection we demonstrate the shape interpolation results of our method on different datasets. Additionally, we show the matching results by texture transfer from the source shape to the target shape. We observe that our method can obtain accurate point-wise correspondences and realistic shape interpolation trajectories even under large non-isometry and pose variations.
\begin{figure}[!bht]
    \centering
    \begin{tabular}{c}
         \includegraphics[width=\columnwidth]{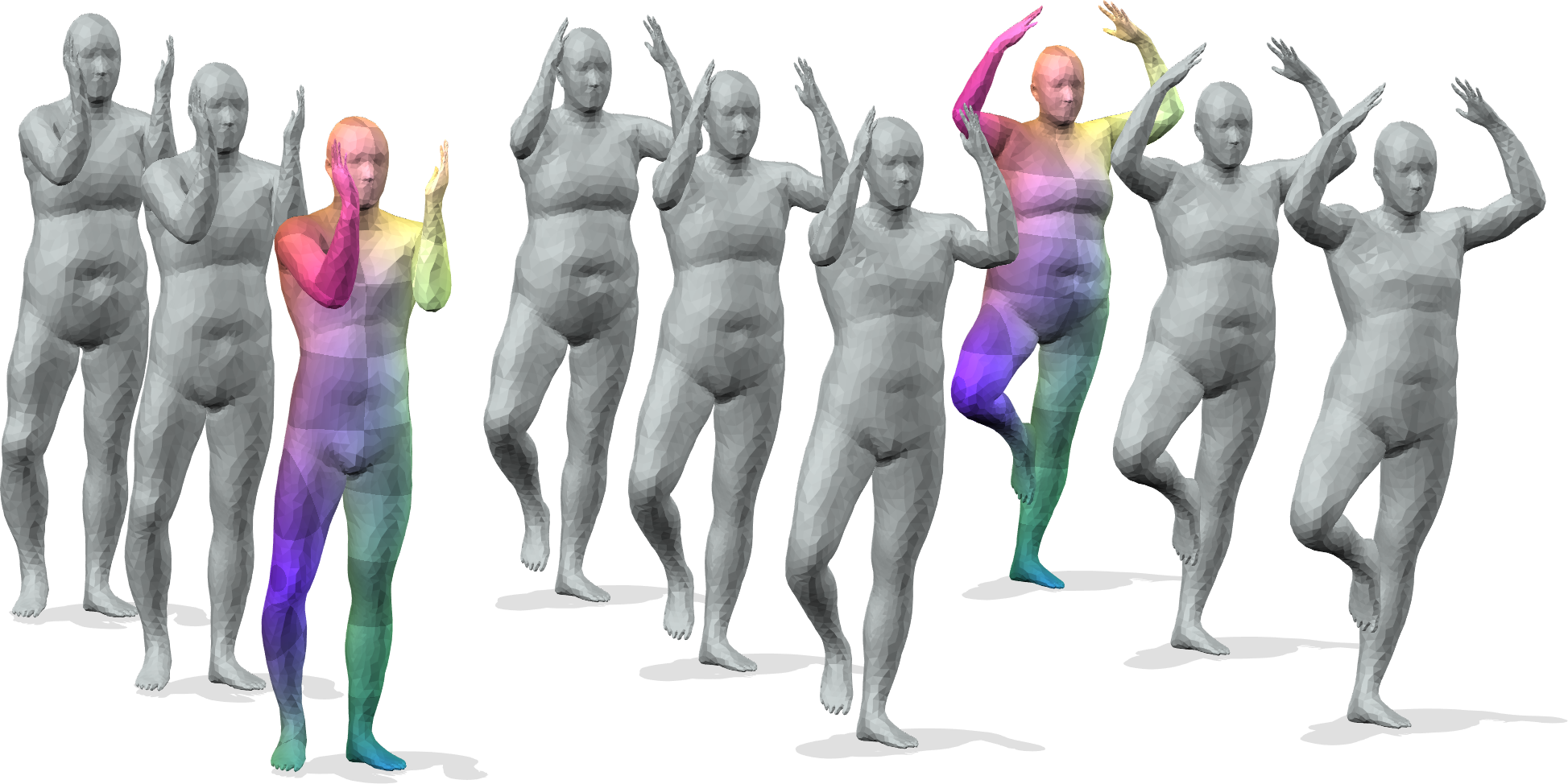} \\
         \includegraphics[width=\columnwidth]{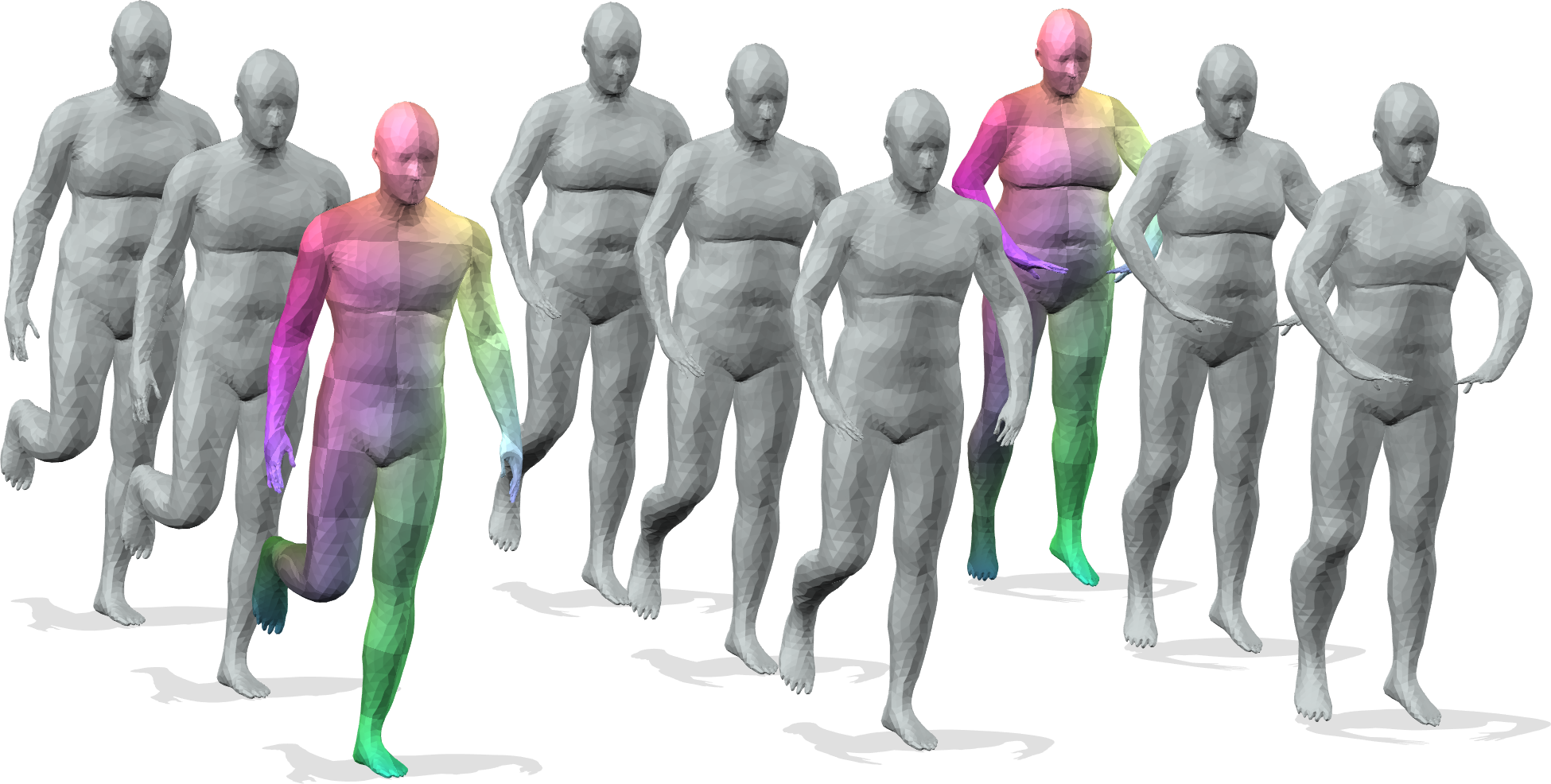} 
    \end{tabular}
    \caption{\textbf{Qualitative shape interpolation results of our method on the FAUST dataset.} Our method obtains realistic interpolation trajectories that capture both pose-dominant (horizontal) and shape-dominant (vertical) deformations.}
    \vspace{-4mm}
    \label{fig:faust_interp_result}
\end{figure}

\begin{figure}[!bht]
    \centering
    \begin{tabular}{c}
         \includegraphics[width=\columnwidth]{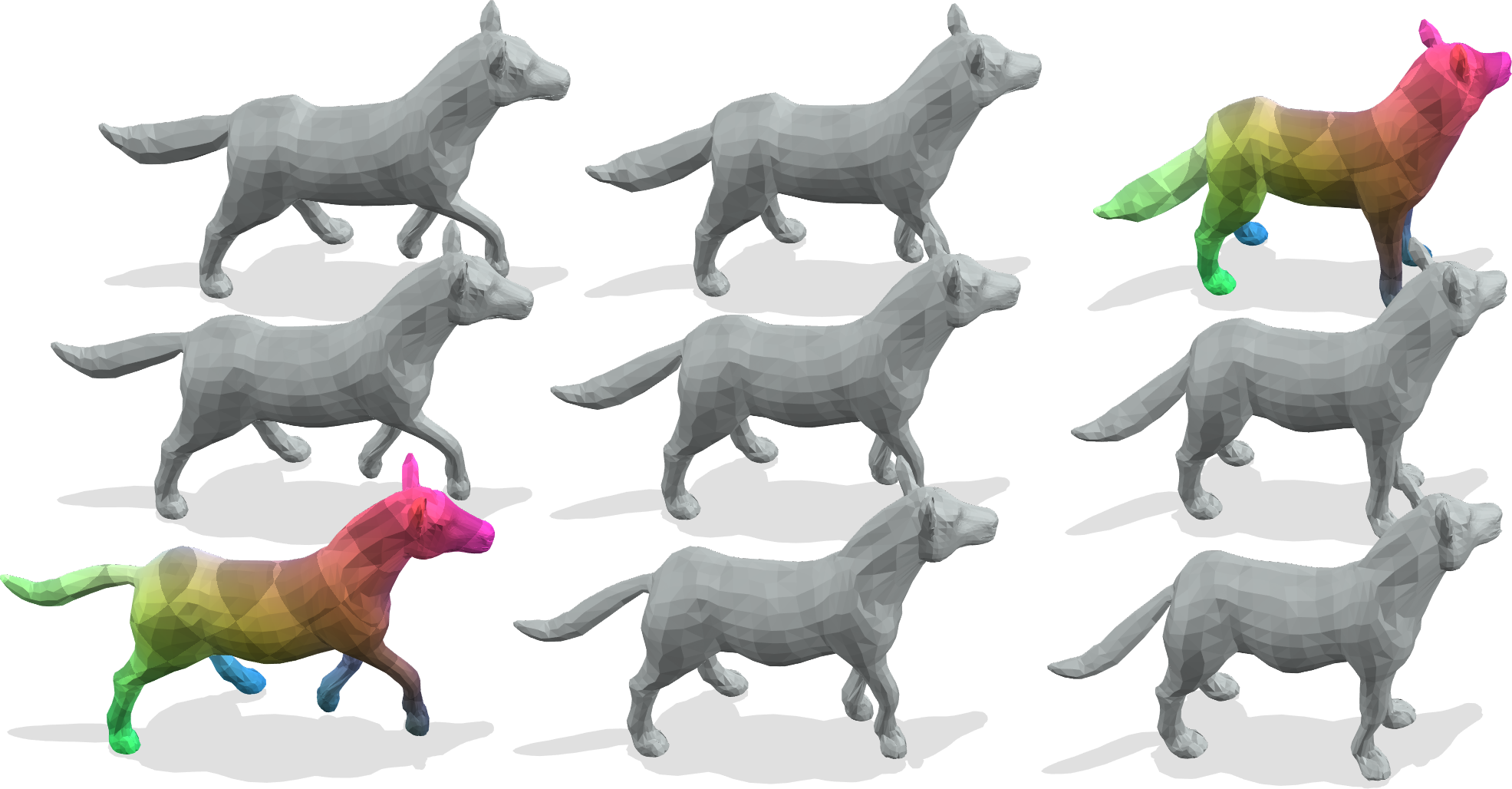} \\
         \includegraphics[width=\columnwidth]{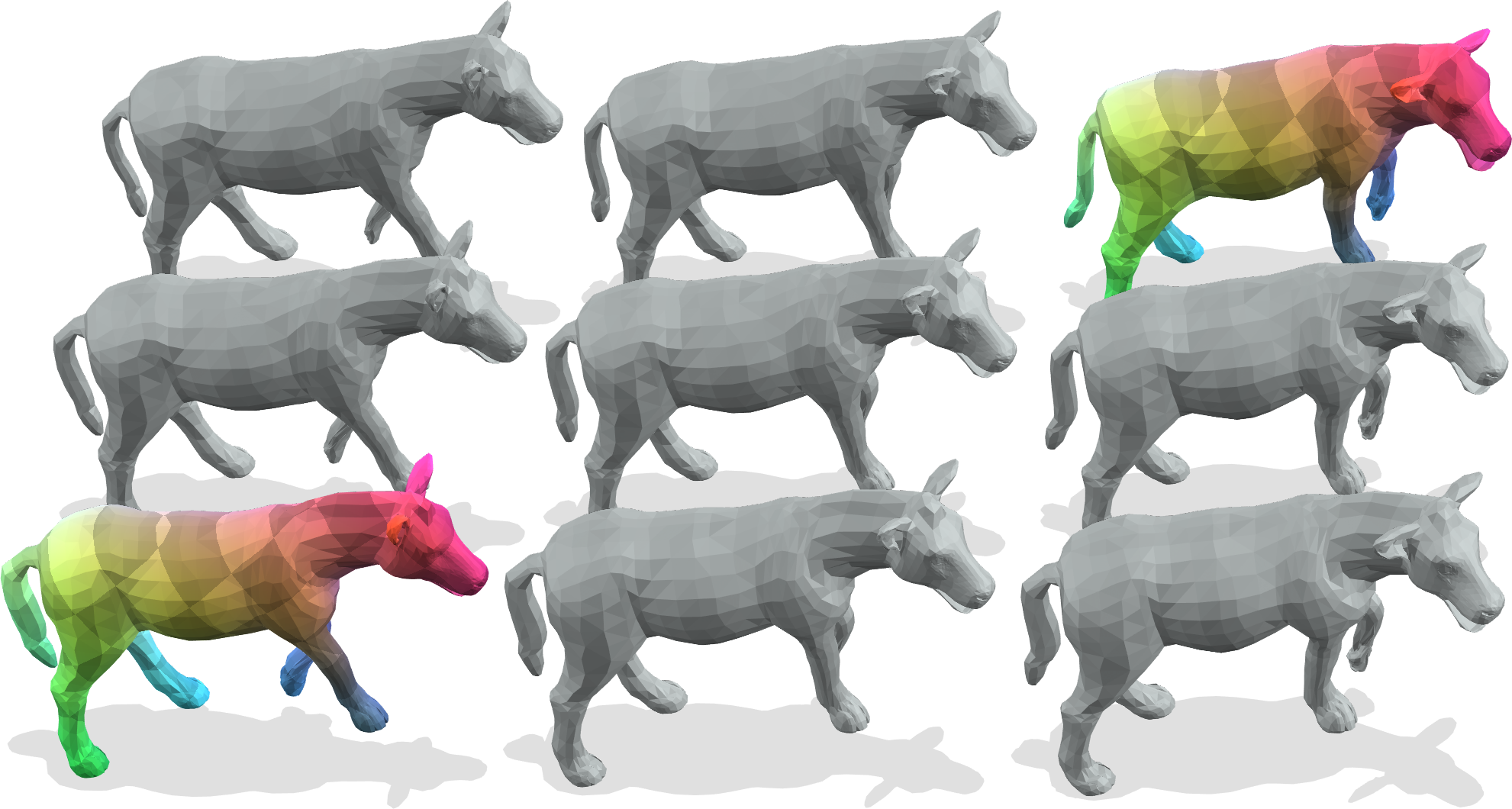} 
    \end{tabular}
    \caption{\textbf{Qualitative shape interpolation results of our method on the SMAL dataset.} Our method obtains realistic interpolation trajectories between different shape categories.}
    \vspace{-4mm}
    \label{fig:smal_interp_result}
\end{figure}

\begin{figure}[!bht]
    \centering
    \begin{tabular}{c}
         \includegraphics[width=\columnwidth]{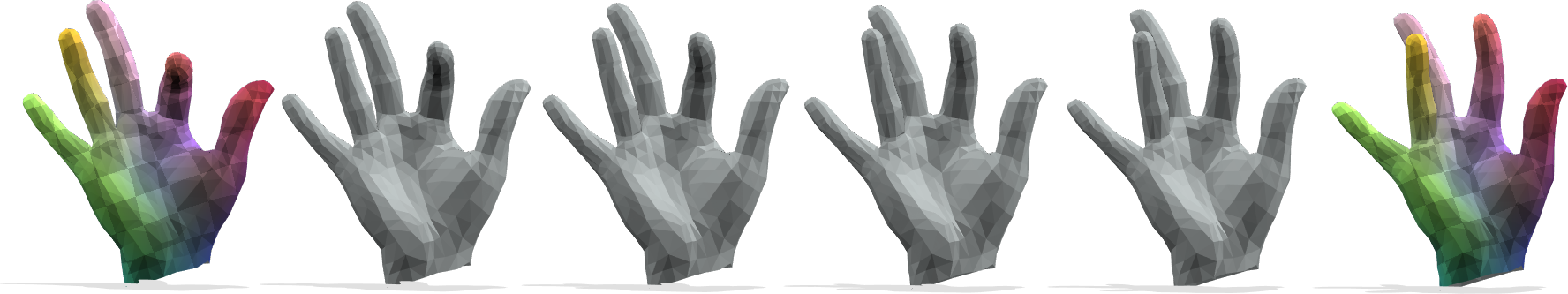} \\
         \includegraphics[width=\columnwidth]{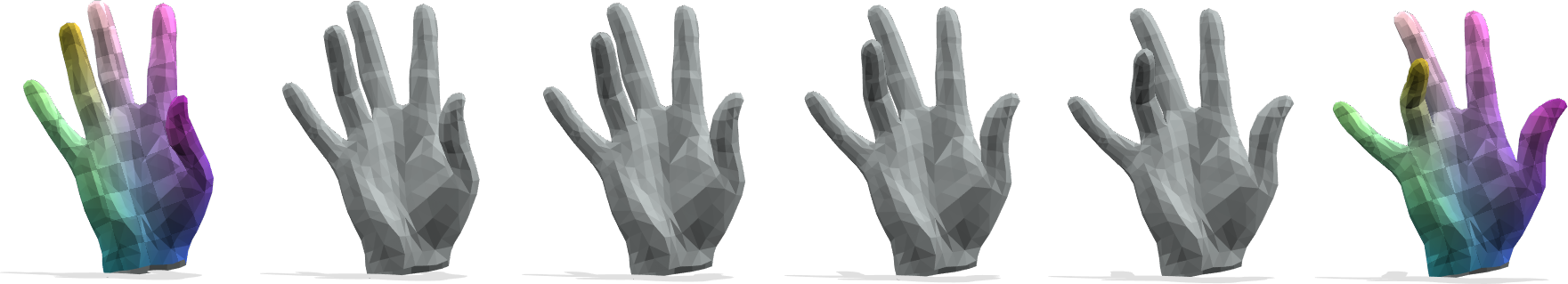} \\
    \end{tabular}
    \caption{\textbf{Qualitative shape interpolation results of our method on the MANO dataset.} Our method obtains realistic interpolation trajectories between hands in different poses.}
    \vspace{-4mm}
    \label{fig:mano_interp_result}
\end{figure}

\begin{figure}[!bht]
    \centering
    \begin{tabular}{c}
         \includegraphics[width=\columnwidth]{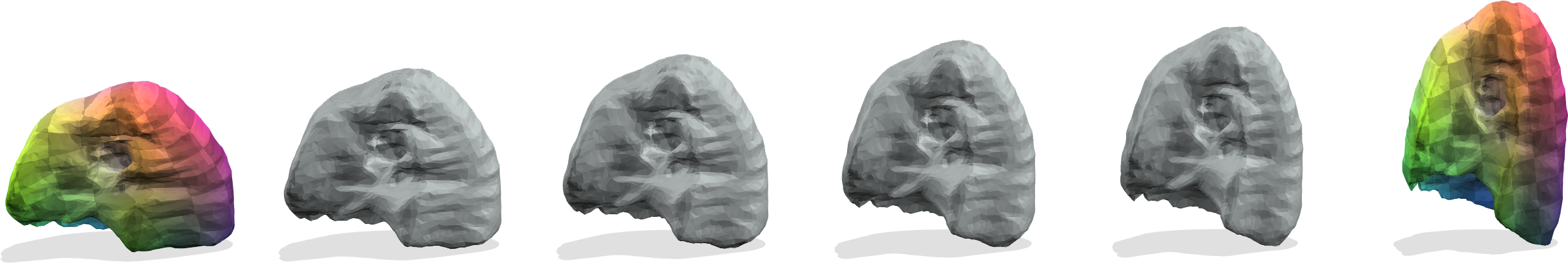} \\
        \includegraphics[width=\columnwidth]{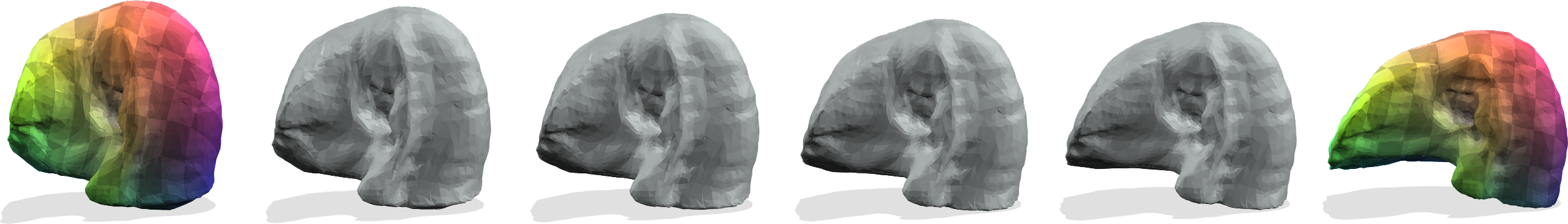} \\
        \includegraphics[width=\columnwidth]{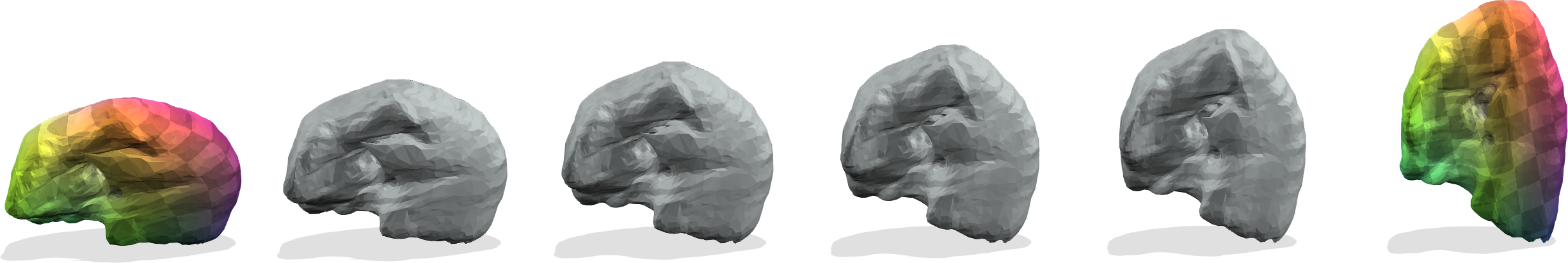} \\
    \end{tabular}
    \caption{\textbf{Qualitative shape interpolation results of our method on the LUNA dataset.} Our method obtains realistic interpolation trajectories between lungs despite large non-isometries.}
    \vspace{-4mm}
    \label{fig:luna_interp_result}
\end{figure}

%% file: figs/tikz/faust_matching_conformal_pck.tex
\newcommand{\pckLineWidth}{1pt}
\newcommand{\plotWidth}{\columnwidth}
\newcommand{\plotHeight}{0.7\columnwidth}
\newcommand{\pckTitle}{\textbf{FAUST}}
\definecolor{cPLOT0}{RGB}{28,213,227}
\definecolor{cPLOT1}{RGB}{80,150,80}
\definecolor{cPLOT2}{RGB}{90,130,213}
\definecolor{cPLOT3}{RGB}{247,179,43}
\definecolor{cPLOT4}{RGB}{124,42,43}
\definecolor{cPLOT5}{RGB}{242,64,0}

\pgfplotsset{%
    label style = {font=\LARGE},
    tick label style = {font=\large},
    title style =  {font=\LARGE},
    legend style={  fill= gray!10,
                    fill opacity=0.6, 
                    font=\large,
                    draw=gray!20, %
                    text opacity=1}
}
\begin{tikzpicture}[scale=0.5, transform shape]
	\begin{axis}[
		width=\plotWidth,
		height=\plotHeight,
		grid=major,
		title=\pckTitle,
		legend style={
			at={(0.97,0.03)},
			anchor=south east,
			legend columns=1},
		legend cell align={left},
        xlabel={\LARGE Conformal distortion},
		xmin=0,
        xmax=2.0,
        ylabel near ticks,
        xtick={0, 0.5, 1.0, 1.5, 2.0},
	ymin=0,
        ymax=1,
        ytick={0, 0.2, 0.40, 0.60, 0.80, 1.0}
	]

\addplot [color=cPLOT1, smooth, line width=\pckLineWidth]
table[row sep=crcr]{%
0.0  0.04998374479833547 \\
0.02531645569620253  0.12370633602752881 \\
0.05063291139240506  0.1844602772887324 \\
0.0759493670886076  0.2358764804737516 \\
0.10126582278481013  0.2797485195262484 \\
0.12658227848101267  0.31748259443021765 \\
0.1518987341772152  0.35052591829385404 \\
0.17721518987341772  0.3796257302336748 \\
0.20253164556962025  0.40574083706786174 \\
0.22784810126582278  0.4291498279449424 \\
0.25316455696202533  0.4503853733194622 \\
0.27848101265822783  0.4698788612355954 \\
0.3037974683544304  0.48764854753521125 \\
0.3291139240506329  0.5040347911331626 \\
0.35443037974683544  0.519238906450064 \\
0.37974683544303794  0.5332884022887324 \\
0.4050632911392405  0.5463935959507042 \\
0.43037974683544306  0.558543984074904 \\
0.45569620253164556  0.5699701404449424 \\
0.4810126582278481  0.5807403369078105 \\
0.5063291139240507  0.5908415693021767 \\
0.5316455696202531  0.6004268866037132 \\
0.5569620253164557  0.6094020086427657 \\
0.5822784810126582  0.6178764704705506 \\
0.6075949367088608  0.6258770306498079 \\
0.6329113924050633  0.6333859334987196 \\
0.6582278481012658  0.6406074943982074 \\
0.6835443037974683  0.6474024187740077 \\
0.7088607594936709  0.6538722391165173 \\
0.7341772151898734  0.6600887283930857 \\
0.7594936708860759  0.6659386003521127 \\
0.7848101265822784  0.6715256382042254 \\
0.810126582278481  0.6768856033930858 \\
0.8354430379746836  0.6819824843950064 \\
0.8607594936708861  0.6868380381722151 \\
0.8860759493670886  0.691350732234315 \\
0.9113924050632911  0.6957999059699104 \\
0.9367088607594937  0.7000217569622279 \\
0.9620253164556962  0.7040265384923176 \\
0.9873417721518987  0.7078605153649168 \\
1.0126582278481013  0.7115499459827145 \\
1.0379746835443038  0.715057568421895 \\
1.0632911392405062  0.7184191441261204 \\
1.0886075949367089  0.7216086647727272 \\
1.1139240506329113  0.7247071562900128 \\
1.139240506329114  0.7276591009122919 \\
1.1645569620253164  0.7304822543213828 \\
1.1898734177215189  0.7331926216389244 \\
1.2151898734177216  0.7357811999839948 \\
1.240506329113924  0.7382724971991037 \\
1.2658227848101267  0.7406312520006402 \\
1.2911392405063291  0.7429182338348271 \\
1.3164556962025316  0.7451014324583867 \\
1.3417721518987342  0.7471988536331626 \\
1.3670886075949367  0.7491987435979514 \\
1.3924050632911391  0.7511316121158771 \\
1.4177215189873418  0.7529601972631242 \\
1.4430379746835442  0.7547237616037132 \\
1.4683544303797469  0.7564785731434059 \\
1.4936708860759493  0.7580833366677336 \\
1.5189873417721518  0.7596483374679898 \\
1.5443037974683544  0.7611490677016646 \\
1.5696202531645569  0.7626147867317542 \\
1.5949367088607596  0.7639809739116518 \\
1.620253164556962  0.7653224031690141 \\
1.6455696202531644  0.7666193181818182 \\
1.6708860759493671  0.7678737195902688 \\
1.6962025316455696  0.7690801056338028 \\
1.7215189873417722  0.7702304737516005 \\
1.7468354430379747  0.7713513324263764 \\
1.7721518987341771  0.7724261763764405 \\
1.7974683544303798  0.7734767625640205 \\
1.8227848101265822  0.7744878361075545 \\
1.8481012658227849  0.7754453925256082 \\
1.8734177215189873  0.7763929457426376 \\
1.8987341772151898  0.7772889824743918 \\
1.9240506329113924  0.7781265004801536 \\
1.9493670886075949  0.778929257362356 \\
1.9746835443037973  0.7797355153649168 \\
2.0  0.7804972591229193 \\
    };
\addlegendentry{\textcolor{black}{Ours (AUC: \textbf{0.64})}}

\addplot [color=cPLOT4, smooth, line width=\pckLineWidth]
table[row sep=crcr]{%
0.0  0.04663417293533931 \\
0.02531645569620253  0.10902788892445582 \\
0.05063291139240506  0.1614441621318822 \\
0.0759493670886076  0.20602917933738796 \\
0.10126582278481013  0.2445287491997439 \\
0.12658227848101267  0.2781372539212548 \\
0.1518987341772152  0.30771271806978234 \\
0.17721518987341772  0.3340976612516005 \\
0.20253164556962025  0.35796554897567223 \\
0.22784810126582278  0.3794841849391805 \\
0.25316455696202533  0.39912647047055055 \\
0.27848101265822783  0.41729878561139566 \\
0.3037974683544304  0.4341164172535211 \\
0.3291139240506329  0.4496301316421255 \\
0.35443037974683544  0.4641087648047375 \\
0.37974683544303794  0.4776158370678617 \\
0.4050632911392405  0.4902556318021767 \\
0.43037974683544306  0.5021746959026888 \\
0.45569620253164556  0.5134388004161332 \\
0.4810126582278481  0.5239676696542894 \\
0.5063291139240507  0.5338648367477593 \\
0.5316455696202531  0.5432763484314981 \\
0.5569620253164557  0.5522012043854033 \\
0.5822784810126582  0.5606729153329065 \\
0.6075949367088608  0.568730243677977 \\
0.6329113924050633  0.5763334266965429 \\
0.6582278481012658  0.5835899987996159 \\
0.6835443037974683  0.5903716689340589 \\
0.7088607594936709  0.5969577764884764 \\
0.7341772151898734  0.6031114956786171 \\
0.7594936708860759  0.6090001300416134 \\
0.7848101265822784  0.6146321822983355 \\
0.810126582278481  0.6199926476472472 \\
0.8354430379746836  0.6251117857714469 \\
0.8607594936708861  0.6299780929897567 \\
0.8860759493670886  0.634587568021767 \\
0.9113924050632911  0.6389889764724712 \\
0.9367088607594937  0.6431733154609475 \\
0.9620253164556962  0.6471993537932138 \\
0.9873417721518987  0.6510508362676056 \\
1.0126582278481013  0.6547340148847631 \\
1.0379746835443038  0.6582336347631242 \\
1.0632911392405062  0.6616059639084507 \\
1.0886075949367089  0.6648132402368758 \\
1.1139240506329113  0.667881972231114 \\
1.139240506329114  0.6708329165332907 \\
1.1645569620253164  0.6736695742637644 \\
1.1898734177215189  0.6763604353393086 \\
1.2151898734177216  0.678945762644046 \\
1.240506329113924  0.6814665693021768 \\
1.2658227848101267  0.683830575784251 \\
1.2911392405063291  0.6861113056177977 \\
1.3164556962025316  0.6882779989596671 \\
1.3417721518987342  0.6903654169334187 \\
1.3670886075949367  0.6923270446542894 \\
1.3924050632911391  0.6942326544494238 \\
1.4177215189873418  0.6960662411971831 \\
1.4430379746835442  0.6977917933738796 \\
1.4683544303797469  0.6995058418693982 \\
1.4936708860759493  0.7011226092349552 \\
1.5189873417721518  0.7026908610755441 \\
1.5443037974683544  0.704188340268886 \\
1.5696202531645569  0.7056190480953906 \\
1.5949367088607596  0.7069752320742637 \\
1.620253164556962  0.7083196622919334 \\
1.6455696202531644  0.7096175776248399 \\
1.6708860759493671  0.7108279649487836 \\
1.6962025316455696  0.7120348511523688 \\
1.7215189873417722  0.7131437059859155 \\
1.7468354430379747  0.7142213008162612 \\
1.7721518987341771  0.7152501300416133 \\
1.7974683544303798  0.7162486995838668 \\
1.8227848101265822  0.7172032550416133 \\
1.8481012658227849  0.7181298015364916 \\
1.8734177215189873  0.7190225872279129 \\
1.8987341772151898  0.7198828625160051 \\
1.9240506329113924  0.7207263824423815 \\
1.9493670886075949  0.7215071322823303 \\
1.9746835443037973  0.7222753781209987 \\
2.0  0.7230043613956466 \\
    };
\addlegendentry{\textcolor{black}{URSSM (AUC: {0.59})}}

\end{axis}
\end{tikzpicture}

%% file: figs/tikz/scape_matching_conformal_pck.tex
\newcommand{\pckLineWidth}{1pt}
\newcommand{\plotWidth}{\columnwidth}
\newcommand{\plotHeight}{0.7\columnwidth}
\newcommand{\pckTitle}{\textbf{SCAPE}}
\definecolor{cPLOT0}{RGB}{28,213,227}
\definecolor{cPLOT1}{RGB}{80,150,80}
\definecolor{cPLOT2}{RGB}{90,130,213}
\definecolor{cPLOT3}{RGB}{247,179,43}
\definecolor{cPLOT4}{RGB}{124,42,43}
\definecolor{cPLOT5}{RGB}{242,64,0}

\pgfplotsset{%
    label style = {font=\LARGE},
    tick label style = {font=\large},
    title style =  {font=\LARGE},
    legend style={  fill= gray!10,
                    fill opacity=0.6, 
                    font=\large,
                    draw=gray!20, %
                    text opacity=1}
}
\begin{tikzpicture}[scale=0.5, transform shape]
	\begin{axis}[
		width=\plotWidth,
		height=\plotHeight,
		grid=major,
		title=\pckTitle,
		legend style={
			at={(0.97,0.03)},
			anchor=south east,
			legend columns=1},
		legend cell align={left},
        xlabel={\LARGE Conformal distortion},
		xmin=0,
        xmax=2.0,
        ylabel near ticks,
        xtick={0, 0.5, 1.0, 1.5, 2.0},
	ymin=0,
        ymax=1,
        ytick={0, 0.2, 0.40, 0.60, 0.80, 1.0}
	]

\addplot [color=cPLOT1, smooth, line width=\pckLineWidth]
table[row sep=crcr]{%
0.0  0.04997574587679906 \\
0.02531645569620253  0.10739150655611454 \\
0.05063291139240506  0.1567286438694578 \\
0.0759493670886076  0.199536921276617 \\
0.10126582278481013  0.23702829481011772 \\
0.12658227848101267  0.2703544602582439 \\
0.1518987341772152  0.3001717792024644 \\
0.17721518987341772  0.3269570827040597 \\
0.20253164556962025  0.3511589470209936 \\
0.22784810126582278  0.3730849244371543 \\
0.25316455696202533  0.393481141794105 \\
0.27848101265822783  0.4121585669563826 \\
0.3037974683544304  0.4294207515277597 \\
0.3291139240506329  0.4454467259434104 \\
0.35443037974683544  0.4603502595441225 \\
0.37974683544303794  0.4742676254963344 \\
0.4050632911392405  0.48727433663722836 \\
0.43037974683544306  0.4994444055489433 \\
0.45569620253164556  0.5108873508496444 \\
0.4810126582278481  0.5217114409449607 \\
0.5063291139240507  0.5320009401598271 \\
0.5316455696202531  0.5415818189092145 \\
0.5569620253164557  0.5507011191902623 \\
0.5822784810126582  0.5593730934258824 \\
0.6075949367088608  0.5675379814568476 \\
0.6329113924050633  0.5753263054719302 \\
0.6582278481012658  0.5827473170438975 \\
0.6835443037974683  0.5898282708060371 \\
0.7088607594936709  0.5965074062590641 \\
0.7341772151898734  0.6029037436364182 \\
0.7594936708860759  0.6089370192932798 \\
0.7848101265822784  0.6147827630697219 \\
0.810126582278481  0.620353460088215 \\
0.8354430379746836  0.6256773651520758 \\
0.8607594936708861  0.6307202224378144 \\
0.8860759493670886  0.6355520438474541 \\
0.9113924050632911  0.6402078353320064 \\
0.9367088607594937  0.6445758278907414 \\
0.9620253164556962  0.6487785423521999 \\
0.9873417721518987  0.6528247302041347 \\
1.0126582278481013  0.6567466469299781 \\
1.0379746835443038  0.6604817819029235 \\
1.0632911392405062  0.6640316353780142 \\
1.0886075949367089  0.6674469659842173 \\
1.1139240506329113  0.6707055199383896 \\
1.139240506329114  0.6738238000460078 \\
1.1645569620253164  0.6767898042667253 \\
1.1898734177215189  0.6796855465429122 \\
1.2151898734177216  0.6824015082564036 \\
1.240506329113924  0.685072212276087 \\
1.2658227848101267  0.6876516507806327 \\
1.2911392405063291  0.6901628276807057 \\
1.3164556962025316  0.6925154776311973 \\
1.3417721518987342  0.6947806127041597 \\
1.3670886075949367  0.6969982397007491 \\
1.3924050632911391  0.6991160997369553 \\
1.4177215189873418  0.7011081883920266 \\
1.4430379746835442  0.7030522688857106 \\
1.4683544303797469  0.7049313383275156 \\
1.4936708860759493  0.7067338947621096 \\
1.5189873417721518  0.7084791914625487 \\
1.5443037974683544  0.7101937329345989 \\
1.5696202531645569  0.7118040066811357 \\
1.5949367088607596  0.7133520198433734 \\
1.620253164556962  0.7148532750567597 \\
1.6455696202531644  0.7163340267845534 \\
1.6708860759493671  0.7177445165678166 \\
1.6962025316455696  0.7191137493373874 \\
1.7215189873417722  0.7204189712251082 \\
1.7468354430379747  0.7216611824010082 \\
1.7721518987341771  0.7228843903463589 \\
1.7974683544303798  0.7240363361771501 \\
1.8227848101265822  0.7251660282247983 \\
1.8481012658227849  0.726282217977056 \\
1.8734177215189873  0.7273534000780133 \\
1.8987341772151898  0.7283833251652782 \\
1.9240506329113924  0.7294079993598912 \\
1.9493670886075949  0.7303801646279867 \\
1.9746835443037973  0.7313325765380114 \\
2.0  0.7322232279487513 \\
    };
\addlegendentry{\textcolor{black}{Ours (AUC: \textbf{0.59})}}

\addplot [color=cPLOT4, smooth, line width=\pckLineWidth]
table[row sep=crcr]{%
0.0  0.04974045587749917 \\
0.02531645569620253  0.10436299170859047 \\
0.05063291139240506  0.15130922256783652 \\
0.0759493670886076  0.19219742356200553 \\
0.10126582278481013  0.22826255463428782 \\
0.12658227848101267  0.2602272386305672 \\
0.1518987341772152  0.288897612594141 \\
0.17721518987341772  0.31481126791554565 \\
0.20253164556962025  0.3384250322554834 \\
0.22784810126582278  0.3598019163257754 \\
0.25316455696202533  0.37952226878569356 \\
0.27848101265822783  0.3976158446935979 \\
0.3037974683544304  0.4144287028794895 \\
0.3291139240506329  0.4299798465739176 \\
0.35443037974683544  0.44451381734894935 \\
0.37974683544303794  0.4580858745986818 \\
0.4050632911392405  0.47070476981086784 \\
0.43037974683544306  0.4826287968954722 \\
0.45569620253164556  0.4938242001140194 \\
0.4810126582278481  0.5044307532280488 \\
0.5063291139240507  0.5144332036446195 \\
0.5316455696202531  0.5238590560395268 \\
0.5569620253164557  0.5327243131332327 \\
0.5822784810126582  0.5411392436714242 \\
0.6075949367088608  0.5490480881749897 \\
0.6329113924050633  0.5566378784393347 \\
0.6582278481012658  0.5638053469089745 \\
0.6835443037974683  0.5705834991948632 \\
0.7088607594936709  0.5771083584209316 \\
0.7341772151898734  0.5832396507406259 \\
0.7594936708860759  0.5891026474500665 \\
0.7848101265822784  0.5947020993568907 \\
0.810126582278481  0.5999932488523049 \\
0.8354430379746836  0.6050673614514468 \\
0.8607594936708861  0.6098594261024374 \\
0.8860759493670886  0.614467959553124 \\
0.9113924050632911  0.6189097146514908 \\
0.9367088607594937  0.6230731724393147 \\
0.9620253164556962  0.6271176099936989 \\
0.9873417721518987  0.630990518388126 \\
1.0126582278481013  0.6346901473250453 \\
1.0379746835443038  0.6382087454867328 \\
1.0632911392405062  0.6415923206945181 \\
1.0886075949367089  0.644811868017563 \\
1.1139240506329113  0.6479449006331076 \\
1.139240506329114  0.6508706480101617 \\
1.1645569620253164  0.653704879829571 \\
1.1898734177215189  0.6564220917555984 \\
1.2151898734177216  0.659074292629747 \\
1.240506329113924  0.6616087234829922 \\
1.2658227848101267  0.6640351359731155 \\
1.2911392405063291  0.6663375273796546 \\
1.3164556962025316  0.66855190382365 \\
1.3417721518987342  0.6706760149225368 \\
1.3670886075949367  0.6727368652670954 \\
1.3924050632911391  0.6747102007341248 \\
1.4177215189873418  0.6765437624396148 \\
1.4430379746835442  0.6783373173439484 \\
1.4683544303797469  0.6800473580508687 \\
1.4936708860759493  0.6817316443795445 \\
1.5189873417721518  0.6833424182110959 \\
1.5443037974683544  0.6848794295030155 \\
1.5696202531645569  0.6863759339087645 \\
1.5949367088607596  0.6878131782403009 \\
1.620253164556962  0.6892071652180871 \\
1.6455696202531644  0.6905656461598472 \\
1.6708860759493671  0.6918388626066432 \\
1.6962025316455696  0.6931083284158307 \\
1.7215189873417722  0.6943175339807768 \\
1.7468354430379747  0.6954739805766981 \\
1.7721518987341771  0.6965964213916366 \\
1.7974683544303798  0.6976873568506646 \\
1.8227848101265822  0.6987242831281317 \\
1.8481012658227849  0.6997409559625136 \\
1.8734177215189873  0.7007021193602913 \\
1.8987341772151898  0.7016487802926498 \\
1.9240506329113924  0.7025856895672264 \\
1.9493670886075949  0.7034885930608203 \\
1.9746835443037973  0.7043327365652161 \\
2.0  0.7051258713981376 \\
    };
\addlegendentry{\textcolor{black}{URSSM (AUC: {0.57})}}

\end{axis}
\end{tikzpicture}

%% file: figs/tikz/shrec19_matching_conformal_pck.tex
\newcommand{\pckLineWidth}{1pt}
\newcommand{\plotWidth}{\columnwidth}
\newcommand{\plotHeight}{0.7\columnwidth}
\newcommand{\pckTitle}{\textbf{SHREC'19}}
\definecolor{cPLOT0}{RGB}{28,213,227}
\definecolor{cPLOT1}{RGB}{80,150,80}
\definecolor{cPLOT2}{RGB}{90,130,213}
\definecolor{cPLOT3}{RGB}{247,179,43}
\definecolor{cPLOT4}{RGB}{124,42,43}
\definecolor{cPLOT5}{RGB}{242,64,0}

\pgfplotsset{%
    label style = {font=\LARGE},
    tick label style = {font=\large},
    title style =  {font=\LARGE},
    legend style={  fill= gray!10,
                    fill opacity=0.6, 
                    font=\large,
                    draw=gray!20, %
                    text opacity=1}
}
\begin{tikzpicture}[scale=0.5, transform shape]
	\begin{axis}[
		width=\plotWidth,
		height=\plotHeight,
		grid=major,
		title=\pckTitle,
		legend style={
			at={(0.97,0.03)},
			anchor=south east,
			legend columns=1},
		legend cell align={left},
        xlabel={\LARGE Conformal distortion},
		xmin=0,
        xmax=2.0,
        ylabel near ticks,
        xtick={0, 0.5, 1.0, 1.5, 2.0},
	ymin=0,
        ymax=1,
        ytick={0, 0.2, 0.40, 0.60, 0.80, 1.0}
	]

\addplot [color=cPLOT1, smooth, line width=\pckLineWidth]
table[row sep=crcr]{%
0.0  0.0 \\
0.02531645569620253  0.03887228649252656 \\
0.05063291139240506  0.07440579826552862 \\
0.0759493670886076  0.1070973472651719 \\
0.10126582278481013  0.13744091056203772 \\
0.12658227848101267  0.16535285932638633 \\
0.1518987341772152  0.19114999090477416 \\
0.17721518987341772  0.2149920032884557 \\
0.20253164556962025  0.237186480478338 \\
0.22784810126582278  0.25771924809649915 \\
0.25316455696202533  0.27679961445692053 \\
0.27848101265822783  0.2944663228592555 \\
0.3037974683544304  0.311025067387355 \\
0.3291139240506329  0.3265590043916948 \\
0.35443037974683544  0.3410305717707814 \\
0.37974683544303794  0.3543436199943775 \\
0.4050632911392405  0.3668551071464851 \\
0.43037974683544306  0.3787086669233804 \\
0.45569620253164556  0.3897904318224234 \\
0.4810126582278481  0.4001117413459517 \\
0.5063291139240507  0.4097933139459342 \\
0.5316455696202531  0.41893862258120146 \\
0.5569620253164557  0.4275285318415588 \\
0.5822784810126582  0.43560461990224403 \\
0.6075949367088608  0.4431257810672834 \\
0.6329113924050633  0.4502864405538402 \\
0.6582278481012658  0.4569710771818502 \\
0.6835443037974683  0.46334718485042486 \\
0.7088607594936709  0.4693623183612529 \\
0.7341772151898734  0.47499899598156387 \\
0.7594936708860759  0.48047526689172426 \\
0.7848101265822784  0.48560969905433277 \\
0.810126582278481  0.49039378784263604 \\
0.8354430379746836  0.4948785610171534 \\
0.8607594936708861  0.4991214248084687 \\
0.8860759493670886  0.503221836101668 \\
0.9113924050632911  0.507133728168505 \\
0.9367088607594937  0.5108195389074862 \\
0.9620253164556962  0.5144051840424854 \\
0.9873417721518987  0.5177203347988065 \\
1.0126582278481013  0.5209601251125091 \\
1.0379746835443038  0.5240121049187454 \\
1.0632911392405062  0.5268956458673419 \\
1.0886075949367089  0.5296915419124543 \\
1.1139240506329113  0.5323234876529356 \\
1.139240506329114  0.5348302263884394 \\
1.1645569620253164  0.53724365046929 \\
1.1898734177215189  0.5395410808908124 \\
1.2151898734177216  0.5417385819479847 \\
1.240506329113924  0.5438333187652227 \\
1.2658227848101267  0.5458441905130889 \\
1.2911392405063291  0.5477964748322108 \\
1.3164556962025316  0.5496823758147314 \\
1.3417721518987342  0.551439526197794 \\
1.3670886075949367  0.5531553346452508 \\
1.3924050632911391  0.554796255129353 \\
1.4177215189873418  0.556406700700923 \\
1.4430379746835442  0.5579122558758702 \\
1.4683544303797469  0.5593804851889562 \\
1.4936708860759493  0.5607941431470427 \\
1.5189873417721518  0.562171656441428 \\
1.5443037974683544  0.5634674308231298 \\
1.5696202531645569  0.56469564067007 \\
1.5949367088607596  0.5659316464248675 \\
1.620253164556962  0.5671296175989077 \\
1.6455696202531644  0.5682871917958701 \\
1.6708860759493671  0.5694015341401705 \\
1.6962025316455696  0.5704934337194276 \\
1.7215189873417722  0.5715229660358281 \\
1.7468354430379747  0.5725227321585924 \\
1.7721518987341771  0.5734506814332186 \\
1.7974683544303798  0.574344612200832 \\
1.8227848101265822  0.5752666554846574 \\
1.8481012658227849  0.576147829312141 \\
1.8734177215189873  0.5769701794712484 \\
1.8987341772151898  0.577800089298581 \\
1.9240506329113924  0.578591728305524 \\
1.9493670886075949  0.5793486400865582 \\
1.9746835443037973  0.5800859439781336 \\
2.0  0.5807984427083457 \\
    };
\addlegendentry{\textcolor{black}{Ours (AUC: \textbf{0.46})}}

\addplot [color=cPLOT4, smooth, line width=\pckLineWidth]
table[row sep=crcr]{%
0.0  0.0 \\
0.02531645569620253  0.03350019726009275 \\
0.05063291139240506  0.06455956664201899 \\
0.0759493670886076  0.09345876082863414 \\
0.10126582278481013  0.12019990597662646 \\
0.12658227848101267  0.14527981403216167 \\
0.1518987341772152  0.16864722099508858 \\
0.17721518987341772  0.19046725836819836 \\
0.20253164556962025  0.21068181120201088 \\
0.22784810126582278  0.22964878253905632 \\
0.25316455696202533  0.24745345488649867 \\
0.27848101265822783  0.2641237045209178 \\
0.3037974683544304  0.279582517322271 \\
0.3291139240506329  0.2943085147850574 \\
0.35443037974683544  0.30820649233756753 \\
0.37974683544303794  0.32119896338049464 \\
0.4050632911392405  0.3334012128542709 \\
0.43037974683544306  0.3449093902891337 \\
0.45569620253164556  0.35577688584192263 \\
0.4810126582278481  0.36602472483988857 \\
0.5063291139240507  0.37569590289606164 \\
0.5316455696202531  0.38480081455425125 \\
0.5569620253164557  0.39341104042296343 \\
0.5822784810126582  0.4016012322259207 \\
0.6075949367088608  0.40931469245143504 \\
0.6329113924050633  0.4166442632748955 \\
0.6582278481012658  0.4234954488434889 \\
0.6835443037974683  0.4300213324387726 \\
0.7088607594936709  0.43626892574752124 \\
0.7341772151898734  0.44216475824417256 \\
0.7594936708860759  0.447762220085566 \\
0.7848101265822784  0.45309792841466673 \\
0.810126582278481  0.4581806240978599 \\
0.8354430379746836  0.4629897542871587 \\
0.8607594936708861  0.4675770554619784 \\
0.8860759493670886  0.4720337161202838 \\
0.9113924050632911  0.4761650748052795 \\
0.9367088607594937  0.48009846467863143 \\
0.9620253164556962  0.4839416110125467 \\
0.9873417721518987  0.48757946510622513 \\
1.0126582278481013  0.49103635964176623 \\
1.0379746835443038  0.49433190250862863 \\
1.0632911392405062  0.49752798849040514 \\
1.0886075949367089  0.5005509108219013 \\
1.1139240506329113  0.5035046149412118 \\
1.139240506329114  0.5062820842950255 \\
1.1645569620253164  0.5090099433261123 \\
1.1898734177215189  0.5115946411401869 \\
1.2151898734177216  0.5140782283917514 \\
1.240506329113924  0.516455507808901 \\
1.2658227848101267  0.5187248257142114 \\
1.2911392405063291  0.5209433520986347 \\
1.3164556962025316  0.5230262769342711 \\
1.3417721518987342  0.5250810892536953 \\
1.3670886075949367  0.5270397520428822 \\
1.3924050632911391  0.5288963593110307 \\
1.4177215189873418  0.5307109159246773 \\
1.4430379746835442  0.5324638139943634 \\
1.4683544303797469  0.5341470213725995 \\
1.4936708860759493  0.5357388040132389 \\
1.5189873417721518  0.5372821575293114 \\
1.5443037974683544  0.5388089742711417 \\
1.5696202531645569  0.5402828733353965 \\
1.5949367088607596  0.5417324397175519 \\
1.620253164556962  0.5430936524773269 \\
1.6455696202531644  0.544427461439786 \\
1.6708860759493671  0.5457265431763364 \\
1.6962025316455696  0.5469457759172595 \\
1.7215189873417722  0.5481437470912995 \\
1.7468354430379747  0.5492741537305782 \\
1.7721518987341771  0.5503662895494674 \\
1.7974683544303798  0.5514700011103263 \\
1.8227848101265822  0.552522448671034 \\
1.8481012658227849  0.5535555245819149 \\
1.8734177215189873  0.5545959239213889 \\
1.8987341772151898  0.5556015960349541 \\
1.9240506329113924  0.5565444284063983 \\
1.9493670886075949  0.5574447376440766 \\
1.9746835443037973  0.5583157531673829 \\
2.0  0.5592049591423557 \\
    };
\addlegendentry{\textcolor{black}{URSSM (AUC: {0.43})}}

\end{axis}
\end{tikzpicture}

%% file: figs/tikz/smal_matching_conformal_pck.tex
\newcommand{\pckLineWidth}{1pt}
\newcommand{\plotWidth}{\columnwidth}
\newcommand{\plotHeight}{0.7\columnwidth}
\newcommand{\pckTitle}{\textbf{SMAL}}
\definecolor{cPLOT0}{RGB}{28,213,227}
\definecolor{cPLOT1}{RGB}{80,150,80}
\definecolor{cPLOT2}{RGB}{90,130,213}
\definecolor{cPLOT3}{RGB}{247,179,43}
\definecolor{cPLOT4}{RGB}{124,42,43}
\definecolor{cPLOT5}{RGB}{242,64,0}

\pgfplotsset{%
    label style = {font=\LARGE},
    tick label style = {font=\large},
    title style =  {font=\LARGE},
    legend style={  fill= gray!10,
                    fill opacity=0.6, 
                    font=\large,
                    draw=gray!20, %
                    text opacity=1}
}
\begin{tikzpicture}[scale=0.5, transform shape]
	\begin{axis}[
		width=\plotWidth,
		height=\plotHeight,
		grid=major,
		title=\pckTitle,
		legend style={
			at={(0.97,0.03)},
			anchor=south east,
			legend columns=1},
		legend cell align={left},
        xlabel={\LARGE Conformal distortion},
		xmin=0,
        xmax=2.0,
        ylabel near ticks,
        xtick={0, 0.5, 1.0, 1.5, 2.0},
	ymin=0,
        ymax=1,
        ytick={0, 0.2, 0.40, 0.60, 0.80, 1.0}
	]

\addplot [color=cPLOT1, smooth, line width=\pckLineWidth]
table[row sep=crcr]{%
0.0  0.04898765114484178 \\
0.02531645569620253  0.10131785438641626 \\
0.05063291139240506  0.13879084126575766 \\
0.0759493670886076  0.16917481348083355 \\
0.10126582278481013  0.1949289297658863 \\
0.12658227848101267  0.217362361718549 \\
0.1518987341772152  0.23760322871108824 \\
0.17721518987341772  0.25604804476459997 \\
0.20253164556962025  0.27272961152559816 \\
0.22784810126582278  0.28818304605093903 \\
0.25316455696202533  0.30244565217391306 \\
0.27848101265822783  0.3158271160277849 \\
0.3037974683544304  0.3285023797272961 \\
0.3291139240506329  0.3403315538976074 \\
0.35443037974683544  0.3514435940313867 \\
0.37974683544303794  0.36199800617442757 \\
0.4050632911392405  0.37202244661692824 \\
0.43037974683544306  0.38143651916645227 \\
0.45569620253164556  0.3904720864419861 \\
0.4810126582278481  0.39906097247234373 \\
0.5063291139240507  0.4073080138924621 \\
0.5316455696202531  0.4153608181116542 \\
0.5569620253164557  0.4230132492925135 \\
0.5822784810126582  0.4300787882685876 \\
0.6075949367088608  0.436933046050939 \\
0.6329113924050633  0.4436033573449961 \\
0.6582278481012658  0.4499649472600978 \\
0.6835443037974683  0.45605351170568564 \\
0.7088607594936709  0.4620314509904811 \\
0.7341772151898734  0.46774890661178287 \\
0.7594936708860759  0.4731762927707744 \\
0.7848101265822784  0.4784030100334448 \\
0.810126582278481  0.4835242474916388 \\
0.8354430379746836  0.4883811422691021 \\
0.8607594936708861  0.49309653974787754 \\
0.8860759493670886  0.4976572549524055 \\
0.9113924050632911  0.5020674684846925 \\
0.9367088607594937  0.5062959866220735 \\
0.9620253164556962  0.5104296372523798 \\
0.9873417721518987  0.5143896321070234 \\
1.0126582278481013  0.5182541162850527 \\
1.0379746835443038  0.5219336892204786 \\
1.0632911392405062  0.5255569848211988 \\
1.0886075949367089  0.5289802546951377 \\
1.1139240506329113  0.5323665423205557 \\
1.139240506329114  0.5356232312837664 \\
1.1645569620253164  0.538789233341909 \\
1.1898734177215189  0.5419182531515307 \\
1.2151898734177216  0.5448945201955235 \\
1.240506329113924  0.5477932853100077 \\
1.2658227848101267  0.5506775791098534 \\
1.2911392405063291  0.5534528556727554 \\
1.3164556962025316  0.5561422691021354 \\
1.3417721518987342  0.558734885515822 \\
1.3670886075949367  0.5612921276048366 \\
1.3924050632911391  0.5637699382557242 \\
1.4177215189873418  0.5661734628248006 \\
1.4430379746835442  0.5685149215333162 \\
1.4683544303797469  0.5707875611011063 \\
1.4936708860759493  0.5730273990223823 \\
1.5189873417721518  0.5752003473115513 \\
1.5443037974683544  0.5773530357602263 \\
1.5696202531645569  0.5794285438641626 \\
1.5949367088607596  0.581417867249807 \\
1.620253164556962  0.5834158734242346 \\
1.6455696202531644  0.585311937226653 \\
1.6708860759493671  0.5871530100334448 \\
1.6962025316455696  0.5889985850270131 \\
1.7215189873417722  0.5907676228453821 \\
1.7468354430379747  0.5924974273218421 \\
1.7721518987341771  0.5942236943658349 \\
1.7974683544303798  0.5958808206843323 \\
1.8227848101265822  0.5974971057370723 \\
1.8481012658227849  0.5990786596346797 \\
1.8734177215189873  0.6006447774633393 \\
1.8987341772151898  0.6021687676871623 \\
1.9240506329113924  0.6036744275791098 \\
1.9493670886075949  0.6051292770774376 \\
1.9746835443037973  0.6065532544378698 \\
2.0  0.6078981219449446 \\
    };
\addlegendentry{\textcolor{black}{Ours (AUC: \textbf{0.47})}}

\addplot [color=cPLOT4, smooth, line width=\pckLineWidth]
table[row sep=crcr]{%
0.0  0.04962471057370723 \\
0.02531645569620253  0.07161210445073321 \\
0.05063291139240506  0.08742507074864934 \\
0.0759493670886076  0.10027302546951376 \\
0.10126582278481013  0.11135805248263442 \\
0.12658227848101267  0.1213352199639825 \\
0.1518987341772152  0.13039587085155646 \\
0.17721518987341772  0.1389033959351685 \\
0.20253164556962025  0.14680666323642913 \\
0.22784810126582278  0.15430698482119887 \\
0.25316455696202533  0.16153878312323128 \\
0.27848101265822783  0.16846989966555184 \\
0.3037974683544304  0.17508843581167996 \\
0.3291139240506329  0.18141593774118858 \\
0.35443037974683544  0.18749324671983536 \\
0.37974683544303794  0.19341619500900437 \\
0.4050632911392405  0.19909988422948288 \\
0.43037974683544306  0.20465526112683302 \\
0.45569620253164556  0.21004920246977102 \\
0.4810126582278481  0.21530293285310007 \\
0.5063291139240507  0.22045504244918962 \\
0.5316455696202531  0.22535985335734499 \\
0.5569620253164557  0.23021481862618987 \\
0.5822784810126582  0.23488101363519423 \\
0.6075949367088608  0.23947453048623618 \\
0.6329113924050633  0.24397736043220994 \\
0.6582278481012658  0.2483936840751222 \\
0.6835443037974683  0.2527038847440185 \\
0.7088607594936709  0.25690056598919475 \\
0.7341772151898734  0.26103260869565215 \\
0.7594936708860759  0.2650594931824029 \\
0.7848101265822784  0.269023990223823 \\
0.810126582278481  0.27282544378698226 \\
0.8354430379746836  0.2765651530743504 \\
0.8607594936708861  0.28017236943658347 \\
0.8860759493670886  0.2837741188577309 \\
0.9113924050632911  0.28725398765114485 \\
0.9367088607594937  0.2906962310264986 \\
0.9620253164556962  0.2941002058142526 \\
0.9873417721518987  0.2973273089786468 \\
1.0126582278481013  0.3006322356573193 \\
1.0379746835443038  0.30374774890661177 \\
1.0632911392405062  0.306833354772318 \\
1.0886075949367089  0.309829560072035 \\
1.1139240506329113  0.31282029843066633 \\
1.139240506329114  0.31575894005659894 \\
1.1645569620253164  0.31858374067404166 \\
1.1898734177215189  0.3213512992024698 \\
1.2151898734177216  0.3240419989709287 \\
1.240506329113924  0.3267407383586313 \\
1.2658227848101267  0.329435618729097 \\
1.2911392405063291  0.33208933624903525 \\
1.3164556962025316  0.3346259969127862 \\
1.3417721518987342  0.3371292127604837 \\
1.3670886075949367  0.3396292127604837 \\
1.3924050632911391  0.34205524826344225 \\
1.4177215189873418  0.3444812837664008 \\
1.4430379746835442  0.34684525340879857 \\
1.4683544303797469  0.34915198096218164 \\
1.4936708860759493  0.3514210830975045 \\
1.5189873417721518  0.35364580653460254 \\
1.5443037974683544  0.3558267944430152 \\
1.5696202531645569  0.35794796758425523 \\
1.5949367088607596  0.36001543606894776 \\
1.620253164556962  0.362103485978904 \\
1.6455696202531644  0.364142655003859 \\
1.6708860759493671  0.3661811808592745 \\
1.6962025316455696  0.3681785438641626 \\
1.7215189873417722  0.37013442243375355 \\
1.7468354430379747  0.3720944816053512 \\
1.7721518987341771  0.37397478775405196 \\
1.7974683544303798  0.3758431952662722 \\
1.8227848101265822  0.37770838693079495 \\
1.8481012658227849  0.3795423848726524 \\
1.8734177215189873  0.38133554154875227 \\
1.8987341772151898  0.38310168510419346 \\
1.9240506329113924  0.38489484178029326 \\
1.9493670886075949  0.3866114612811937 \\
1.9746835443037973  0.3883203627476203 \\
2.0  0.39000771803447387 \\
    };
\addlegendentry{\textcolor{black}{URSSM (AUC: {0.28})}}

\end{axis}
\end{tikzpicture}

%% file: figs/tikz/dt4d_intra_matching_conformal_pck.tex
\newcommand{\pckLineWidth}{1pt}
\newcommand{\plotWidth}{\columnwidth}
\newcommand{\plotHeight}{0.7\columnwidth}
\newcommand{\pckTitle}{\textbf{DT4D-H intra-class}}
\definecolor{cPLOT0}{RGB}{28,213,227}
\definecolor{cPLOT1}{RGB}{80,150,80}
\definecolor{cPLOT2}{RGB}{90,130,213}
\definecolor{cPLOT3}{RGB}{247,179,43}
\definecolor{cPLOT4}{RGB}{124,42,43}
\definecolor{cPLOT5}{RGB}{242,64,0}

\pgfplotsset{%
    label style = {font=\LARGE},
    tick label style = {font=\large},
    title style =  {font=\LARGE},
    legend style={  fill= gray!10,
                    fill opacity=0.6, 
                    font=\large,
                    draw=gray!20, %
                    text opacity=1}
}
\begin{tikzpicture}[scale=0.5, transform shape]
	\begin{axis}[
		width=\plotWidth,
		height=\plotHeight,
		grid=major,
		title=\pckTitle,
		legend style={
			at={(0.97,0.03)},
			anchor=south east,
			legend columns=1},
		legend cell align={left},
        xlabel={\LARGE Conformal distortion},
		xmin=0,
        xmax=2.0,
        ylabel near ticks,
        xtick={0, 0.5, 1.0, 1.5, 2.0},
	ymin=0,
        ymax=1,
        ytick={0, 0.2, 0.40, 0.60, 0.80, 1.0}
	]

\addplot [color=cPLOT1, smooth, line width=\pckLineWidth]
table[row sep=crcr]{%
0.0  0.0974353452688137 \\
0.02531645569620253  0.14104926172247317 \\
0.05063291139240506  0.1809113194971539 \\
0.0759493670886076  0.21758740625092413 \\
0.10126582278481013  0.2513108378519168 \\
0.12658227848101267  0.2825553736579788 \\
0.1518987341772152  0.31139994828683065 \\
0.17721518987341772  0.33805945194656106 \\
0.20253164556962025  0.3628389316756811 \\
0.22784810126582278  0.38581845947833393 \\
0.25316455696202533  0.40723483920768144 \\
0.27848101265822783  0.4272142449308583 \\
0.3037974683544304  0.4458014896729104 \\
0.3291139240506329  0.4632189703923149 \\
0.35443037974683544  0.47940557152530344 \\
0.37974683544303794  0.49465815092814525 \\
0.4050632911392405  0.5088871583765098 \\
0.43037974683544306  0.5222348429989695 \\
0.45569620253164556  0.5347310043197936 \\
0.4810126582278481  0.5464864237765323 \\
0.5063291139240507  0.5575021629298468 \\
0.5316455696202531  0.567885515232258 \\
0.5569620253164557  0.577676516685838 \\
0.5822784810126582  0.586856741630542 \\
0.6075949367088608  0.595497845410987 \\
0.6329113924050633  0.6036472191281099 \\
0.6582278481012658  0.6113340557000879 \\
0.6835443037974683  0.618591945939265 \\
0.7088607594936709  0.6254260459974235 \\
0.7341772151898734  0.6319402371678168 \\
0.7594936708860759  0.6381180652601999 \\
0.7848101265822784  0.6439636248656936 \\
0.810126582278481  0.6495030000462537 \\
0.8354430379746836  0.6547389963550556 \\
0.8607594936708861  0.659706341990866 \\
0.8860759493670886  0.6644368837735145 \\
0.9113924050632911  0.6689327448243231 \\
0.9367088607594937  0.6731927877568692 \\
0.9620253164556962  0.6772261116625329 \\
0.9873417721518987  0.681086931960786 \\
1.0126582278481013  0.6847892764175058 \\
1.0379746835443038  0.6882994783945866 \\
1.0632911392405062  0.6916371009384954 \\
1.0886075949367089  0.6948314886189323 \\
1.1139240506329113  0.6978804424888138 \\
1.139240506329114  0.7007865406240309 \\
1.1645569620253164  0.7035560007370264 \\
1.1898734177215189  0.7061914767294529 \\
1.2151898734177216  0.7087077546248027 \\
1.240506329113924  0.7111814942679515 \\
1.2658227848101267  0.7135071461988924 \\
1.2911392405063291  0.7157430962539799 \\
1.3164556962025316  0.7178919983348663 \\
1.3417721518987342  0.7199496061989077 \\
1.3670886075949367  0.721917133058288 \\
1.3924050632911391  0.7238142177852357 \\
1.4177215189873418  0.7256442725390181 \\
1.4430379746835442  0.7273894782656829 \\
1.4683544303797469  0.7290673507161364 \\
1.4936708860759493  0.7306828185648762 \\
1.5189873417721518  0.732241492918253 \\
1.5443037974683544  0.7337426913444135 \\
1.5696202531645569  0.7351786037899232 \\
1.5949367088607596  0.7365728120666084 \\
1.620253164556962  0.7379151555224281 \\
1.6455696202531644  0.7392155673321388 \\
1.6708860759493671  0.7404629011088001 \\
1.6962025316455696  0.7416588250191649 \\
1.7215189873417722  0.7427981829114514 \\
1.7468354430379747  0.7439102435295982 \\
1.7721518987341771  0.7449970541691657 \\
1.7974683544303798  0.7460381418745493 \\
1.8227848101265822  0.7470423024340828 \\
1.8481012658227849  0.7480301604548939 \\
1.8734177215189873  0.7489700965944376 \\
1.8987341772151898  0.7498881949146695 \\
1.9240506329113924  0.7507719441649423 \\
1.9493670886075949  0.7516307467396818 \\
1.9746835443037973  0.7524612663053822 \\
2.0  0.753252356475103 \\
    };
\addlegendentry{\textcolor{black}{Ours (AUC: \textbf{0.62})}}

\addplot [color=cPLOT4, smooth, line width=\pckLineWidth]
table[row sep=crcr]{%
0.0  0.09742958251093976 \\
0.02531645569620253  0.14039480957497386 \\
0.05063291139240506  0.17979213125742619 \\
0.0759493670886076  0.21591067118689306 \\
0.10126582278481013  0.2491387330880117 \\
0.12658227848101267  0.2800028510486324 \\
0.1518987341772152  0.30851508136483335 \\
0.17721518987341772  0.3348695379860526 \\
0.20253164556962025  0.3594345066358915 \\
0.22784810126582278  0.3821945188590043 \\
0.25316455696202533  0.40335786802222906 \\
0.27848101265822783  0.4231596898119749 \\
0.3037974683544304  0.441560251529216 \\
0.3291139240506329  0.4588103089672304 \\
0.35443037974683544  0.4748715701164608 \\
0.37974683544303794  0.4899703748749823 \\
0.4050632911392405  0.5041007330076364 \\
0.43037974683544306  0.5173404417457214 \\
0.45569620253164556  0.5297235468561502 \\
0.4810126582278481  0.5413944964145788 \\
0.5063291139240507  0.5523263722756752 \\
0.5316455696202531  0.5626673379774084 \\
0.5569620253164557  0.5723619648881229 \\
0.5822784810126582  0.5815323324838321 \\
0.6075949367088608  0.5901560721648937 \\
0.6329113924050633  0.5982974841770592 \\
0.6582278481012658  0.6059787096426863 \\
0.6835443037974683  0.6132480495718499 \\
0.7088607594936709  0.6200838936225227 \\
0.7341772151898734  0.6265383340928548 \\
0.7594936708860759  0.6326880308277216 \\
0.7848101265822784  0.6385044733407996 \\
0.810126582278481  0.6440122291788146 \\
0.8354430379746836  0.6492230513347987 \\
0.8607594936708861  0.6542215613585853 \\
0.8860759493670886  0.6590001766740243 \\
0.9113924050632911  0.6635155249455381 \\
0.9367088607594937  0.6677864109619787 \\
0.9620253164556962  0.6718349000199422 \\
0.9873417721518987  0.6757373486612582 \\
1.0126582278481013  0.6794465932622745 \\
1.0379746835443038  0.6829766615888682 \\
1.0632911392405062  0.6863386242022183 \\
1.0886075949367089  0.6895476462546244 \\
1.1139240506329113  0.6926099454585297 \\
1.139240506329114  0.6954951915093345 \\
1.1645569620253164  0.6982676846527901 \\
1.1898734177215189  0.7009148378124874 \\
1.2151898734177216  0.7034835871347948 \\
1.240506329113924  0.7059407209361752 \\
1.2658227848101267  0.7082964756944313 \\
1.2911392405063291  0.7105627560541183 \\
1.3164556962025316  0.7127271265903505 \\
1.3417721518987342  0.7148029326371518 \\
1.3670886075949367  0.7167753881710296 \\
1.3924050632911391  0.7187017416419159 \\
1.4177215189873418  0.7205484780632281 \\
1.4430379746835442  0.7223024037524652 \\
1.4683544303797469  0.7240013557646156 \\
1.4936708860759493  0.7256198566438153 \\
1.5189873417721518  0.7271927104145925 \\
1.5443037974683544  0.7286845822720887 \\
1.5696202531645569  0.7301427116657175 \\
1.5949367088607596  0.7315395738440552 \\
1.620253164556962  0.7328929120352924 \\
1.6455696202531644  0.7342059867471734 \\
1.6708860759493671  0.7354825892677733 \\
1.6962025316455696  0.7366936025046765 \\
1.7215189873417722  0.7378537366556138 \\
1.7468354430379747  0.7389927912448543 \\
1.7721518987341771  0.7400827865664048 \\
1.7974683544303798  0.7411378262119042 \\
1.8227848101265822  0.742160260779959 \\
1.8481012658227849  0.743156611286058 \\
1.8734177215189873  0.7441201292374279 \\
1.8987341772151898  0.7450527861038676 \\
1.9240506329113924  0.7459401749906924 \\
1.9493670886075949  0.7467975368759635 \\
1.9746835443037973  0.7476327576388768 \\
2.0  0.7484399228700354 \\
    };
\addlegendentry{\textcolor{black}{URSSM (AUC: {0.61})}}

\end{axis}
\end{tikzpicture}

%% file: figs/tikz/dt4d_inter_matching_conformal_pck.tex
\newcommand{\pckLineWidth}{1pt}
\newcommand{\plotWidth}{\columnwidth}
\newcommand{\plotHeight}{0.7\columnwidth}
\newcommand{\pckTitle}{\textbf{DT4D-H inter-class}}
\definecolor{cPLOT0}{RGB}{28,213,227}
\definecolor{cPLOT1}{RGB}{80,150,80}
\definecolor{cPLOT2}{RGB}{90,130,213}
\definecolor{cPLOT3}{RGB}{247,179,43}
\definecolor{cPLOT4}{RGB}{124,42,43}
\definecolor{cPLOT5}{RGB}{242,64,0}

\pgfplotsset{%
    label style = {font=\LARGE},
    tick label style = {font=\large},
    title style =  {font=\LARGE},
    legend style={  fill= gray!10,
                    fill opacity=0.6, 
                    font=\large,
                    draw=gray!20, %
                    text opacity=1}
}
\begin{tikzpicture}[scale=0.5, transform shape]
	\begin{axis}[
		width=\plotWidth,
		height=\plotHeight,
		grid=major,
		title=\pckTitle,
		legend style={
			at={(0.97,0.03)},
			anchor=south east,
			legend columns=1},
		legend cell align={left},
        xlabel={\LARGE Conformal distortion},
		xmin=0,
        xmax=2.0,
        ylabel near ticks,
        xtick={0, 0.5, 1.0, 1.5, 2.0},
	ymin=0,
        ymax=1,
        ytick={0, 0.2, 0.40, 0.60, 0.80, 1.0}
	]

\addplot [color=cPLOT1, smooth, line width=\pckLineWidth]
table[row sep=crcr]{%
0.0  0.0 \\
0.02531645569620253  0.02200583346643408 \\
0.05063291139240506  0.04269777987955165 \\
0.0759493670886076  0.06219474257271738 \\
0.10126582278481013  0.0805741287642718 \\
0.12658227848101267  0.09793171879518249 \\
0.1518987341772152  0.11431772557444717 \\
0.17721518987341772  0.12986175269150588 \\
0.20253164556962025  0.14457246474396115 \\
0.22784810126582278  0.15853149685621257 \\
0.25316455696202533  0.1717976221421182 \\
0.27848101265822783  0.18439537289609478 \\
0.3037974683544304  0.19635246539119613 \\
0.3291139240506329  0.20771441486302106 \\
0.35443037974683544  0.21852062957174964 \\
0.37974683544303794  0.22885029141233984 \\
0.4050632911392405  0.23870126033357658 \\
0.43037974683544306  0.24810245312504894 \\
0.45569620253164556  0.2571117033659352 \\
0.4810126582278481  0.2656715428516548 \\
0.5063291139240507  0.2738339591678228 \\
0.5316455696202531  0.2816111140689055 \\
0.5569620253164557  0.28910713931524623 \\
0.5822784810126582  0.2963040271588158 \\
0.6075949367088608  0.3031876323830462 \\
0.6329113924050633  0.30974589762621346 \\
0.6582278481012658  0.31610002286201055 \\
0.6835443037974683  0.32217416676322996 \\
0.7088607594936709  0.32800136963277765 \\
0.7341772151898734  0.3335636759189954 \\
0.7594936708860759  0.33894392126281814 \\
0.7848101265822784  0.34408380231772767 \\
0.810126582278481  0.3490073816108009 \\
0.8354430379746836  0.3537755201107398 \\
0.8607594936708861  0.3583737072263788 \\
0.8860759493670886  0.3628098246097538 \\
0.9113924050632911  0.36707327639753173 \\
0.9367088607594937  0.37118358403250373 \\
0.9620253164556962  0.37514701107920173 \\
0.9873417721518987  0.37900223504861047 \\
1.0126582278481013  0.3827047324361821 \\
1.0379746835443038  0.3862753817903564 \\
1.0632911392405062  0.3897126172200004 \\
1.0886075949367089  0.3930316800654751 \\
1.1139240506329113  0.39623121322113203 \\
1.139240506329114  0.3993255706890234 \\
1.1645569620253164  0.40232226874658766 \\
1.1898734177215189  0.40520413478773304 \\
1.2151898734177216  0.40799658844518494 \\
1.240506329113924  0.4107172720923751 \\
1.2658227848101267  0.4133376343143119 \\
1.2911392405063291  0.41588612213324483 \\
1.3164556962025316  0.4183699386483854 \\
1.3417721518987342  0.4207815153859243 \\
1.3670886075949367  0.42308786423932665 \\
1.3924050632911391  0.4253543350652089 \\
1.4177215189873418  0.427530453972718 \\
1.4430379746835442  0.4296359511901295 \\
1.4683544303797469  0.4316736975178538 \\
1.4936708860759493  0.43365094825147377 \\
1.5189873417721518  0.43558800777934714 \\
1.5443037974683544  0.4374883732583376 \\
1.5696202531645569  0.43932302350611374 \\
1.5949367088607596  0.44110427686625514 \\
1.620253164556962  0.442817570551225 \\
1.6455696202531644  0.44450001618087503 \\
1.6708860759493671  0.4461380948950905 \\
1.6962025316455696  0.4477281529478945 \\
1.7215189873417722  0.44927728904575637 \\
1.7468354430379747  0.4507684871716979 \\
1.7721518987341771  0.45222288685601425 \\
1.7974683544303798  0.4536536415842225 \\
1.8227848101265822  0.4550400293969962 \\
1.8481012658227849  0.45639123685564886 \\
1.8734177215189873  0.45770481073073876 \\
1.8987341772151898  0.4590024125162722 \\
1.9240506329113924  0.46025042774926117 \\
1.9493670886075949  0.4614754243825952 \\
1.9746835443037973  0.4626500515178183 \\
2.0  0.4638053137993634 \\
    };
\addlegendentry{\textcolor{black}{Ours (AUC: \textbf{0.34})}}

\addplot [color=cPLOT4, smooth, line width=\pckLineWidth]
table[row sep=crcr]{%
0.0  0.0 \\
0.02531645569620253  0.01809027049203431 \\
0.05063291139240506  0.03522889674618262 \\
0.0759493670886076  0.05139609633779821 \\
0.10126582278481013  0.06672825838456407 \\
0.12658227848101267  0.08130378183587164 \\
0.1518987341772152  0.09514782534259088 \\
0.17721518987341772  0.10828111086404829 \\
0.20253164556962025  0.12076945802420033 \\
0.22784810126582278  0.1326843236550822 \\
0.25316455696202533  0.14409105761331048 \\
0.27848101265822783  0.15495541907944393 \\
0.3037974683544304  0.16531373672776775 \\
0.3291139240506329  0.17523005550562104 \\
0.35443037974683544  0.18473000383121363 \\
0.37974683544303794  0.19381566955938956 \\
0.4050632911392405  0.2025373177955176 \\
0.43037974683544306  0.21091389582230685 \\
0.45569620253164556  0.2189199318106608 \\
0.4810126582278481  0.22661477303451955 \\
0.5063291139240507  0.23403516573913716 \\
0.5316455696202531  0.24115454197162398 \\
0.5569620253164557  0.24801643350547697 \\
0.5822784810126582  0.2546070083023548 \\
0.6075949367088608  0.2609394198477745 \\
0.6329113924050633  0.26705046658336123 \\
0.6582278481012658  0.27293028339497727 \\
0.6835443037974683  0.2785947901758078 \\
0.7088607594936709  0.2840755657303682 \\
0.7341772151898734  0.28933659456259964 \\
0.7594936708860759  0.2944389464266886 \\
0.7848101265822784  0.29933063373702007 \\
0.810126582278481  0.30408274795103146 \\
0.8354430379746836  0.30866647667187586 \\
0.8607594936708861  0.31308834444594075 \\
0.8860759493670886  0.31739167426123865 \\
0.9113924050632911  0.32152082478617755 \\
0.9367088607594937  0.32551791849223477 \\
0.9620253164556962  0.32940717449560036 \\
0.9873417721518987  0.3331784667002811 \\
1.0126582278481013  0.33685157753092376 \\
1.0379746835443038  0.340406985544737 \\
1.0632911392405062  0.34386676980669595 \\
1.0886075949367089  0.347196480711875 \\
1.1139240506329113  0.35043265572004373 \\
1.139240506329114  0.3535744074928935 \\
1.1645569620253164  0.3566236672961549 \\
1.1898734177215189  0.35958977828025485 \\
1.2151898734177216  0.36247973475892065 \\
1.240506329113924  0.3652775646425958 \\
1.2658227848101267  0.3679976219333327 \\
1.2911392405063291  0.3706588017174694 \\
1.3164556962025316  0.37321809418521984 \\
1.3417721518987342  0.3757399096585209 \\
1.3670886075949367  0.37818160370218423 \\
1.3924050632911391  0.3805692223053884 \\
1.4177215189873418  0.38289921611489885 \\
1.4430379746835442  0.38514308591209506 \\
1.4683544303797469  0.38731513350265834 \\
1.4936708860759493  0.3894416658576229 \\
1.5189873417721518  0.3915237269044107 \\
1.5443037974683544  0.3935639264615767 \\
1.5696202531645569  0.3955407074278568 \\
1.5949367088607596  0.39747045946377624 \\
1.620253164556962  0.3993261970454766 \\
1.6455696202531644  0.401170921192875 \\
1.6708860759493671  0.4029514960001921 \\
1.6962025316455696  0.40469386306386435 \\
1.7215189873417722  0.4063924373721841 \\
1.7468354430379747  0.40805776259211357 \\
1.7721518987341771  0.40968180048250324 \\
1.7974683544303798  0.41127707817241715 \\
1.8227848101265822  0.4128162969597702 \\
1.8481012658227849  0.414307651674825 \\
1.8734177215189873  0.4157825645329834 \\
1.8987341772151898  0.4172228711971029 \\
1.9240506329113924  0.4186373406575281 \\
1.9493670886075949  0.42001997033158267 \\
1.9746835443037973  0.42137530130355216 \\
2.0  0.42271121522547267 \\
    };
\addlegendentry{\textcolor{black}{URSSM (AUC: {0.30})}}

\end{axis}
\end{tikzpicture}

%% file: figs/topkids_results.tex
\def\heightIQ{2.6cm}
\def\widthIQ{2.2cm}
\def\hspaceColsIQ{-0.0cm}
\def\firstRow{kid00-kid17}
\begin{tabular}{cccc}%
    \setlength{\tabcolsep}{0pt} 
    \includegraphics[height=\heightIQ, width=\widthIQ]{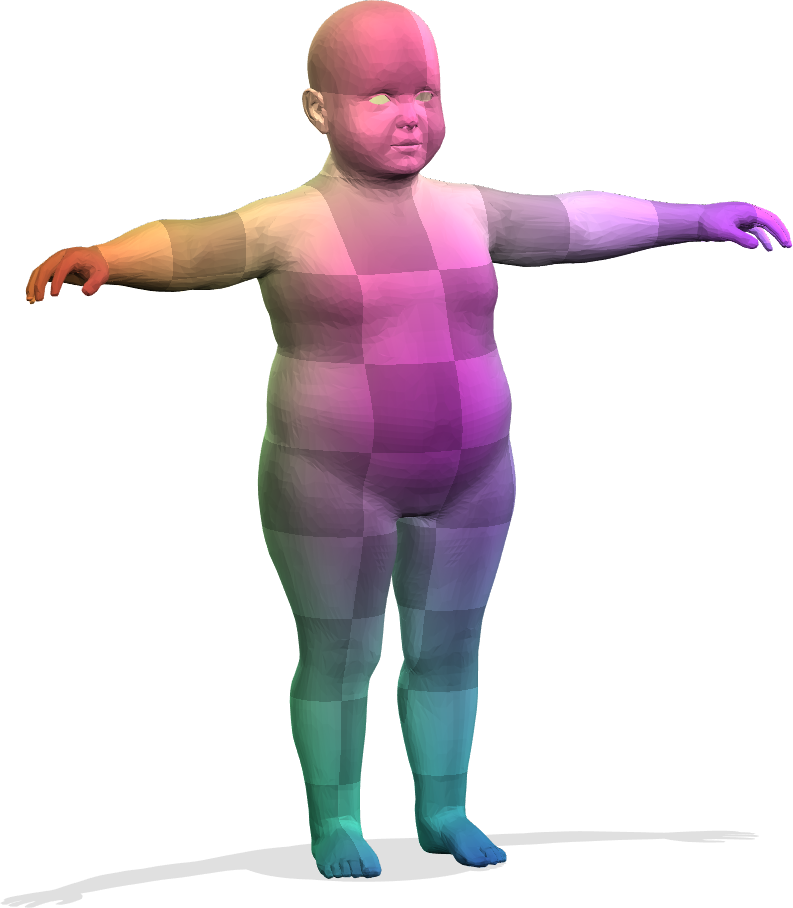}&
    \hspace{\hspaceColsIQ}
    \begin{overpic}[height=\heightIQ, width=\widthIQ]{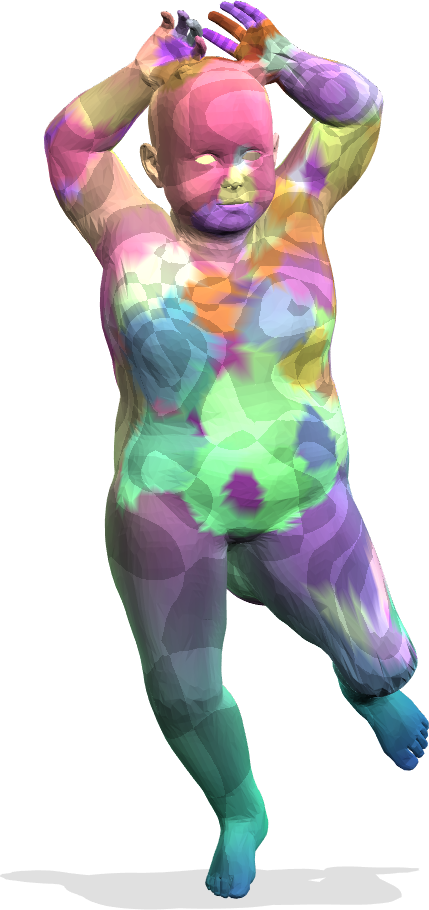}
    \end{overpic}
    &
    \hspace{\hspaceColsIQ}
    \begin{overpic}[height=\heightIQ, width=\widthIQ]{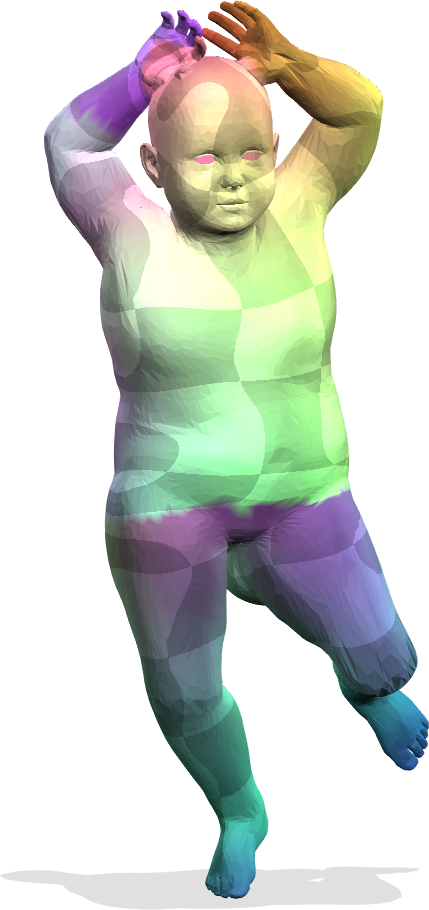}
    \end{overpic}
    &
    \hspace{\hspaceColsIQ}
    \begin{overpic}[height=\heightIQ, width=\widthIQ]{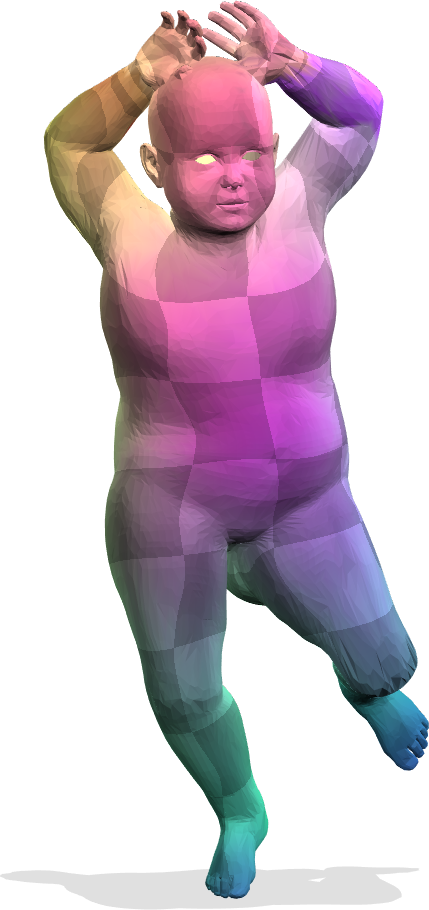}
    \end{overpic} \\

    \textbf{\small{Source}} & \textbf{\small{AttnFMaps}} & \textbf{\small{URSSM}} & \textbf{\small{Ours}}
    \\
\end{tabular}

%% file: figs/shrec19_results.tex
\def\rowOnecolumnOne{14-1}
\def\rowOnecolumnTwo{14-2}
\def\rowOnecolumnThree{14-21}
\def\rowOnecolumnFour{14-3}
\def\rowTwocolumnOne{14-44}
\def\rowTwocolumnTwo{14-8}
\def\rowTwocolumnThree{14-9}
\def\rowTwocolumnFour{14-42}

\def\hspaceCols{-0.4cm}
\def\heightT{2.6cm}
\def\widthT{2.2cm}
\begin{tabular}{ccccc}
        \setlength{\tabcolsep}{0pt} 
        \hspace{\hspaceCols}
        \includegraphics[height=\heightT, width=\widthT]{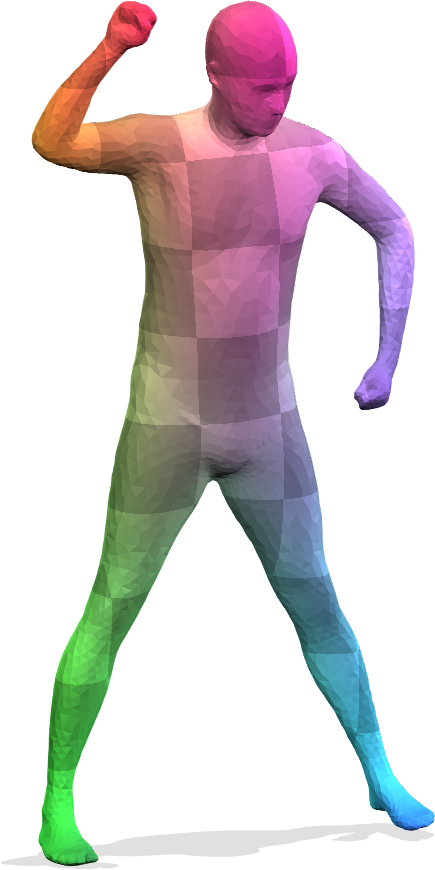}&
        \hspace{\hspaceCols}
        \includegraphics[height=\heightT, width=\widthT]{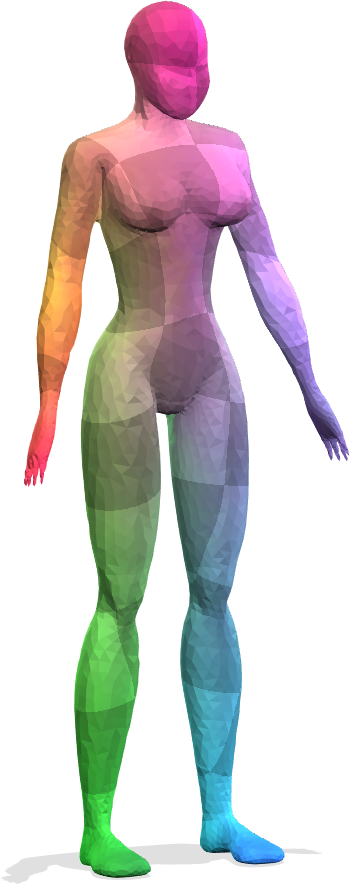}&
        \hspace{\hspaceCols}
        \includegraphics[height=\heightT, width=\widthT]{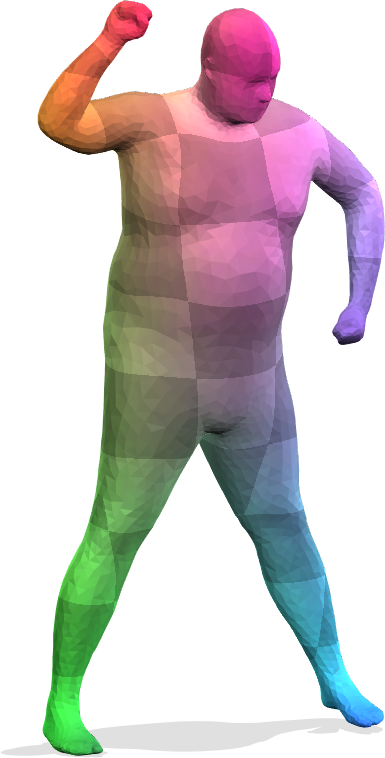}&
        \hspace{\hspaceCols}
        \includegraphics[height=\heightT, width=\widthT]{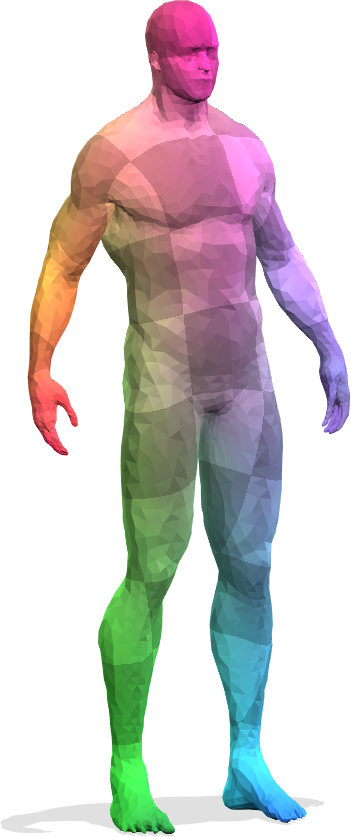}&
        \hspace{\hspaceCols}
        \includegraphics[height=\heightT, width=\widthT]{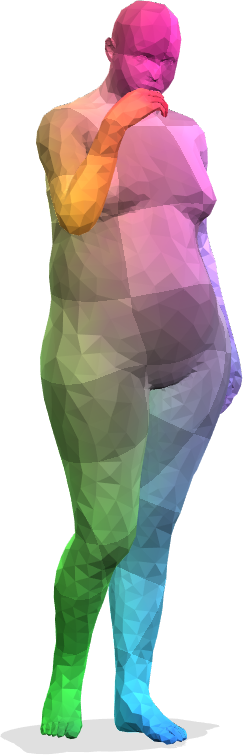}
        \\
        \hspace{\hspaceCols}
        &
        \hspace{\hspaceCols}
        \includegraphics[height=\heightT, width=\widthT]{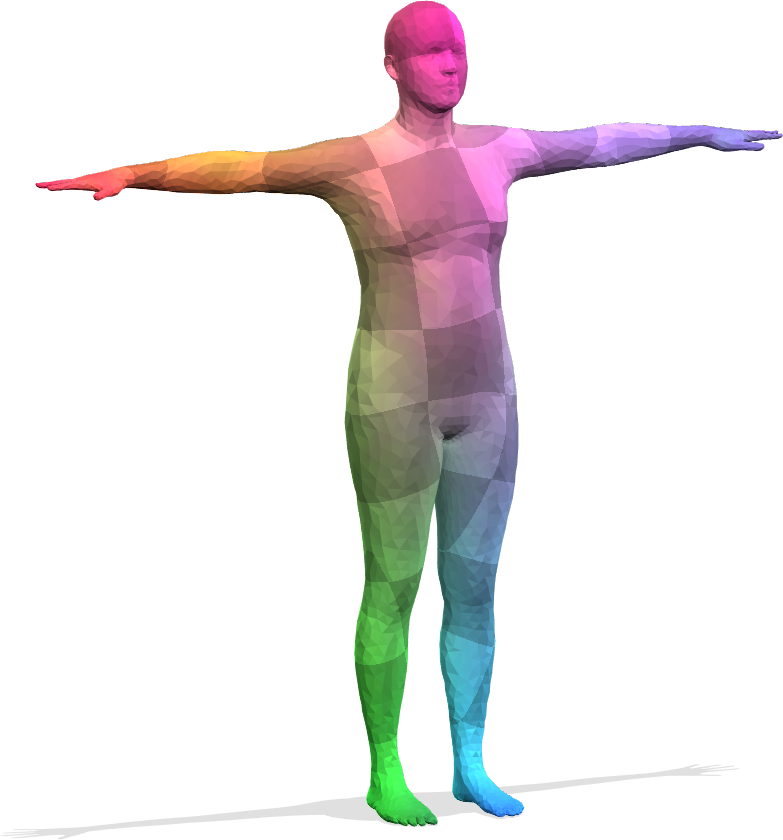}&
        \hspace{\hspaceCols}
        \includegraphics[height=\heightT, width=\widthT]{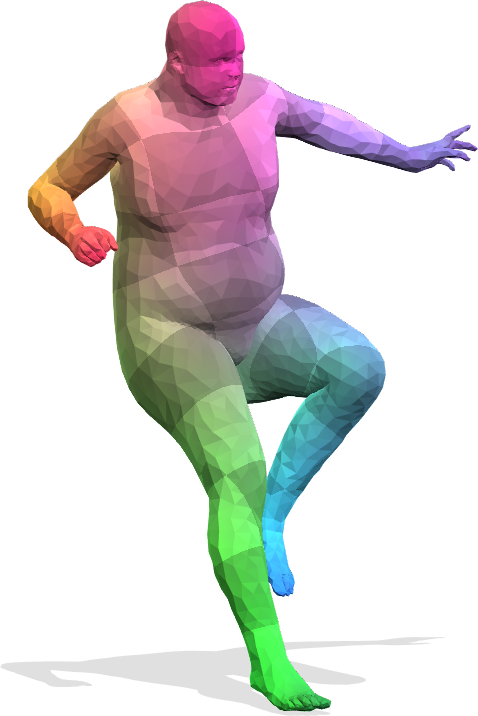}&
        \hspace{\hspaceCols}
        \includegraphics[height=\heightT, width=\widthT]{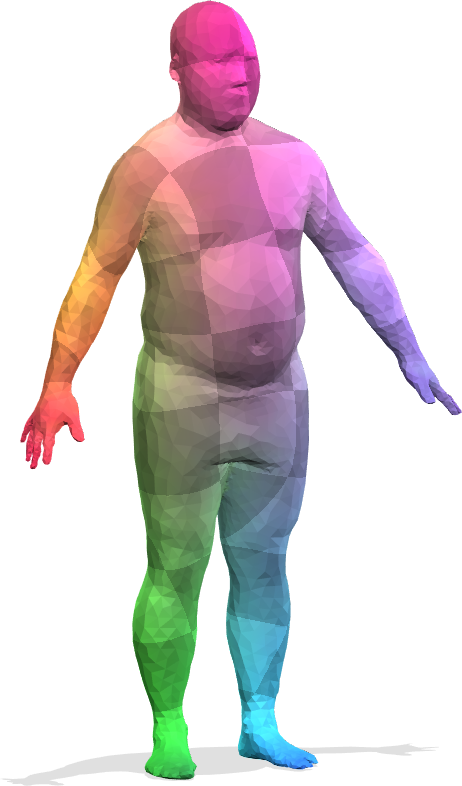}&
        \hspace{\hspaceCols}
        \includegraphics[height=\heightT, width=\widthT]{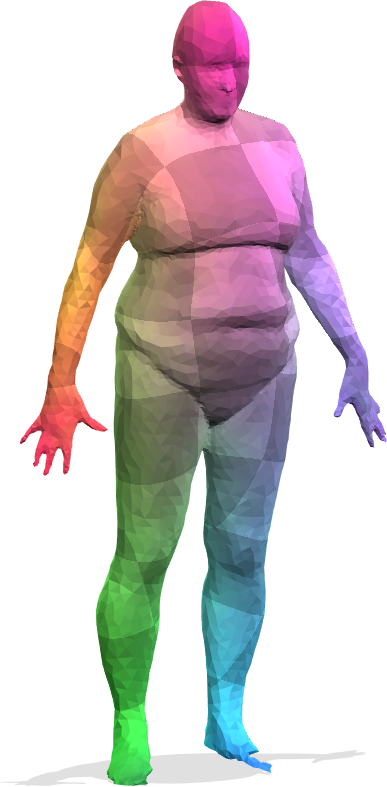}
        \\
    \end{tabular}

%% file: figs/smal_results.tex
\def\rowOnecolumnOne{shape_029-shape_031}
\def\rowOnecolumnTwo{shape_029-shape_037}
\def\rowOnecolumnThree{shape_029-shape_048}
\def\rowOnecolumnFour{shape_029-shape_040}
\def\rowTwocolumnOne{shape_029-shape_044}
\def\rowTwocolumnTwo{shape_029-shape_038}
\def\rowTwocolumnThree{shape_029-shape_039}
\def\rowTwocolumnFour{shape_029-shape_034}

\def\hspaceCols{-0.4cm}
\def\heightT{1.3cm}
\def\widthT{1.7cm}
\begin{tabular}{ccccc}%
        \setlength{\tabcolsep}{0pt} 
        \hspace{\hspaceCols}
        \includegraphics[height=\heightT, width=\widthT]{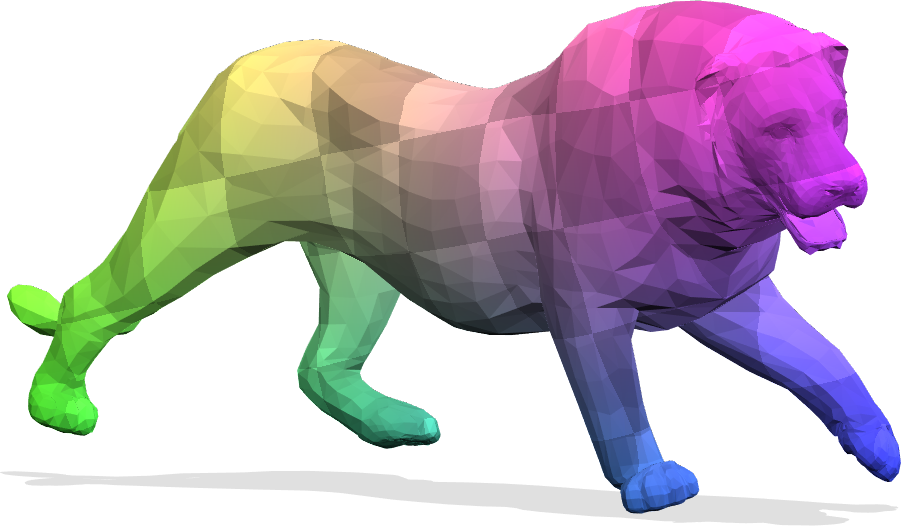}&
        \hspace{\hspaceCols}
        \includegraphics[height=\heightT, width=\widthT]{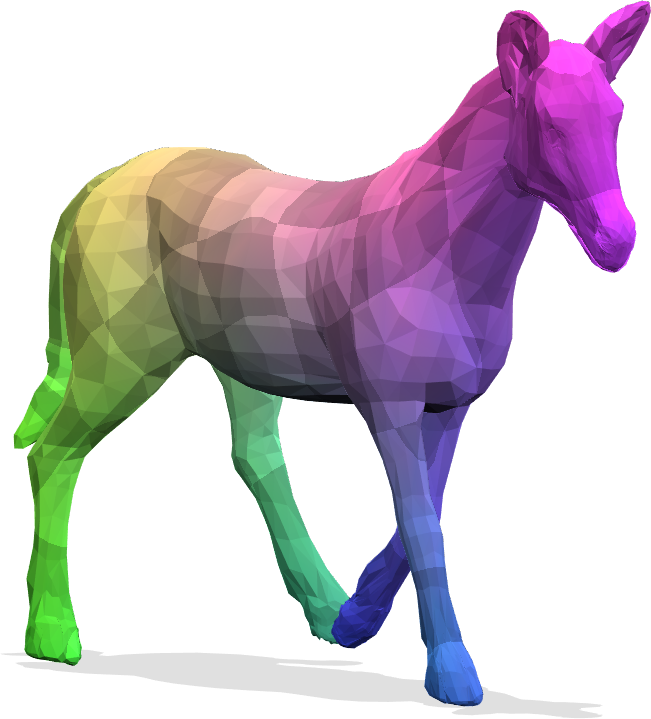}&
        \hspace{\hspaceCols}
        \includegraphics[height=\heightT, width=\widthT]{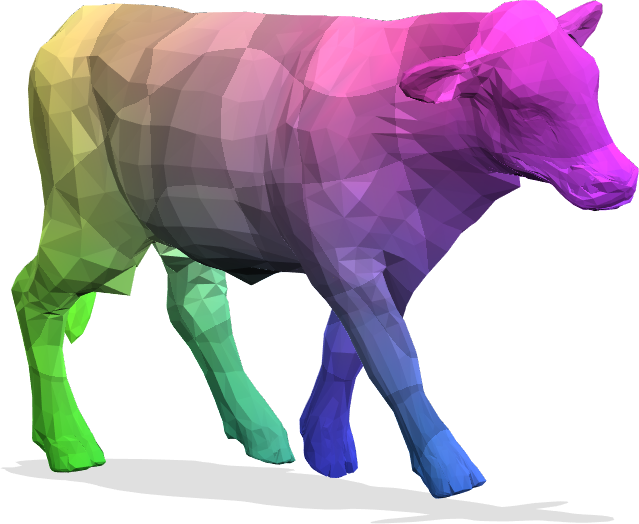}&
        \hspace{\hspaceCols}
        \includegraphics[height=\heightT, width=\widthT]{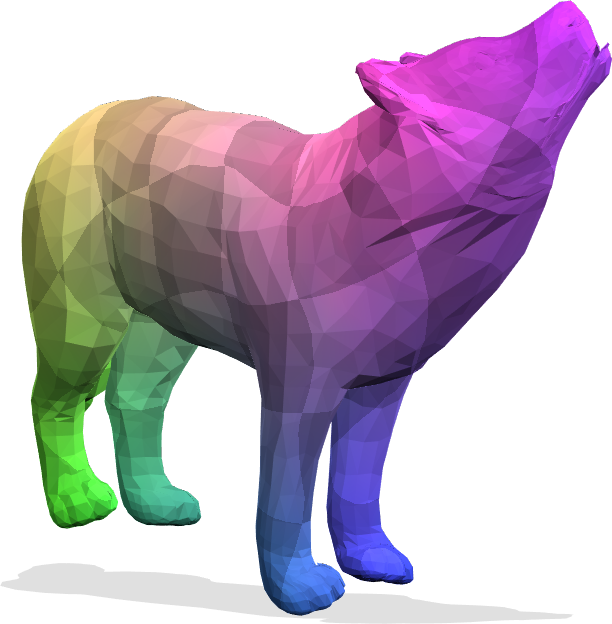}&
        \hspace{\hspaceCols}
        \includegraphics[height=\heightT, width=\widthT]{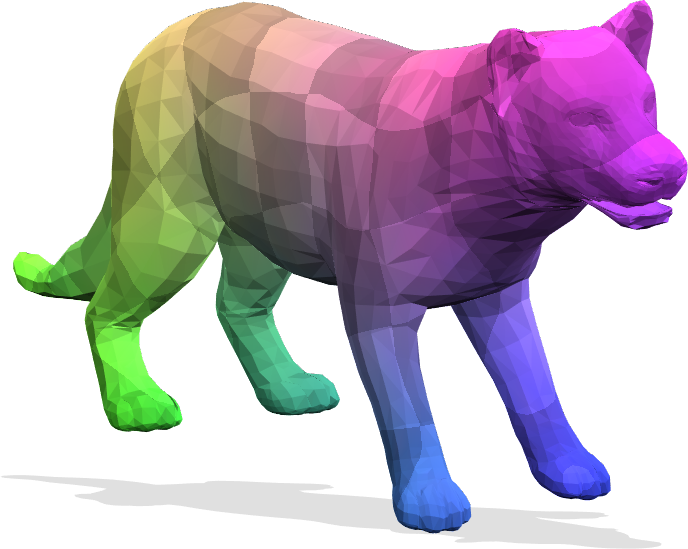}
        \\
        \hspace{\hspaceCols}
        &
        \hspace{\hspaceCols}
        \includegraphics[height=\heightT, width=\widthT]{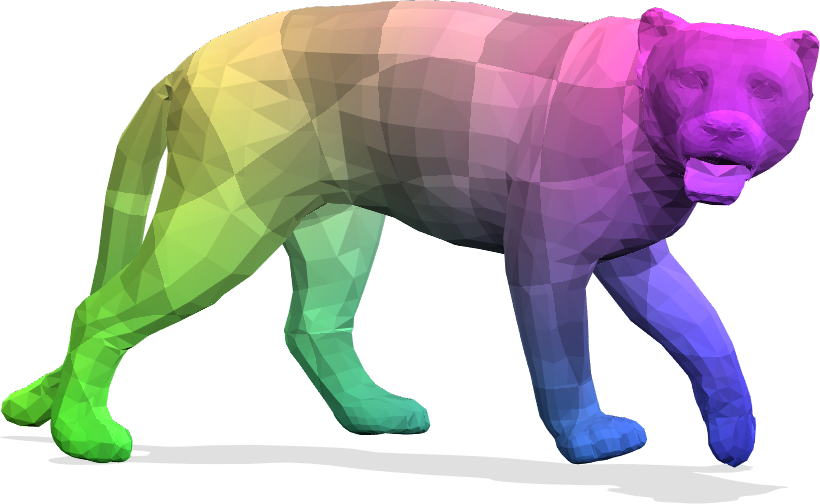}&
        \hspace{\hspaceCols}
        \includegraphics[height=\heightT, width=\widthT]{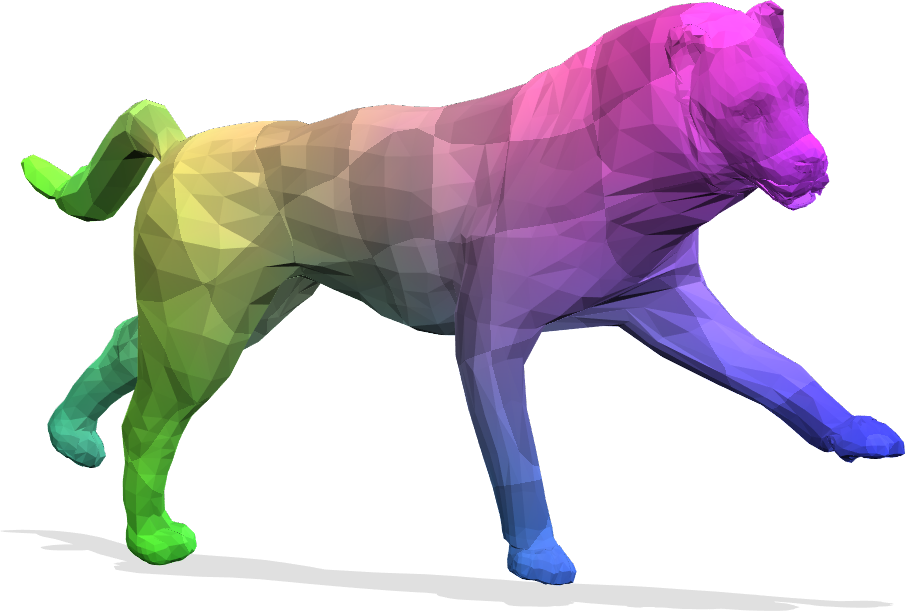}&
        \hspace{\hspaceCols}
        \includegraphics[height=\heightT, width=\widthT]{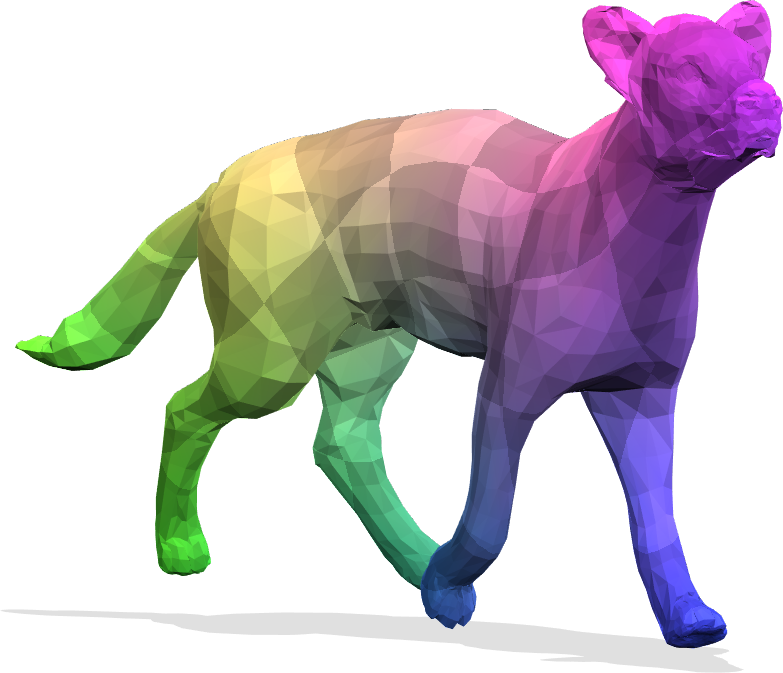}&
        \hspace{\hspaceCols}
        \includegraphics[height=\heightT, width=\widthT]{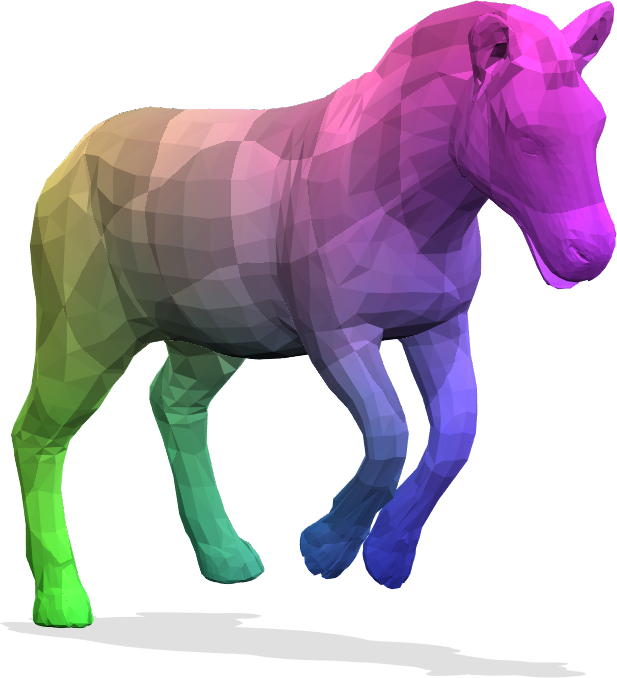}
        \\
    \end{tabular}

%% file: figs/dt4d_results.tex
\def\rowOnecolumnOne{Standing2HMagicAttack01036-StandingCoverTurn007}
\def\rowOnecolumnTwo{Standing2HMagicAttack01036-Floating007}
\def\rowOnecolumnThree{Standing2HMagicAttack01036-GoalkeeperScoop061}
\def\rowOnecolumnFour{Standing2HMagicAttack01036-Strafing000}
\def\rowTwocolumnOne{Standing2HMagicAttack01036-KettlebellSwing057}
\def\rowTwocolumnTwo{Standing2HMagicAttack01036-InvertedDoubleKickToKipUp200}
\def\rowTwocolumnThree{Standing2HMagicAttack01036-Falling271}
\def\rowTwocolumnFour{Standing2HMagicAttack01036-DancingRunningMan259}

\def\hspaceCols{-0.42cm}
\def\heightT{2.5cm}
\def\widthT{2.2cm}
\begin{tabular}{ccccc}
        \setlength{\tabcolsep}{0pt} 
        \hspace{\hspaceCols}
        \includegraphics[height=\heightT, width=\widthT]{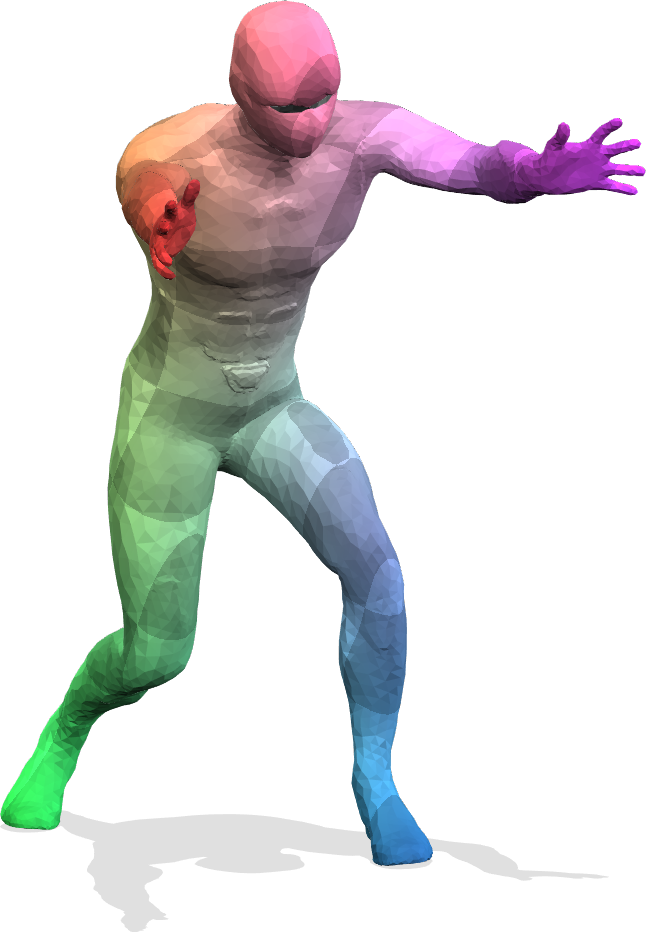}&
        \hspace{\hspaceCols}
        \includegraphics[height=\heightT, width=\widthT]{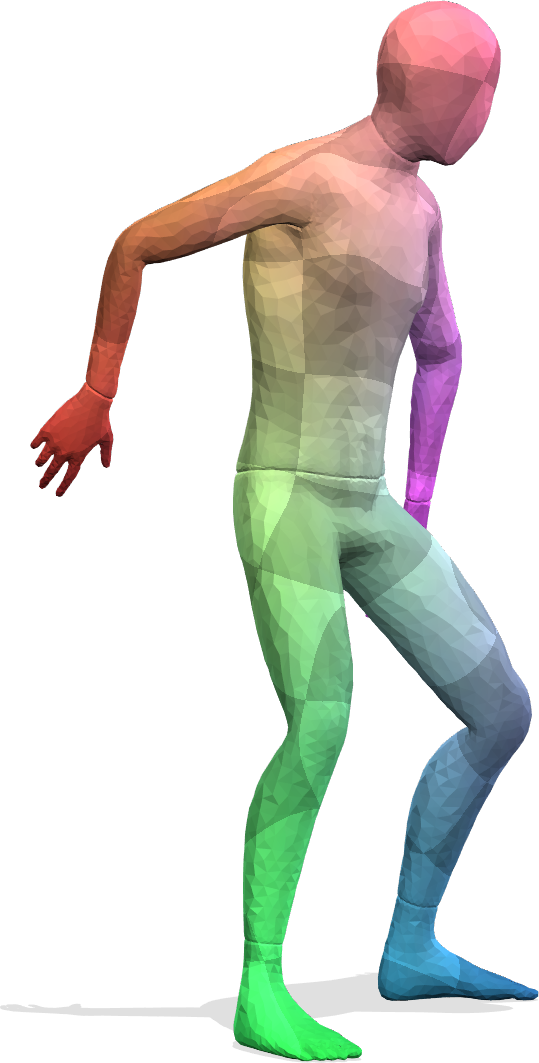}&
        \hspace{\hspaceCols}
        \includegraphics[height=\heightT, width=\widthT]{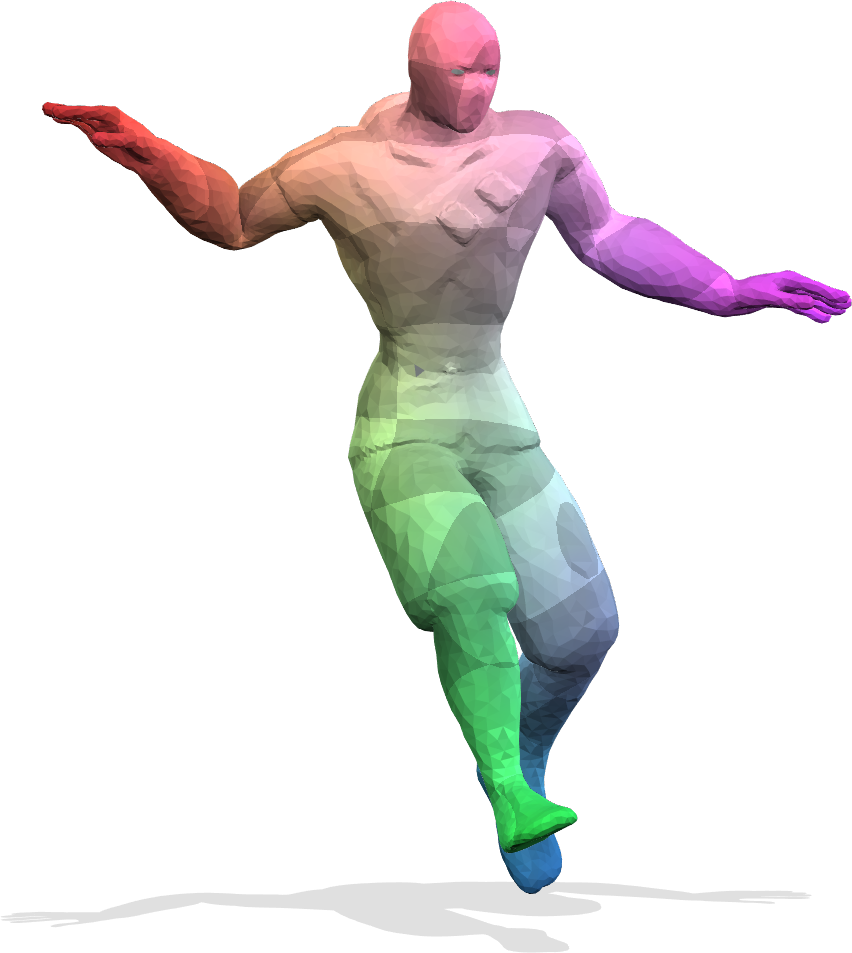}&
        \hspace{\hspaceCols}
        \includegraphics[height=\heightT, width=\widthT]{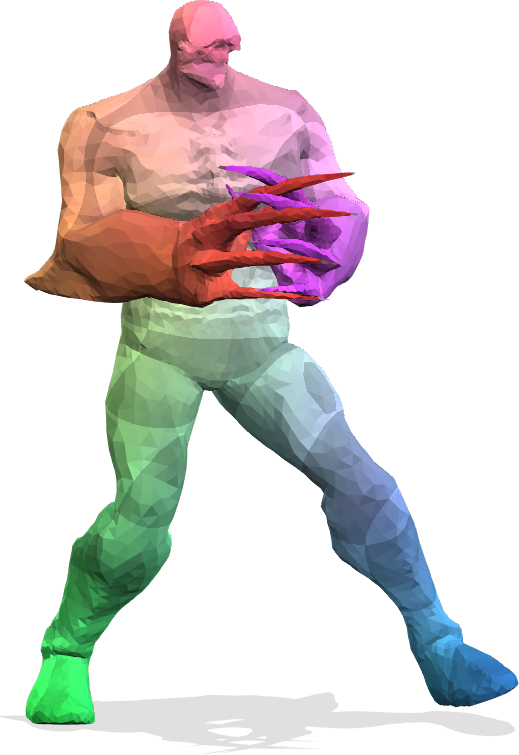}&
        \hspace{\hspaceCols}
        \includegraphics[height=\heightT, width=\widthT]{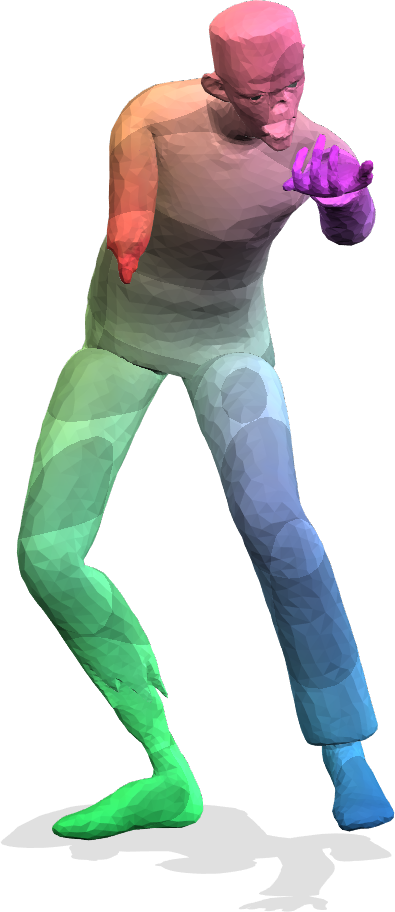}
        \\
        \hspace{\hspaceCols}
        &
        \hspace{\hspaceCols}
        \includegraphics[height=\heightT, width=\widthT]{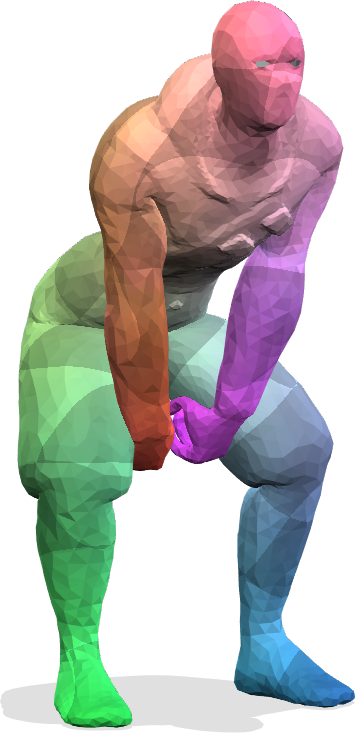}&
        \hspace{\hspaceCols}
        \includegraphics[height=\heightT, width=\widthT]{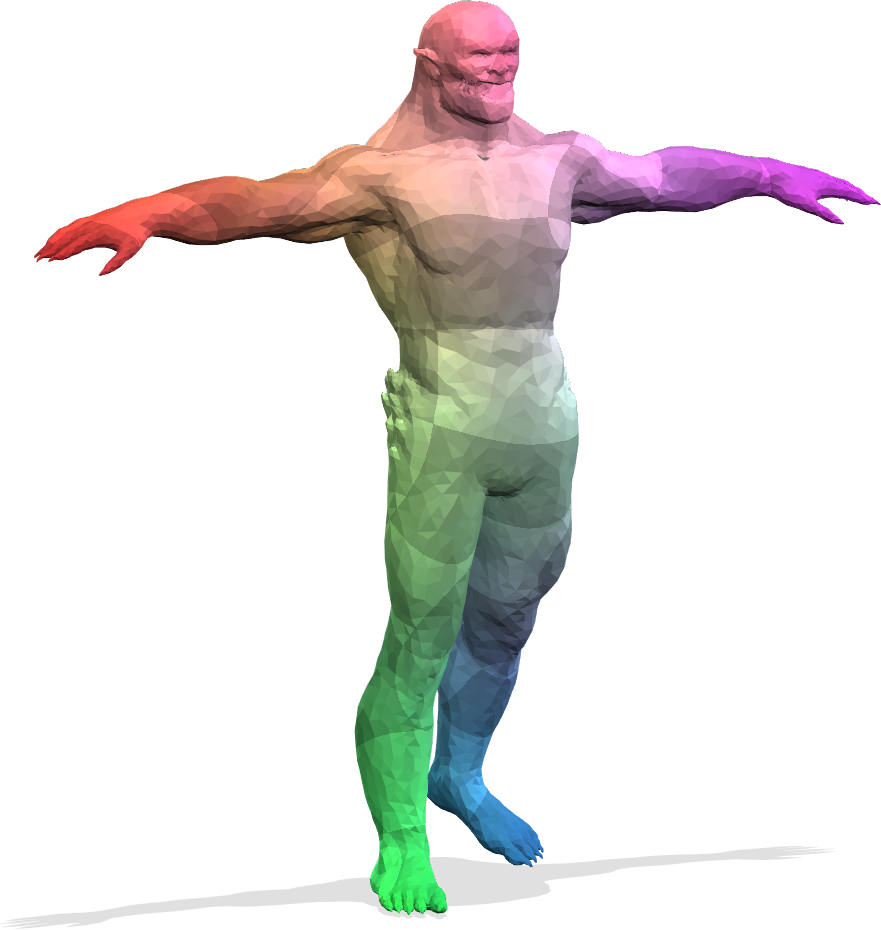}&
        \hspace{\hspaceCols}
        \includegraphics[height=\heightT, width=\widthT]{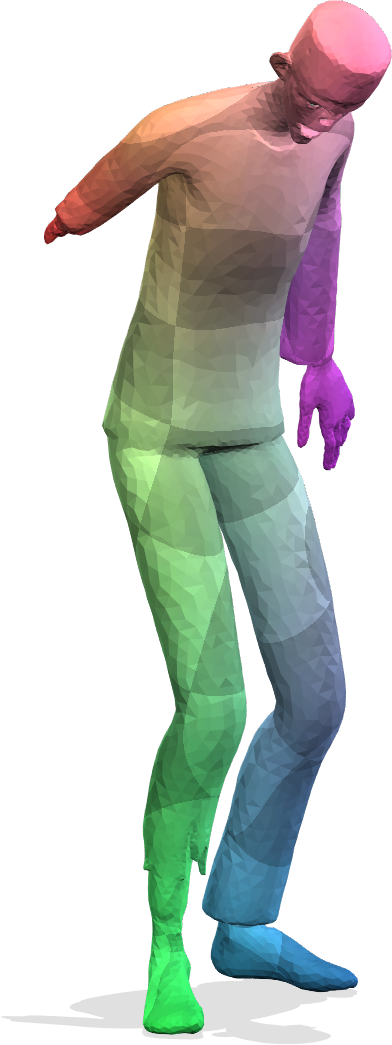}&
        \hspace{\hspaceCols}
        \includegraphics[height=\heightT, width=\widthT]{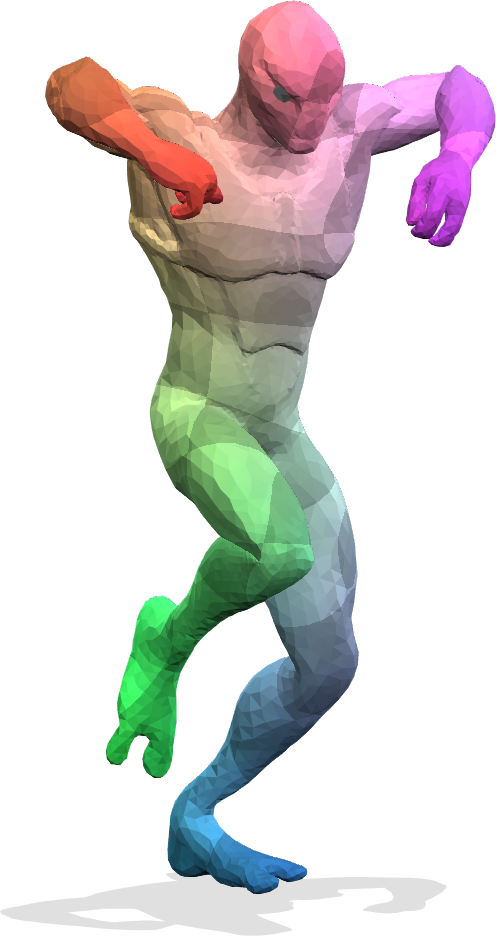}
        \\
    \end{tabular}